\documentclass{article}

\PassOptionsToPackage{numbers, compress, sort}{natbib}
\usepackage[preprint]{neurips_2021}

\newif\ifconf
\newif\iftr
\newif\ifEXT
\EXTfalse

\trtrue
\conffalse

\usepackage[utf8]{inputenc} 
\usepackage[T1]{fontenc}    
\usepackage{url}            
\usepackage{booktabs}       
\usepackage{amsfonts}       
\usepackage{nicefrac}       
\usepackage{microtype}      
\usepackage{xcolor}         

\usepackage{wrapfig}

\newif\ifsq     

\newif\ifsqCAP
\newif\ifsqVS
\newif\ifsqEN
\newif\ifsqTIT

\newcommand{\ignore}[1]{}

\sqtrue
\sqCAPfalse
\sqENtrue
\sqVStrue
\sqTITtrue

\sqfalse
\sqCAPfalse
\sqENfalse
\sqVSfalse
\sqTITfalse

\usepackage{etex}
\usepackage{balance}
\usepackage{epstopdf}
\usepackage{placeins}

\usepackage{graphicx}
\usepackage{float}
\usepackage{dblfloatfix}
\usepackage{multirow}
\usepackage{rotating}
\usepackage{makecell}
\usepackage{tabulary}
\usepackage{parcolumns}
\usepackage{tikz}
\usetikzlibrary{tikzmark}

\usepackage{xpatch}
\expandafter\xpatchcmd
\csname pgfk@/tikz/every picture/.@cmd\endcsname
{\thepage}{\arabic{page}}{}{}

\tikzstyle{comment} = [draw, fill=blue!70, text=white, text width=3cm, minimum height=1cm, rounded corners, align=left, font=\scriptsize]
\tikzstyle{background_alg} = [draw, fill=blue!20, opacity=0.4, inner sep=4pt, rounded corners=2pt]

\usetikzlibrary{shapes}
\usetikzlibrary{plotmarks}
\usetikzlibrary{calc, fit}

\usepackage{enumitem}

\usepackage{amsthm}
\usepackage{amsmath,amssymb,amsfonts}
\usepackage{mathtools,mathrsfs}

\usepackage{soul}
\usepackage{fontawesome}
\usepackage{pifont}
\usepackage{textcomp}
\usepackage{booktabs}
\usepackage{url}
\usepackage{pbox}
\usepackage[normalem]{ulem}
\usepackage[10pt]{moresize}

\ifsqCAP
\usepackage[font={normalfont, scriptsize}]{caption}
\usepackage[font={normalfont, scriptsize}]{subcaption}
\else
\usepackage[font={normalfont, scriptsize}]{caption}
\usepackage[font={normalfont, scriptsize}]{subcaption}
\fi

\newcommand{\vspaceSQ}[1]{\ifsqVS\vspace{#1}\fi}
\newcommand{\enlargeSQ}[1]{\ifsqEN\enlargethispage{\baselineskip}\fi}

\ifsqTIT
\usepackage[compact]{titlesec}
\titlespacing*{\section}{0pt}{3pt}{-1pt}
\titlespacing*{\subsection}{0pt}{0pt}{-3pt}
\titlespacing*{\subsubsection}{0pt}{2pt}{1pt}
\fi

\usepackage{xcolor}
\definecolor{darkgrey}{RGB}{70,70,70}
\definecolor{lightgrey}{RGB}{200,200,200}
\definecolor{lyellow}{RGB}{255,255,100}
\definecolor{llyellow}{RGB}{250,250,180}
\definecolor{lgreen}{RGB}{144,238,144}
\definecolor{raphael_comments}{RGB}{13, 145, 24}

\usepackage[customcolors]{hf-tikz}
\hfsetbordercolor{white}
\hfsetfillcolor{vlgray}

\definecolor{vlgray}{rgb}{0.77 0.77 0.77}
\definecolor{ablack}{rgb}{0.2 0.2 0.2}
\definecolor{vllgray}{rgb}{0.9 0.9 0.9}
\definecolor{bblue}{rgb}{0.7 0.7 0.99}

\usepackage{colortbl}

\usepackage{inconsolata}
\usepackage{listings}

\ifsq
\else
\fi

\newcommand{\maciej}[1]{\textcolor{blue}{[Maciej: #1]}}

\definecolor{hlL}{rgb}{0.8 0.8 0.99}

\newcounter{highlight}

\newcounter{hlLR}

\newcounter{hlLIR}

\newcounter{hlLIIR}

\newcounter{Ahighlight}

\usepackage{scalerel,stackengine}
\stackMath
\newcommand\rwh[1]{%
\savestack{\tmpbox}{\stretchto{%
  \scaleto{%
        \scalerel*[\widthof{\ensuremath{#1}}]{\kern-.6pt\bigwedge\kern-.6pt}%
                  {\rule[-\textheight/2]{1ex}{\textheight}}
                              }{\textheight}%
}{0.5ex}}%
\stackon[1pt]{#1}{\tmpbox}%
}

\usepackage[hang,flushmargin]{footmisc}

\renewcommand{\epsilon}{\ensuremath\varepsilon}

\renewcommand{\phi}{\ensuremath{\varphi}}

\newcommand{\fRB}[1]{\left(#1\right)}
\newcommand{\fSB}[1]{\left[#1\right]}

\usepackage{stmaryrd}

\renewcommand{\arraystretch}{1.0}

\definecolor{lightgray1}{gray}{0.6}
\definecolor{lightgray2}{gray}{0.8}

\renewcommand{\marginpar}[1]{}
\renewcommand{\hl}[1]{#1}
\renewcommand{\colorbox}[2]{#2}

\title{Neural Graph Databases}

\iftr 
\author{%
Maciej Besta$^{1,\dagger}$ \quad Patrick Iff$^1$ \quad Florian Scheidl$^1$ \quad Kazuki Osawa$^1$ \quad Nikoli Dryden$^1$ \\
\textbf{Michal Podstawski}$^{2,3}$ \quad \textbf{Tiancheng Chen}$^1$ \quad \textbf{Torsten Hoefler}$^{1,\dagger}$\\\\
       $^1$Department of Computer Science, ETH Zurich\\
       $^2$Warsaw University of Technology, Warsaw, Poland\\
       \vspace{0.3em}$^3$TCL Research Europe, Warsaw, Poland\\
$^{\dagger}$Corresponding authors:\\ \texttt{\{maciej.besta, torsten.hoefler\}@inf.ethz.ch}
}
\else

\author{Anonymous authors}

\fi

\begin{document}

\maketitle

\vspaceSQ{-1em}
\begin{abstract}
\vspaceSQ{-1em}
Graph databases (GDBs) enable processing and analysis of unstructured, complex,
rich, and usually vast graph datasets. Despite the large significance of GDBs
in both academia and industry, little effort has been made into integrating
them with the predictive power of graph neural networks (GNNs).
In this work, we show how to seamlessly combine nearly any GNN model with the
computational capabilities of GDBs. For this, we observe that the majority of
these systems are based on, or support, a graph data model called the Labeled
Property Graph (LPG), where vertices and edges can have arbitrarily complex
sets of labels and properties. We then develop LPG2vec, an encoder that
transforms an arbitrary LPG dataset into a representation that can be directly
used with a broad class
of GNNs, including convolutional, attentional, message-passing, and even
higher-order or spectral models.
In our evaluation, we show that the rich information represented as LPG labels
and properties is properly preserved by LPG2vec, and it increases the accuracy
of predictions regardless of the targeted learning task or the used GNN model,
by up to 34\% compared to graphs with no LPG labels/properties.
In general, LPG2vec enables combining predictive power of the most powerful
GNNs with the full scope of information encoded in the LPG model, paving the
way for neural graph databases, a class of systems where the vast complexity of
maintained data will benefit from modern and future graph machine learning
methods.
\end{abstract}


\vspaceSQ{-0.5em}
\section{Introduction}
\label{sec:intro}

Graph databases are a class of systems that enable storing, processing,
analyzing, and the overall management of large and rich graph
datasets~\cite{besta2019demystifying}.
They are heavily used in computational biology and chemistry,
medicine, social network analysis, recommendation and online purchase
infrastructure, and many others~\cite{besta2019demystifying}.
A plethora of such systems exist, for example Neo4j~\cite{miller2013graph} (a
leading industry graph database)\footnote{According to the DB engines ranking
(\url{https://db-engines.com/en/ranking/graph+dbms})},
TigerGraph~\cite{tigergraph2022ldbc, tiger_graph_links},
JanusGraph~\cite{janus_graph_links}, Azure Cosmos
DB~\cite{azure_cosmosdb_links}, Amazon Neptune~\cite{amazon_neptune_links},
Virtuoso~\cite{virtuoso_links}, ArangoDB~\cite{arangodb_links,
arangodb_indexing_links, arangodb_starter},
OrientDB~\cite{orientdb_lwedge_links, tesoriero2013getting}, and
others~\cite{graphdb_links, redisgraph_links, dgraph_links,
allegro_graph_links, anzo_graph_links, datastax_links, infinite_graph_links,
blaze_graph_links, oracle_spatial, stardog_links, cayley_links, weaver_links}. 
Graph databases differ from other classes of graph-related systems and
workloads such as graph streaming frameworks~\cite{besta2019practice} in that
they deal with transactional support, persistence, physical/logical data
independence, data integrity, consistency, and complex graph data models where
both vertices and edges may be of different classes and may be associated with
arbitrary properties.

\begin{figure}[t]
\vspaceSQ{-1em}
\centering
\includegraphics[width=1\textwidth]{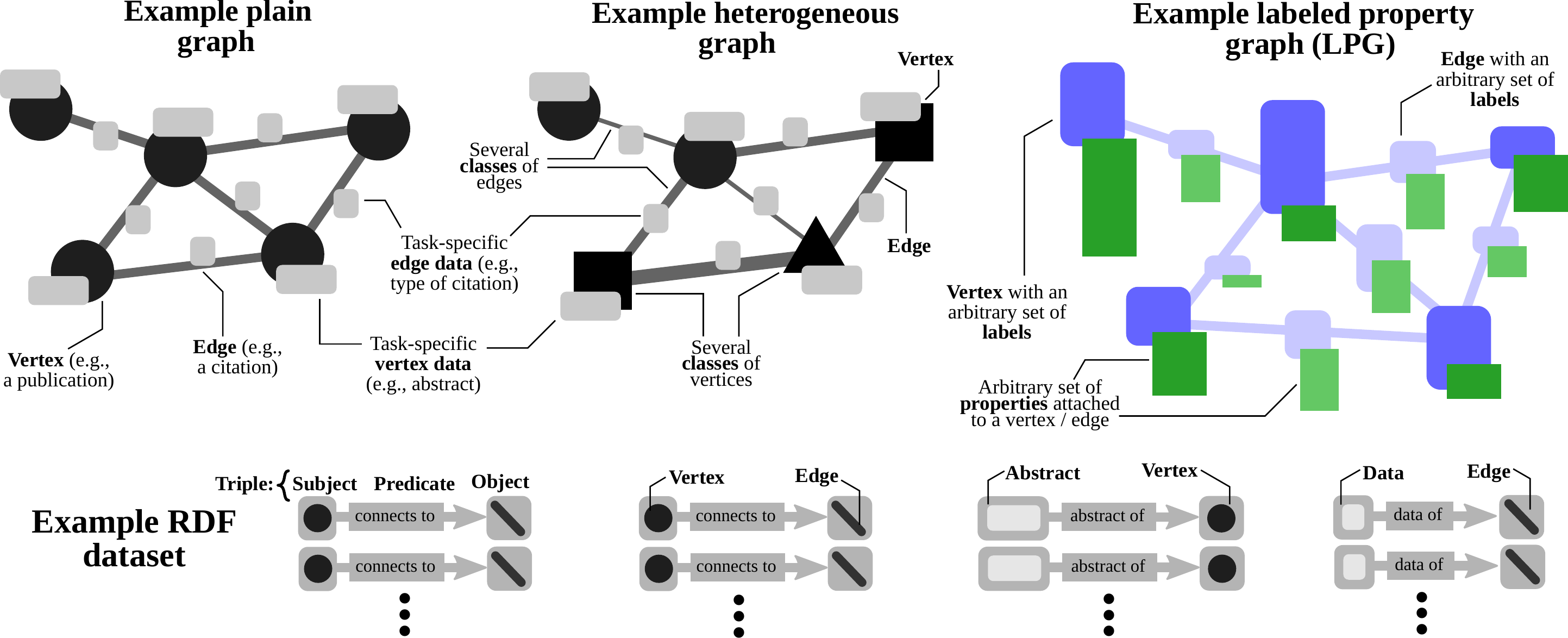}
\caption{Overview of the Labeled Property Graph (LPG) data model used in graph databases,
vs.~plain and heterogeneous graphs used in broad graph processing and graph machine learning, and RDF triples.}
\label{fig:lpg-quick}
\vspaceSQ{-1em}
\end{figure}

An established data model used in the majority of graph databases is called the
\emph{Labeled Property Graph} (LPG)~\cite{besta2019demystifying}.
It is the model of choice for the leading industry Neo4j graph database system.
LPG has several advantages over other
graph data models, such as heterogeneous graphs~\cite{wang2020survey,
yang2020heterogeneous, xie2021survey} or the Resource Description Framework
(RDF)~\cite{lassila1998resource} graphs, often referred to as knowledge graphs
(see Figure~\ref{fig:lpg-quick}).
First, while heterogeneous graphs support different classes of vertices and
edges, LPGs offer \emph{arbitrary sets of labels as well as key-value property
pairs} that can be attached to vertices and edges. This facilitates modeling
very rich and highly complex data. For example, when modeling publications with
graph vertices, one can use labels to model an arbitrarily complex hierarchy of
types of publications (journal, conference, workshop papers; best papers, best
student papers, best paper runner-ups, etc.).  We discuss this example further
in Section~\ref{sec:back}.
Second, LPG explicitly stores the neighborhood structure of the graph, very
often in the form of adjacency lists~\cite{besta2019demystifying}. Hence, it
enables very fast accesses to vertex neighborhoods and consequently fast and
scalable graph algorithms and graph queries. This may be more difficult to
achieve in data representations such as sets of triples. First, \emph{any
possible relation between any two entities in a graph} (i.e., edges, vertices,
and \emph{any} other data) is explicitly maintained as a separate triple (see
Figure~\ref{fig:lpg-quick}). Second, \emph{any entity is fundamentally the same
``resource''}, where ``vertex'' or ``edge'' are just roles assigned to a given
resource; these roles can differ in different triples (i.e., one resource can
be both a vertex and an edge, depending on a specific triple). Hence, RDF
graphs may need more storage, and they may require more complex indexing
structures for vertex neighborhoods, than in the corresponding LPGs.

\marginpar{\Large\vspace{-10em}\colorbox{yellow}{\textbf{x3Ya}}}

Graph neural networks (GNNs)
have recently become an established part of the machine
learning (ML) landscape~\cite{wu2020comprehensive, zhou2020graph,
zhang2020deep, chami2020machine, hamilton2017representation,
bronstein2017geometric,
hamilton2020graph,
besta2021motif, gianinazzi2021learning,
scarselli2008graph, besta2022parallel}.
Example applications are node, link, or graph classification or regression in
social sciences, bioinformatics, chemistry, medicine, cybersecurity,
linguistics, transportation, and others.
GNNs have been successfully used to provide cost-effective and fast placement
of chips~\cite{mirhoseini2021graph}, improve the accuracy of protein folding
prediction~\cite{jumper2021highly}, simulate complex
physics~\cite{pfaff2020learning, sanchez2020learning}, or guide mathematical
discoveries~\cite{davies2021advancing}.
The versatility of GNNs brings a promise of enhanced analytics capabilities
in the graph database landscape.

\marginpar{\Large\vspace{1em}\colorbox{yellow}{\textbf{bJap}}}

Recently, \hl{Neo4j Inc., Amazon, and others} have
started to investigate harnessing graph ML capabilities into
their graph database architectures. However, current efforts only enable
limited learning functionalities that do {not} take advantage of the full richness
of data enabled by LPG. For example, Neo4j's Graph Data Science
module~\cite{hodlergraph} supports obtaining embeddings and using them for node
or graph classification. However, these embeddings are based on the
graph {structure}, with limited support for taking advantage of the full scope of information
provided by LPG labels and properties.

Combining LPG-based graph databases with GNNs could facilitate reaching new
frontiers in analyzing complex unstructured datasets, and it could also
illustrate the potential of GNNs for broad industry.
%
%
%
In this work, we first broadly investigate both the graph database setting
and GNNs to find the best approach for combining these two.
As a result, we develop \textbf{LPG2vec}, an encoder that enables harnessing
the predictive power of GNNs for LPG graph databases.
In LPG2vec, we treat labels and properties attached to a vertex~$v$ as an
additional source of information that should be integrated with $v$'s input
feature vectors.  For this, we show how to encode different forms of data
provided in such labels/properties.
This data is transformed into embeddings that can seamlessly be used with
different GNN models. LPG2vec is orthogonal to the software design
and can also be used with any GNN framework or graph database implementation.

We combine LPG2vec with three established GNN models (GCN~\cite{kipf2016semi},
GIN~\cite{xu2018powerful}, and GAT~\cite{velivckovic2017graph}), and we show that  
the information preserved by LPG2vec consistently 
enhances the accuracy of graph ML tasks, i.e., classification or
regression of nodes and edges, by up to 34\%.
Moreover, LPG2vec supports the completion of missing labels and properties in
often noisy LPGs.
Overall, it enables \textbf{Neural Graph Databases}: the first learning
architecture that enables harnessing {both} the structure {and} rich data
(labels, properties) of LPG for highly accurate predictions in graph databases.

\enlargeSQ

\section{Background}
\vspaceSQ{-0.3em}
\label{sec:back}

We first introduce fundamental concepts and notation for the LPG model and GNNs.

\subsection{Labeled Property Graph Data Model}

{Labeled Property Graph Model (LPG)}~\cite{besta2019demystifying} (also
called the property graph~\cite{gdb_query_language_Angles}) is a primary
established data model used in graph
databases.
We focus on LPG because it is supported by the majority of systems, and
is a model of choice in many leading ones~\cite{besta2019demystifying} (see Section~\ref{sec:intro}).
%

\begin{wrapfigure}{r}{0.65\linewidth}
%
%
\centering
\includegraphics[width=0.65\columnwidth]{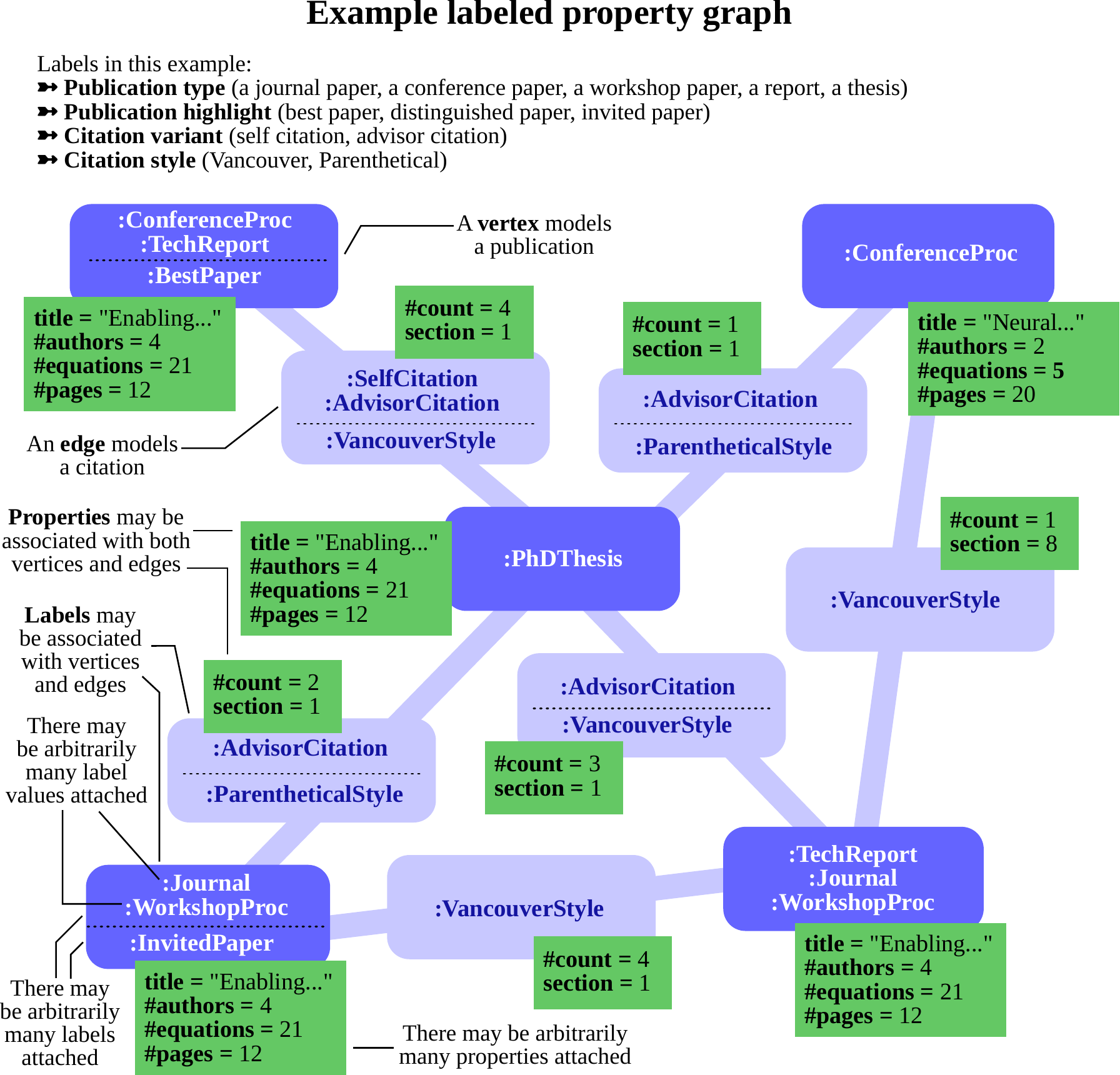}
\vspaceSQ{-0.5em}
\caption{An example of an LPG graph modeling
publications and citations between them.}
\label{fig:lpg-example}
\vspaceSQ{-0.5em}
%
%
\end{wrapfigure}

At its core, LPG is based on the {plain graph model} $G = (V,E)$,
where $V$ is a set of vertices and $E \subseteq V \times V$ is a set of edges;
$|V|=n$ and $|E|=m$. An edge $e=(u,v)\in E$ is a tuple of the out-vertex $u$
(origin) and the in-vertex $v$ (target). If $G$ is undirected, then an edge
$e=\{u,v\}\in E$ is a set of $u$ and $v$.
$N_i$ and $d_i$ denote the neighbors and the degree of a given vertex~$i$
($N_i \subset V$); $d$ is $G$'s maximum degree.
LPG then adds arbitrary \emph{labels} and \emph{properties} to vertices and
edges. An LPG is formally modeled as a tuple
$(V,E,L,l,K,W,p)$.
$L$ is a set of labels. $l: V \cup E \mapsto \mathcal{P}(L)$ is a labeling function, mapping --
respectively -- each vertex and each edge to a subset of labels, where $\mathcal{P}(L)$ is
the power set of $L$, containing all possible subsets of $L$.
In addition to labels, each vertex and edge can have arbitrarily many
\emph{properties} (sometimes referenced as attributes). A property is a $(key,
value)$ pair, with $key$ being an identifier and $value$ being a corresponding
value.
Here, $\mathnormal{K}$ is a set with all possible keys and $\mathnormal{W}$ is
a set with all possible values. For any property, we have $key \in
\mathnormal{K}$ and $value \in \mathnormal{W}$. Then, $p: (V \cup E) \times \mathnormal{K} \mapsto \mathnormal{W}$ 
is a mapping function from vertices/edges to property values.
Specifically,
$p(u, key)$ and $p(e, key)$ assign -- respectively -- a value to a property indexed with a key $key$, of a vertex $u$ of an edge $e$.
Note that one can assign multiple properties with the
same key to vertices and edges (i.e., only the pair $(key, value)$ must be
unique).

We illustrate an example of an LPG graph in Figure~\ref{fig:lpg-example}.

\subsection{Graph Neural Networks}

Graph neural networks (GNNs) are a class of neural networks that enable
learning over irregular graph datasets~\cite{scarselli2008graph}.
Each vertex (and often each edge) of the input graph usually comes with an
\emph{input feature vector} that encodes the semantics of a given task. For
example, when vertices and edges model publications and citations between these
papers, then a vertex input feature vector is a encoding of the publication
abstract (e.g., a one-hot bag-of-words encoding specifying which words are
present).  Input feature vectors are transformed in a series of GNN layers. In
this process, intermediate {hidden latent vectors} are created. The last GNN
layer produces output feature vectors, which are then used for the
\emph{downstream ML tasks} such as node classification or graph classification.

\begin{wrapfigure}{r}{0.5\linewidth}
\vspaceSQ{-1.0em}
%
%
\centering
\includegraphics[width=0.5\columnwidth]{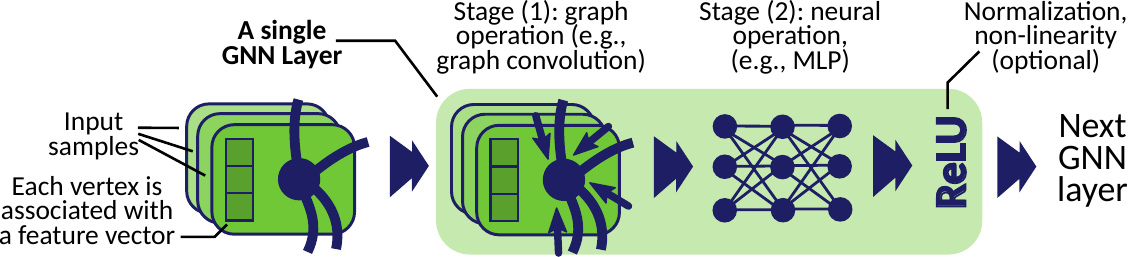}
\vspaceSQ{-0.5em}
\caption{Overview of a single GNN layer.}
\label{fig:gnn-layer}
\vspaceSQ{-2.5em}
%
%
\end{wrapfigure}

Many GNN models exist~\cite{zhang2019heterogeneous, zhou2020graph,
thekumparampil2018attention, wu2020comprehensive, sato2020survey, wu2020graph,
zhang2020deep, chen2020bridging, cao2020comprehensive, besta2022parallel}. Most
of such models consist of a series of GNN layers, and a single layer has two
stages: (1) the aggregation stage that -- for each vertex -- combines the
features of the neighbors of that vertex, and (2) the neural stage that
combines the results of the aggregation with the vertex score from the previous
layer into a new score. One may also explicitly distinguish stage~(3), a
non-linear activation over feature vectors (e.g., ReLU~\cite{kipf2016semi})
and/or normalization.
%
%
We illustrate a simplified view of a GNN layer in Figure~\ref{fig:gnn-layer}.

The input, output, and hidden feature vector of a vertex~$i$ are denoted with,
respectively, $\mathbf{x}_i, \mathbf{y}_i, \mathbf{h}_i$. We have $\mathbf{x}_i
\in \mathbb{R}^k$ and $\mathbf{y}_i, \mathbf{h}_i \in \mathbb{R}^{O(k)}$, $k$
is the dimensionality of vertex input feature vectors. 
%
%
``$(l)$'' denotes the $l$-th GNN layer; $\mathbf{h}^{(l)}_i$ are latent features in layer~$l$.

Formally, the graph aggregation stage of a GNN layer can be described using two
functions, $\psi$ and $\bigoplus$.  First, the feature vector of each neighbor
of~$i$ is transformed by a function~$\psi$. Then, the resulting neighbor
feature vectors are aggregated using a function~$\bigoplus$, such as sum or
max.
The outcome of $\bigoplus$ is then processed using a third function, $\phi$,
that models the neural operation and non-linearity.  This gives the latent
feature vector~$\mathbf{h}_i$ in the next GNN layer. 
Combined, we have
\iftr

\vspace{-2em}
\begin{gather}
\mathbf{h}^{(l+1)}_i = \phi \left( \mathbf{h}^{(l)}_i, \bigoplus_{j \in N(i)} \psi\left(\mathbf{h}^{(l)}_i, \mathbf{h}^{(l)}_j \right) \right) \label{eq:mpgnn}
\end{gather}

\else
$\mathbf{h}^{(l+1)}_i = \phi \left( \mathbf{h}^{(l)}_i, \bigoplus_{j \in N(i)} \psi\left(\mathbf{h}^{(l)}_i, \mathbf{h}^{(l)}_j \right) \right)$.
\fi
This is a generic form of GNNs, which can be used to define 
three major classes of GNNs~\cite{bronstein2021geometric}: \emph{convolutional}
GNNs (C-GNNs; examples are GCN~\cite{kipf2016semi},
GraphSAGE~\cite{hamilton2017inductive}, GIN~\cite{xu2018powerful}, and
CommNet~\cite{sukhbaatar2016learning}), \emph{attentional} GNNs (A-GNNs; examples are
MoNet~\cite{monti2017geometric}, GAT~\cite{velivckovic2017graph}, and
AGNN~\cite{thekumparampil2018attention}), and the most generic \emph{message-passing}
GNNs (MP-GNNs; examples are G-GCN~\cite{bresson2017residual},
EdgeConv~\cite{wang2019dynamic}, MPNN~\cite{gilmer2017neural}, and
GraphNets~\cite{battaglia2018relational}).

\iftr
One can alternatively formulate a GNN model using matrices grouping
the graph structure (the adjacency matrix $\mathbf{A}$), the feature vectors ($\mathbf{H}$), and the parameters ($\mathbf{W}$), and the
operations on these matrices such as sparse matrix
products~\cite{gleinig2022io} and many others~\cite{petersen2008matrix}.
Many of these models can be formulated as

\vspace{-2em}
\begin{gather}
\mathbf{H}^{(l+1)} = \fRB{ f\fRB{\mathbf{A}} \odot g \fRB{\mathbf{H}^{(l)}, \mathbf{W}^{(l)}} } \times \mathbf{H}^{(l)} \times \mathbf{W}^{(l)}
\end{gather}

\noindent
where $f$ is usually some preprocessing transformation applied to the adjacency matrix (e.g., row normalization),
$\odot$ is the Hadamard product, and $g$ is some function of the feature vectors and model parameters,
such as the vanilla attention ($g = \mathbf{H}^{(l)} \times {\mathbf{H}^{(l)}}^T$).
Our work is also compatible with such global
formulations~\cite{hamilton2020graph, besta2022parallel}, and with higher-order
associated models such as Chebnet~\cite{defferrard2016convolutional},
DeepWalk~\cite{perozzi2014deepwalk}, Node2vec~\cite{grover2016node2vec}, and
others~\cite{lin2015learning, wang2016structural, ying2018graph,
klicpera2018predict, bojchevski2020scaling, bianchi2021graph}. 

An extensive description of these and other GNN model classes is provided in a recent analysis~\cite{besta2022parallel}. 

\fi

\hl{To avoid confusion, we always use a term ``label'' to denote an LPG label, while a term ``class''
indicates a prediction target in classification tasks.}

\vspaceSQ{-0.5em}
\section{Marrying Graph Databases and Graph Neural Networks}
\vspaceSQ{-0.5em}

\enlargeSQ

We first investigate how LPG-based graph databases and GNNs can
be combined to reach new frontiers of complex graph data analytics.

\marginpar{\Large\vspace{1em}\colorbox{yellow}{\textbf{Cmds}}\\ \colorbox{yellow}{\textbf{bJap}}}

\vspaceSQ{-0.25em}
\subsection{\colorbox{yellow}{How to Use GNNs with GDBs?}} 
\vspaceSQ{-0.5em}

\hl{It is not immediately clear on how to use GNNs in combination with the
LPG data model.
Specifically, labels and properties of a vertex~$v$ are often seen as additional
``vertices'' attached to $v$~\mbox{\cite{neo3j_release, tdan}}. From this
perspective, it would seem natural to use them during the aggregation phase of
a GNN computation, together with the neighbors of $v$. Similarly, one
could consider incorporating attentional GNNs~\mbox{\cite{bronstein2021geometric}},
by attending to individual labels and properties.
In general, there can be many different approaches for integrating GNNs and LPGs.}

\hl{Here, we first extensively investigated both the graph database (GDB) and the GNN settings.
The goal was to determine the best approach for
using GNNs with GDBs in order to benefit the maximum number of different GDB
workloads while ensuring a seamless integration with as many GNN models as
possible.
We consider all major classes of GDB workloads: online transactional,
analytical, and serving processing (respectively, OLTP, OLAP,
OLSP)~\mbox{\cite{besta2019demystifying}}, and the fundamental GNN model
classes (e.g., C-GNN, A-GNN, MP-GNN)~\mbox{\cite{bronstein2021geometric}}, for
a total of more than 280 analyzed publications or reports.}

\hl{Our analysis indicated that the most versatile approach for extracting the
information from LPG 
labels and properties is based on encoding labels and properties directly into the input feature vectors, 
and subsequently feeding such vectors into a selected GNN model. First, this approach only
requires modifications to the input feature vectors, which makes LPG2vec fully
compatible with any C-GNN, A-GNN, or MP-GNN model (and many others). Second, this approach is
very similar in its workflow to schemes such as positional encodings: is is based 
on preprocessing and feeding additional information into input feature
vectors. Hence, it is straightforward to integrate into existing GNN infrastructures.}
%

\vspaceSQ{-0.25em}
\subsection{Use Cases and Advantages} 


The first advantage of combining graph databases with GNNs is enhancing the
accuracy of traditional GNN tasks: classification and regression of nodes,
edges, and graphs (note that tasks such as clustering or link prediction can be
expressed as node/edge classification/regression).
This is because LPG labels and properties, when incorporated into input feature
vectors, carry additional information.
This is similar to how different classes of vertices/edges in heterogeneous
graphs enhance prediction tasks~\cite{yang2020heterogeneous,
xie2021survey}. However, the challenge is how to incorporate the full rich set
of LPG information, i.e., multiple labels and properties, into the learning
workflow, while achieving high accuracy and without exacerbating running times or
memory pressure.
%


GNNs can also be used to deliver novel prediction tasks suited for LPG, namely
\textbf{label prediction} and \textbf{property prediction}. In the former, one
is interested in assessing whether a given vertex or edge potentially has a
specified label, i.e., whether $label \in l(v)$ or $label \in l(e)$, where $label \in
L, v \in V, e \in E$ are -- respectively -- a label, a vertex, and an edge of
interest, and $l$ is a labeling function.
In the latter, one analogously asks whether a given vertex or edge potentially
has a specified property, and -- if yes -- what its value is, i.e., whether $property
\in p(v)$ or $property \in p(e)$, where $p = (k, w), k \in K, w \in W, v \in V, e
\in E$ are -- respectively -- a property, a vertex, and an edge of interest,
and $p$ maps $v$ and $e$ to their corresponding
properties.

Here, we observe that predicting new labels can be seamlessly resolved
with node/edge classification, with the target learned label being
$l$. Similarly, property prediction is effectively node/edge regression, where
$w$ is the learned value. 
Thus, it means that one can easily use existing GNN models for \emph{LPG
graph completion} tasks, i.e., finding missing labels or properties
in the often noisy datasets.

\subsection{LPG2vec + GNN: Towards A Neural Graph Database}

\begin{figure}[t]
\centering
\includegraphics[width=1\textwidth]{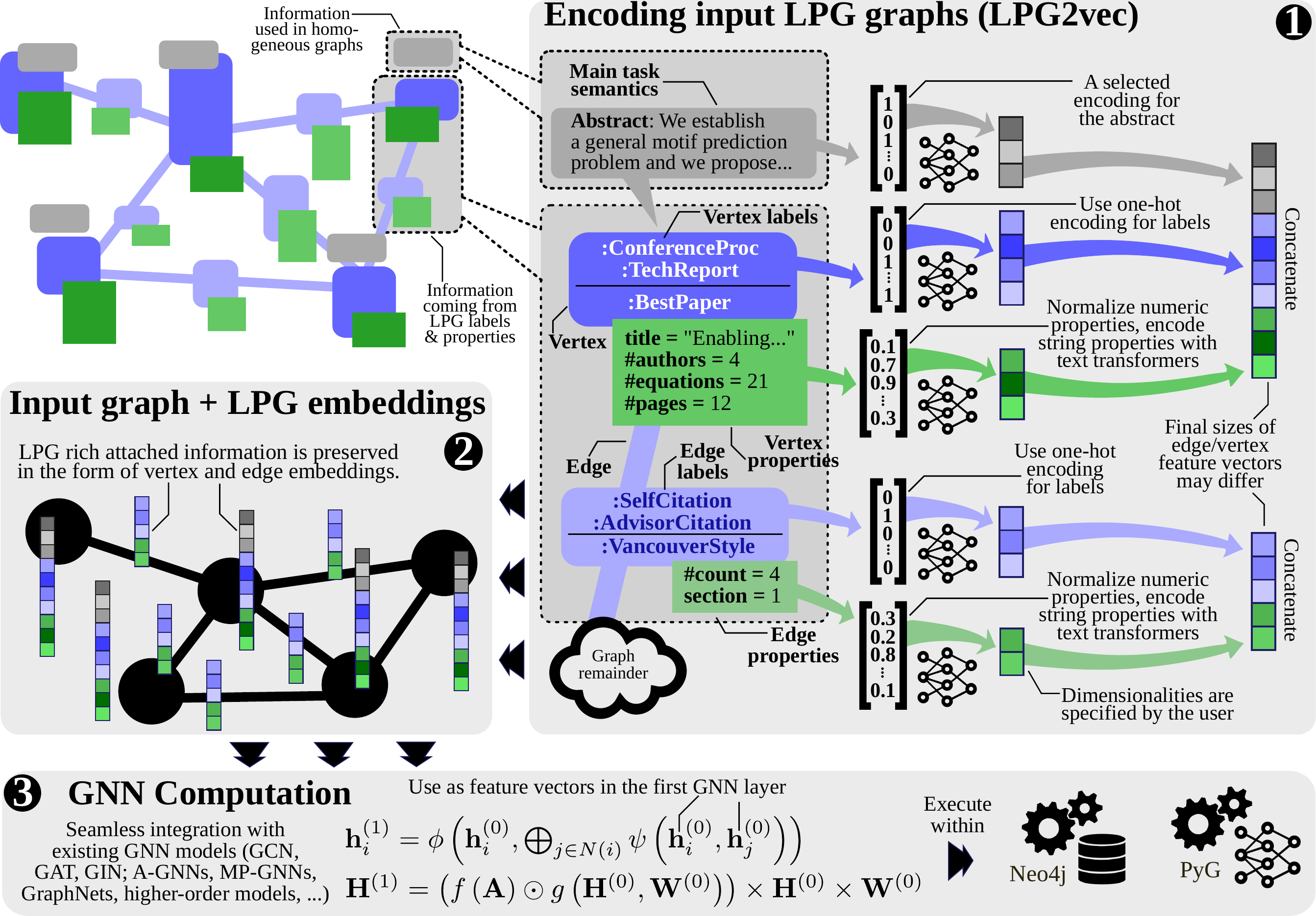}
\caption{An overview of LPG2vec in the context of processing LPG graphs with
GNNs.
First (``\protect\includegraphics[scale=0.15,trim=0 16 0 0]{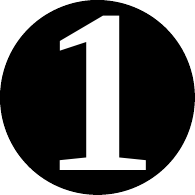}''), the
graph data is loaded from disk and encoded using LPG2vec. 
Here, we differentiate the additional data usually used with homogeneous
graphs, that determines the task semantics (in this case, publication
abstracts), from the LPG-related additional data (labels, properties). Note
that, in practice, encoding the abstract could be just implemented as encoding
an additional property.
The encoding process gives a graph dataset
(``\protect\includegraphics[scale=0.15,trim=0 16 0 0]{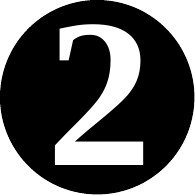}'') that is ready
for the actual GNN computation that can be executed in a dedicated module of a
  graph database (e.g., Neo Graph Data Science) or in a dedicated ML framework
  (e.g., PyG).
Importantly, LPG2vec preserves all the rich LPG information in the
form of vertex and edge embeddings. Thus, the actual input to the GNN
computation is a homogeneous graph structure together with the embeddings. This
makes the integration with existing GNN models straightforward
(``\protect\includegraphics[scale=0.15,trim=0 16 0 0]{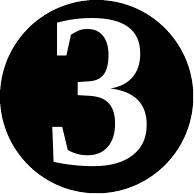}'').
The computation itself is conducted in a dedicated
module of a graph database (e.g., Neo4j's Graph Data Science),
but - thanks to the seamless LPG2vec design (i.e., the fact that
the output of LPG2vec is a homogeneous graph with enhanced feature vectors) - it can
also be conducted in a standalone GNN framework (e.g., PyG).}
\label{fig:overview}
\end{figure}

Our architecture for neural graph databases can be seen as an encoder
combined with a selected GNN model.
An overview is provided in Figure~\ref{fig:overview}.

In the first step to construct an embedding of an LPG, we apply one-hot
encoding for labels and properties of each vertex and edge.
For labels, the encoding is $\{0,1\}^{|L|}$, where ``1'' indicates that a given
$i$th label is attached to a given vertex/edge.
For properties, the encoding details depend on the property type:
If a property can have \emph{discretely many ($C$) values}, then we encode it
using a plain one-hot vector with $C$ entries.
A \emph{continuous scalar} property is normalized to $[0;1]$ or, alternatively,
discretized and encoded as a one-hot vector. Importantly, one must use the same
norm or discretization for all property instances for a given property type.
A \emph{numerical vector} is standardized and normalized.
Finally, for properties that contain a \emph{string of text}, we use Sentence
Transformers, based on sentence-BERT~\cite{reimers2019sentence}, to embed such
a property. String embeddings are usually much longer than other numerical
properties to preserve most information in strings.

\marginpar{\Large\vspace{1em}\colorbox{yellow}{\textbf{Cmds}}}

After encoding, labels and properties are concatenated into \hl{input} feature
vectors for each vertex \hl{and for each edge}. Importantly, the concatenation
is done after ordering the elements of a set $\text{Labels} \cup \text{Property
keys}$ (i.e., $L \cup K$) and applying the same ordering for each vertex and
for each edge.  This ensures that the embeddings of labels/properties follow
  the same order in each feature vector and that the lengths of feature vectors
  for, respectively, vertices and edges, are the same.

\subsection{\colorbox{yellow}{Seamless Integration with GNN Models and Encodings}}

\marginpar{\Large\vspace{-1em}\colorbox{yellow}{\textbf{Cmds}}}

\marginpar{\Large\vspace{1em}\colorbox{yellow}{\textbf{Cmds}}}

\hl{Both vertex and edge information is straightforwardly harnessed by LPG2vec by first
encoding the input vertex or edge labels/properties within LPG2vec. Then, we
feed such vertex and edge encodings, as input feature vectors $\mathbf{x}_i$
(for any vertex $i$) and $\mathbf{x}_{ij}$ (for any edge $(i,j)$), into a
selected GNN model.
Here, LPG2vec enables seamless integration with virtually any GNN model, encoding, or architecture. 
This is due to the simplicity of our solution: all LPG2vec does is providing ``enriched''
input feature vectors $\mathbf{x}_i$ and $\mathbf{x}_{ij}$. Vectors $\mathbf{x}_i$ can be directly fed to any convolutional,
attentional, or message-passing GNN model, as the input vertex feature vectors. Vectors $\mathbf{x}_{ij}$ are fed into
any model that also incorporates edge feature vectors.} 

\marginpar{\Large\vspace{1em}\colorbox{yellow}{\textbf{Cmds}}}

\hl{Moreover,
LPG2vec also enables easy integration with encoding schemes such as MPGNNs-LSPE~\mbox{\cite{dwivedi2021graph}}.
This can be achieved by, for example, concatenating the LPG2vec vectors with any additional encodings,
and then using the resulting feature vectors with the selected GNN model.}

\vspaceSQ{-0.5em}
\section{Evaluation}
\label{sec:eval}
\vspaceSQ{-0.25em}

Our main goal in the evaluation is to show that LPG2vec 
successfully harnesses the label and property information from the LPG graph
datasets to offer more accurate predictions in graph ML tasks.
Our analysis comes with a large evaluation space. Thus, we show selected representative results; full data is in the appendix due to space constraints.

\vspaceSQ{-0.5em}
\subsection{Experimental Setup}
\vspaceSQ{-0.25em}

\marginpar{\Large\vspace{5em}\colorbox{yellow}{\textbf{n77k}}}

An important part of the experimental setup is finding the \textbf{appropriate
graph datasets} that \emph{have many labels and properties}. 
First, we use the Microsoft Academic Knowledge Graph
(MAKG)~\cite{DBLP:conf/semweb/Farber19}.  The original graph is in the RDF
format. We extracted data from RDF triples describing consecutive vertices, and
we built LPG vertex entries containing the gathered data; single triples
containing edges were parsed directly into the LPG format. Due to the huge size
of MAKG, we extracted two subgraphs. For this, we consider the following \hl{LPG labels
of vertices}: \emph{:Paper, :Author, :Affiliation,
:ConferenceSeries, :ConferenceInstance, :FieldOfStudy}, as well as the links
between them. Then, we additionally limit the number of the considered research
areas (and thus vertices) in the \emph{:FieldOfStudy} field (four for a small
MAKG dataset, 25 for a large MAKG dataset); \hl{they form classes to be predicted}. For diversified analysis, we make
sure that these two datasets differ in their degree distributions, implying
different connectivity structure.
Second, we use example LPG graphs provided by Neo4j\footnote{Available at
\url{https://github.com/neo4j-graph-examples}}; While these datasets are small,
they are original excerpts from industry LPG databases. Most importantly, we use a
``citations'' network (modeling publications and citations between them), a
``Twitter trolls'' network (modeling anonymized Twitter trolls and the
interaction of retweets), and a network modeling crime investigations.
The details of datasets are in Table~\ref{tab:datasets}; the appendix
provides a full specification of the associated labels/properties in
selected datasets, as well as additional results.

While there are many heterogeneous graphs available online, they have usually
single labels (often called types) per vertex or edge.
We considered some of these graphs; we first convert them appropriately
into the LPG model by transforming certain information from the graph structure
into labels and properties.
Note that we do \emph{not} compete with heterogeneous representations,
datasets, and the associated heterogeneous GNN models (they are outside the scope of this work); instead, we focus on LPG
because this is the main established graph data model in graph databases.


\begin{table}[h]
\vspaceSQ{-0.75em}
\centering
\setlength{\tabcolsep}{2pt}
\renewcommand{\arraystretch}{0.5}
\scriptsize
%
\begin{tabular}{@{}lllllll@{}}
\toprule
\makecell[l]{\textbf{Dataset}} &
\makecell[l]{\#vertices} &
\makecell[l]{\#edges} &
\makecell[l]{\#labels} &
\makecell[l]{\#properties} &
\makecell[l]{size} &
\makecell[c]{\textbf{Prediction target \& ML task details}} \\
%
\midrule
\makecell[l]{\textbf{[MAKG]} (small)} & \makecell[l]{3.06M} & \makecell[l]{12.3M} & \makecell[l]{20} & \makecell[l]{28} & 1.2 GB & Publication area (node classification, 4 classes) \\ 
\makecell[l]{\textbf{[MAKG]} (large)} & \makecell[l]{50.7M} & \makecell[l]{190M} & \makecell[l]{20} & \makecell[l]{28} & 19.5 GB & Publication area (node classification, 25 classes) \\
\makecell[l]{\textbf{[Neo4j]} citations} & \makecell[l]{132k} & \makecell[l]{221k} & \makecell[l]{5} & \makecell[l]{6} & 51 MB & Citation count (node regression) \\ 
\makecell[l]{\textbf{[Neo4j]} Twitter trolls} & \makecell[l]{281k} & \makecell[l]{493k} & \makecell[l]{13} & \makecell[l]{14} & 79 MB & Retweet count (node regression) \\ 
\makecell[l]{\textbf{[Neo4j]} crime investigations} & \makecell[l]{61.5k} & \makecell[l]{106k} & \makecell[l]{28} & \makecell[l]{29} & 17 MB & Crime type (node classification) \\ 
%
%
\bottomrule
\end{tabular}
%
%
\caption{\textbf{Considered LPG datasets \& ML tasks}.
\textbf{[Neo4j]}: provided by the Neo4j online repository,
\textbf{[MAKG]}: extracted from MS Academic Knowledge Graph.
%
Additional results for all the datasets are provided in the appendix.
}
\label{tab:datasets}
\vspaceSQ{-1.5em}
\end{table}


We consider different established GNN models: GCN~\cite{kipf2016semi} (a
seminal convolutional GNN model), GAT~\cite{velivckovic2017graph} (a seminal
attentional GNN model), and GIN~\cite{xu2018powerful} (a seminal model having
more expressive power than GCN or GAT).
We test these models with and without the LPG2vec encoding scheme. Then, when
considering models enhanced with LPG2vec, we test variants that harness the
additional LPG information coming from only labels, only properties, and from
both labels and properties. Our goal is to investigate how exactly the rich
additional LPG information influences the accuracy of the established graph ML tasks,
focusing on
node classification (assigning each vertex to one of a given number of
classes) and node regression (predicting a real value for each vertex)~\cite{wu2020comprehensive}.

\marginpar{\Large\vspace{3em}\colorbox{yellow}{\textbf{n77k}}}

We split the datasets into train, val, and test by the ratio of $[0.8, 0.1,
0.1]$.  We set the mini-batch size to 32, use the Adam
optimizer~\cite{kingma2014adam}, the learning rate of 0.01 augmented with the
cosine annealing decay, and we train for 100 epochs.  The node mini-batch
sampling is conducted using the GraphSAINT established
scheme~\cite{zeng2019graphsaint}.
%
%
\hl{We use the cross-entropy and MSE loss functions for classification
and regression, respectively.}
In the design of used GNN models (GCN, GAT, GIN), following the established
practice~\cite{you2020design}, we incorporate one preprocessing MLP layer,
followed by two actual GNN layers, and then one additional
post-processing MLP layer. We use the PReLU non-linearity.
%

Our implementation is integrated into PyG~\cite{fey2019fast}. We use
GraphGym~\cite{you2020design} as well as Weights \& Biases~\cite{wandb} for
managing experiments.

\vspaceSQ{-0.5em}
\subsection{Improving the Accuracy with LPG Labels and Properties}
\vspaceSQ{-0.25em}

\enlargeSQ

\marginpar{\Large\vspace{1em}\colorbox{yellow}{\textbf{bJap}}}

\begin{figure}
\vspaceSQ{-1em}
    \centering
    \includegraphics[width=1.0\textwidth]{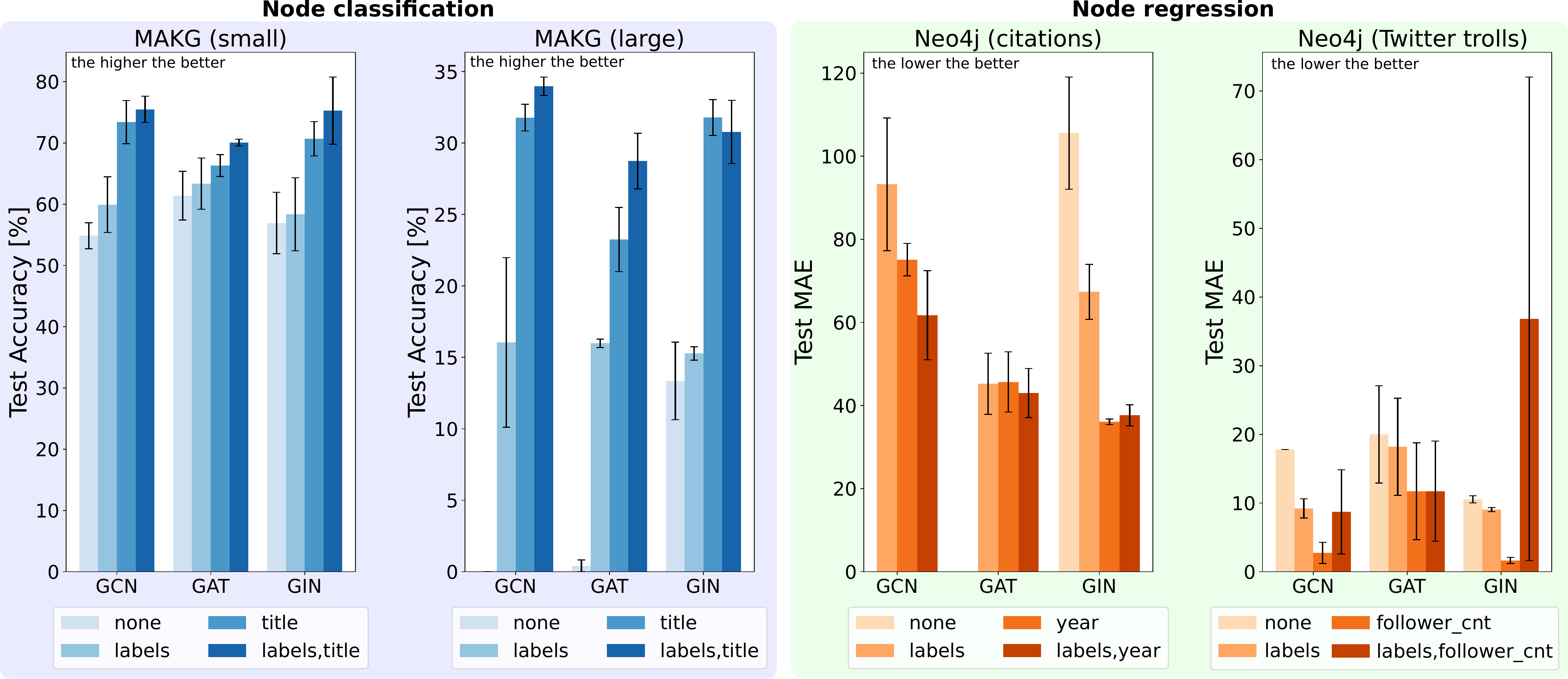}
    \caption{Advantages of preserving the information encoded in LPG labels and properties, for node classification (the MAKG datasets in the left panel; 4 classes for small and 25 classes for large)
    and node regression (the Neo4j datasets in the right panel).}
    \label{fig:node-class}
\end{figure}

We first analyze how LPG2vec \hl{appropriately harnesses the rich information from LPG labels/properties,
enabling accuracy improvements for different GNN models}.
Example results are in Figure~\ref{fig:node-class}, showing both node classification
and node regression, with MAKG and Neo4j datasets. We plot the 
the final test accuracy (with the standard
deviation) for classification and the mean absolute error (MAE) for node regression.
The task is to predict the research area of the publication (for MAKG) and
the citation count of a paper (for Neo4j citations).
In the results, the baseline with no LPG labels/properties (i.e., only the
neighborhood structure) consistently delivers the lowest accuracy (MAKG), or --
in some cases such as for the GCN/GAT models and Neo4j citations dataset -- is unable to converge. Then, for MAKG, including, 
  respectively, labels (describing \emph{paper types}), a property (\emph{paper title}),
  and both the labels and the title property, steadily improves the accuracy,
  reaching nearly 35\% for GCN. \hl{The trend is similar across all the studied
  models, and they achieve similarly high accuracy, which indicates that harnessing the
  appropriate labels/properties is very relevant and - when this information is present -
  different GNN models will perform similarly well}. Neo4j citations and Twitter trolls are similar (note the
  different metric as this is a node regression task). The main difference is that,
  for GIN and the Twitter dataset, combining the {labels} and the \emph{follower count} leads to worse
  results than only using one of these two individually. This illustrates that certain
  {combinations} of LPG information might not always enhance the accuracy;
  we study this in more detail in Sec.~\ref{sec:detrimen}.
\hl{Another interesting effect takes place when considering the bare graph
structure on the small MAKG. Here, GCN performs worst, GAT is somewhat better,
while GIN delivers much higher accuracy than GAT. We conjecture this is because 
 GIN is provably highly expressive in the Weisfeiler-Lehman
  sense (when considering the bare graph structure)~\mbox{\cite{xu2018powerful}}.}
Overall, the results show the importance of including both the labels and
properties when analyzing LPG graphs.

\marginpar{\Large\vspace{-10em}\colorbox{yellow}{\textbf{n77k}}}

\marginpar{\Large\vspace{-3em}\colorbox{yellow}{\textbf{n77k}}}

\marginpar{\Large\vspace{1em}\colorbox{yellow}{\textbf{bJap}}}

\vspaceSQ{-0.5em}
\subsection{\colorbox{yellow}{Selecting the Right Labels and Properties}}
\label{sec:detrimen}
\vspaceSQ{-0.25em}

\begin{figure}
\vspace{-1.5em}
    \centering
    \includegraphics[width=0.93\textwidth]{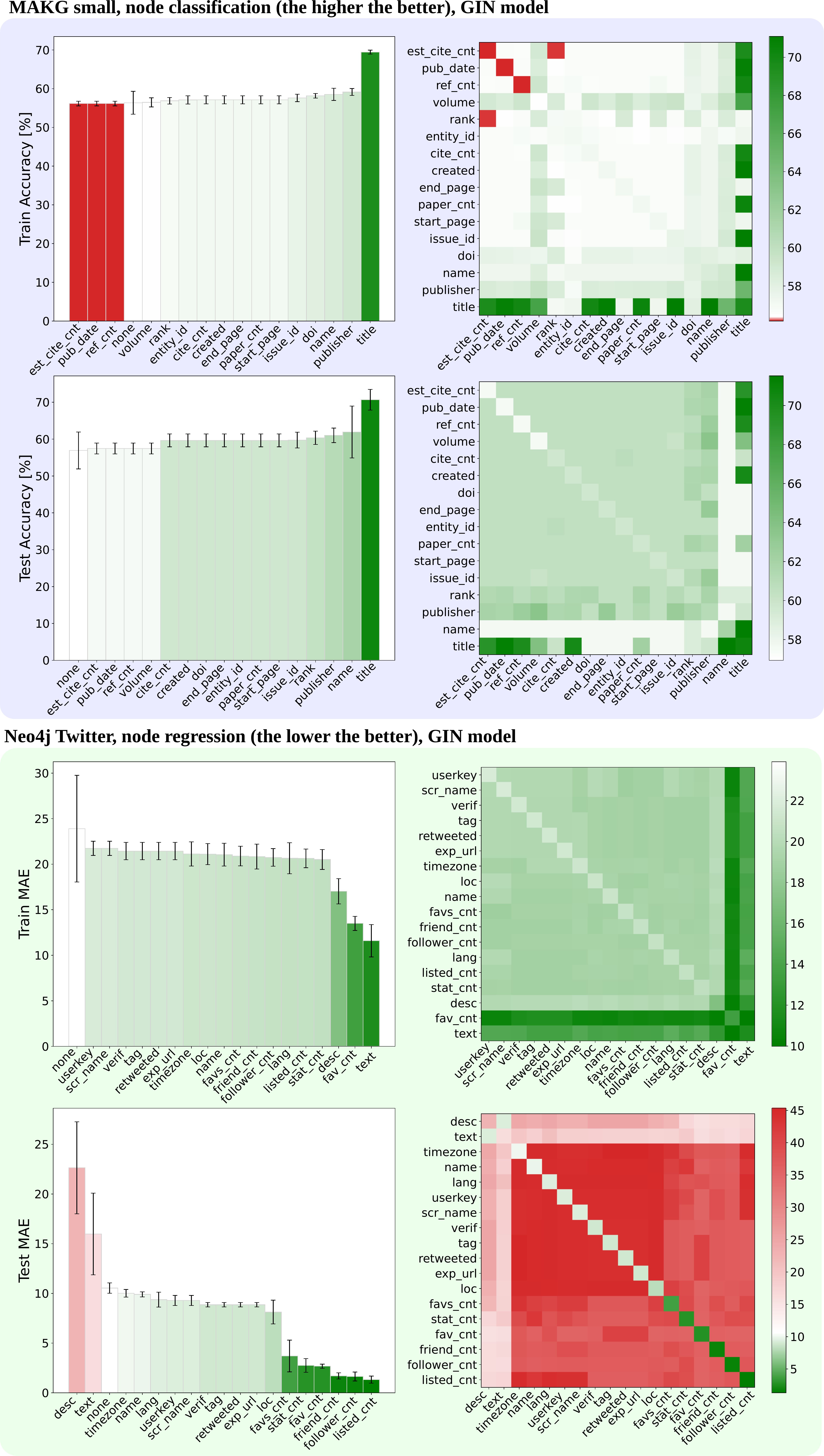}
    \vspaceSQ{-0.25em}
    \caption{Analysis of the impact from different properties on the accuracy,
    considering both individual properties and the combinations of property
    pairs, for the MAKG and Neo4j, with the node classification and regression tasks, for the GIN model.
    The green color indicates that a given result
    is better than for a graph with no labels/properties.
The red color indicates the results are worse than for a graph with
    no labels/properties.}
    \label{fig:makg-small-dets}
\end{figure}

In some experiments, we observed that selecting certain properties was not
improving the accuracy. Moreover, in certain cases, the accuracy was actually
{diminishing}. We analyze this effect in more detail in
Figure~\ref{fig:makg-small-dets} for the node classification and regression on MAKG
small and Neo4j Twitter trolls, with the GIN model, plotting both train and test accuracy.
The plots show the impact of using each of the many available properties
on the final prediction accuracy. For example, on the small MAKG, using the
\emph{title} property significantly improves the accuracy, and the majority of other
properties also increase it, although by much smaller (often negligible)
factor. Still, using the \emph{publication date} property in many cases decreases the
accuracy (see the bottom-left plot).
We further analyze this effect with heatmaps, by considering each possible
\emph{pair of properties}, and how using this pair impacts the results.
The accuracy is almost always enhanced, when using \emph{title}
together with nearly any other property. Some properties, such
as \emph{entity id}, have no effect. Many pair combinations result in
slight accuracy improvements.
However, in the Neo4j Twitter case (the bottom panel), in the test accuracy,
while using many individual properties significantly enhances the accuracy,
  most combinations of property pairs decrease it. Interestingly, this only happens
  for the GIN model; \emph{the GCN models and GAT models are able to extract useful 
    knowledge from most property pairs} (these results are provided in the appendix, see
    Figures~\ref{fig:tweeter-hm-gcn} and~\ref{fig:tweeter-hm-gat}).
This illustrates that it is important to understand the data and select the right encoded LPG
information \emph{and} the model for a given selected graph ML task.

\section{Related Work and Discussion}

\marginpar{\Large\vspace{2em}\colorbox{yellow}{\textbf{Cmds}}}

Our work touches on many areas. We now briefly discuss
related works.
\hl{ We do not compare LPG2vec to non-GNN baselines because our main goal is to
illustrate how to integrate GNN capabilities into GDBs, and \emph{not} to
argue that neural methods outperform those of traditional non-GNN baselines.
Hence, we do not focus on experiments with traditional GDB 
non-neural tasks such as BFS or Connected Components.}

\textbf{Graph Neural Networks and Graph Machine Learning}
Graph neural networks (GNNs) emerged as a highly successful part of the graph
machine learning field~\cite{hamilton2017representation}. Numerous GNN models
have been developed~\cite{wu2020comprehensive, zhou2020graph, zhang2020deep,
chami2020machine, hamilton2017representation, bronstein2017geometric,
besta2021motif, gianinazzi2021learning, scarselli2008graph, besta2022parallel},
including convolutional~\cite{kipf2016semi, hamilton2017inductive,
wu2019simplifying, xu2018powerful, sukhbaatar2016learning},
attentional~\cite{monti2017geometric, velivckovic2017graph,
thekumparampil2018attention}, message-passing~\cite{wang2019dynamic,
bresson2017residual, sanchez2020learning, battaglia2018relational,
gilmer2017neural}, or -- more recently -- higher-order
ones~\cite{abu-el-haijaMixHopHigherOrderGraph2019,
rossiHigherorderNetworkRepresentation2018, rossiHONEHigherOrderNetwork2018,
benson2018simplicial, abu2019mixhop, morris2019weisfeiler,
bodnar2021weisfeiler}.
%
%
Moreover, a large number of software frameworks~\cite{wang2019deep,
fey2019fast, li2020pytorch, zhang2020agl, jia2020improving, zhu2019aligraph,
wang2020gnnadvisor, hu2020featgraph, wu2021seastar, ma2019neugraph,
tripathy2020reducing, wan2022pipegcn, wan2022bns, zheng2021distributed,
waleffe2022marius++}, and even hardware accelerators~\cite{kiningham2020greta,
liang2020engn, yan2020hygcn, geng2020awb, kiningham2020grip} for processing
GNNs have been introduced over the last years.
LPG2vec enables using all these designs together with the LPG graphs and
consequently with LPG-based graph databases. This is because of the fact that
the information within LPG labels and properties is encoded into the input
features vectors, which can then be seamlessly used with essentially any
GNN model or framework of choice.
%
%

\textbf{Graph Databases (GDBs)}~\cite{besta2019demystifying} are systems used to manage,
process, analyze, and store vast amounts of rich and complex graph datasets.
GDBs have a long history of development and focus in both academia
and in the industry, and there has been significant work on
them~\cite{angles2018introduction, davoudian2018survey, han2011survey,
gajendran2012survey, gdb_survey_paper_Kaliyar, kumar2015domain,
gdb_survey_paper_Angles}. A lot of research has been dedicated to graph query
languages~\cite{gdb_query_language_Angles, bonifati2018querying,
gdb_query_language_Angles}, GDB 
management~\cite{gdb_management_huge_unstr_data, pokorny2015graph,
junghanns2017management, bonifati2018querying, miller2013graph}, compression in
GDBs and data models~\cite{lyu2016scalable, ma2016big,
nabti2017querying, besta2018survey, besta2019slim, besta2018log,
besta2022probgraph}, execution in novel environments such as the serverless
setting~\cite{toader2019graphless, copik2020sebs, mao2022ermer}, and others.
Many GDBs exist~\cite{tigergraph2022ldbc, cge_paper,
tiger_graph_links, janus_graph_links, azure_cosmosdb_links,
amazon_neptune_links, virtuoso_links, arangodb_links, arangodb_indexing_links,
arangodb_starter, orientdb_lwedge_links, tesoriero2013getting,
profium_sense_links, triplebit_links, gbase_paper, graphbase_links, graphflow,
livegraph, memgraph_links, dubey2016weaver, sparksee_paper, graphdb_links,
redisgraph_links, dgraph_links, allegro_graph_links, apache_jena_tbd_links,
mormotta_links, brightstardb_links, gstore, anzo_graph_links, datastax_links,
infinite_graph_links, blaze_graph_links, oracle_spatial, stardog_links,
cayley_links, weaver_links}.
We enhance the learning capabilities of graph databases by illustrating how to
harness all the information encoded in Labeled Property Graph (LPG), a data
model underlying the majority of graph databases, and use it for graph ML tasks
such as node classification.

\iftr
\textbf{Heterogeneous Graphs}
In heterogeneous graphs, each vertex and each edge may belong to a different
type~\cite{wang2020survey, yang2020heterogeneous, xie2021survey}.  There exist
GNN models targeting such graphs, for example HAN~\cite{wang2019heterogeneous},
HGT~\cite{hu2020heterogeneous}, HGNN~\cite{zhang2019heterogeneous},
MAGNN~\cite{fu2020magnn}, GATNE~\cite{cen2019representation}, or
Simple-HGN~\cite{lv2021we}.
Contrarily to LPG, heterogeneous graphs only enable attaching a single type 
to vertices and/or edges, and they do not enable arbitrary key-values pairs
being attached with vertices and edges. In this work, we focus on LPG graphs
as the primary data model in graph databases, and illustrate how to enhance
their capabilities with GNNs.

\textbf{Resource Description Framework and Knowledge Graphs}
Resource Description Framework (RDF)~\cite{lassila1998resource} is a standard
originally developed as a data model for metadata. It consists of triples,
i.e., 3-tuples used to encode any information in a graph. Hence, while it has
been used to encode knowledge in ontological models and in frameworks called
RDF stores~\cite{modoni2014survey, harris20094store, papailiou2012h2rdf}, it is
less common in graph databases that focus on achieving high performance, low
latency, and large scale. The reason is that LPG facilitates explicit storage
of graph structure, and thus makes it easier to achieve high performance of
different graph algorithms and complex business-intelligence graph queries that
commonly require accessing graph neighborhoods~\cite{besta2019demystifying}.
We focus on graph databases built on top of LPG, and thus RDF and the associated
graph ML models such as RDF2vec~\cite{ristoski2016rdf2vec, portisch2020rdf2vec,
jurisch2018rdf2vec, ristoski2019rdf2vec} are outside the scope of this work.
Note that the notions of label and property prediction are analogous to the
concepts of knowledge graph completion~\cite{lin2015learning, arora2020survey,
wang2017knowledge, dai2020survey, urbani2016kognac, shi2017proje,
trouillon2017knowledge, akrami2020realistic, yao2019kg, sun2019re,
chen2020knowledge, lin2015learning, shi2018open}.
\fi

\iftr
\textbf{Dynamic, Temporal, and Streaming Graph Processing Frameworks}
There also exist systems for processing dynamic, temporal, and streaming
graphs~\cite{besta2019practice, sakr2020future, choudhury2017nous, besta2021enabling}.
Their setting partially overlaps with graph databases, because they also focus
on high-performance graph processing and on solving 
graph problems such as
Betweenness Centrality~\cite{solomonik2017scaling, madduri2009faster}, Graph
Traversals~\cite{Besta:2015:AIC:2749246.2749263, besta2017slimsell,
kepner2016mathematical, bulucc2011combinatorial}, Connected
Components~\cite{gianinazzi2018communication}, Graph
Coloring~\cite{besta2020highcolor}, Matchings~\cite{besta2020substream}, and
many others~\cite{thorup2000near, lee2016efficient, henzinger1999randomized,
demetrescu2009dynamic, eppstein1999dynamic, gianinazzi2018communication,
gianinazzi2021parallel, ivkovic1993fully, besta2021sisa, gms, besta2019communication, besta2017push}.
However, the rate of updates in such systems is much higher than in graph
databases, thus requiring usually significantly different system designs and
architectures. More importantly, such frameworks do not focus on rich data and
do not use the LPG model. Hence, these systems differ fundamentally from graph
  databases, and are outside the focus of this paper.
\fi

\section{Conclusion}

Graph databases (GDBs), despite being an important part of the graph analytics
landscape, have still not embraced the full predictive capabilities of graph
neural networks (GNNs).
To address this, we first observe that the majority of graph databases use, 
or support, the Labeled Property Graph (LPG) as their data model. In LPG, the graph
structure, stored explicitly in the compressed-sparse row format, is combined
with labels and key-value properties that can be attached, in any
configuration, to vertices and edges.
To integrate GDBs with graph machine learning capabilities, we develop LPG2vec,
an encoder that converts LPG labels and properties into input vertex and edge
embeddings. This enables seamless integration of any GDB with any GNN model of
interest.

Our evaluation shows that incorporating labels and properties into GNN models
consistently improves accuracy. For example, GCN, GAT, and GIN models achieve
even up to 34\% better accuracy in node classification for the LPG
representation of the Microsoft Academic Knowledge Graph, compared to a setting
without LPG labels and properties. We conclude that LPG2vec will facilitate the
development of neural graph databases, a learning architecture that harnesses
both the structure and rich data (labels, properties) of LPG for highly
accurate predictions in graph databases.  It will lead to the wider adoption of GNNs
in the broad graph database industry setting.

\iftr
\vspace{2em}


\textbf{Acknowledgements}
We thank Hussein Harake, Colin McMurtrie, Mark Klein, Angelo Mangili, and the
whole CSCS team granting access to the Ault and Daint machines, and for their
excellent technical support. We thank Timo Schneider for immense help with
computing infrastructure at SPCL.
This research received funding from the European Research Council
\raisebox{-0.25em}{\includegraphics[height=1em]{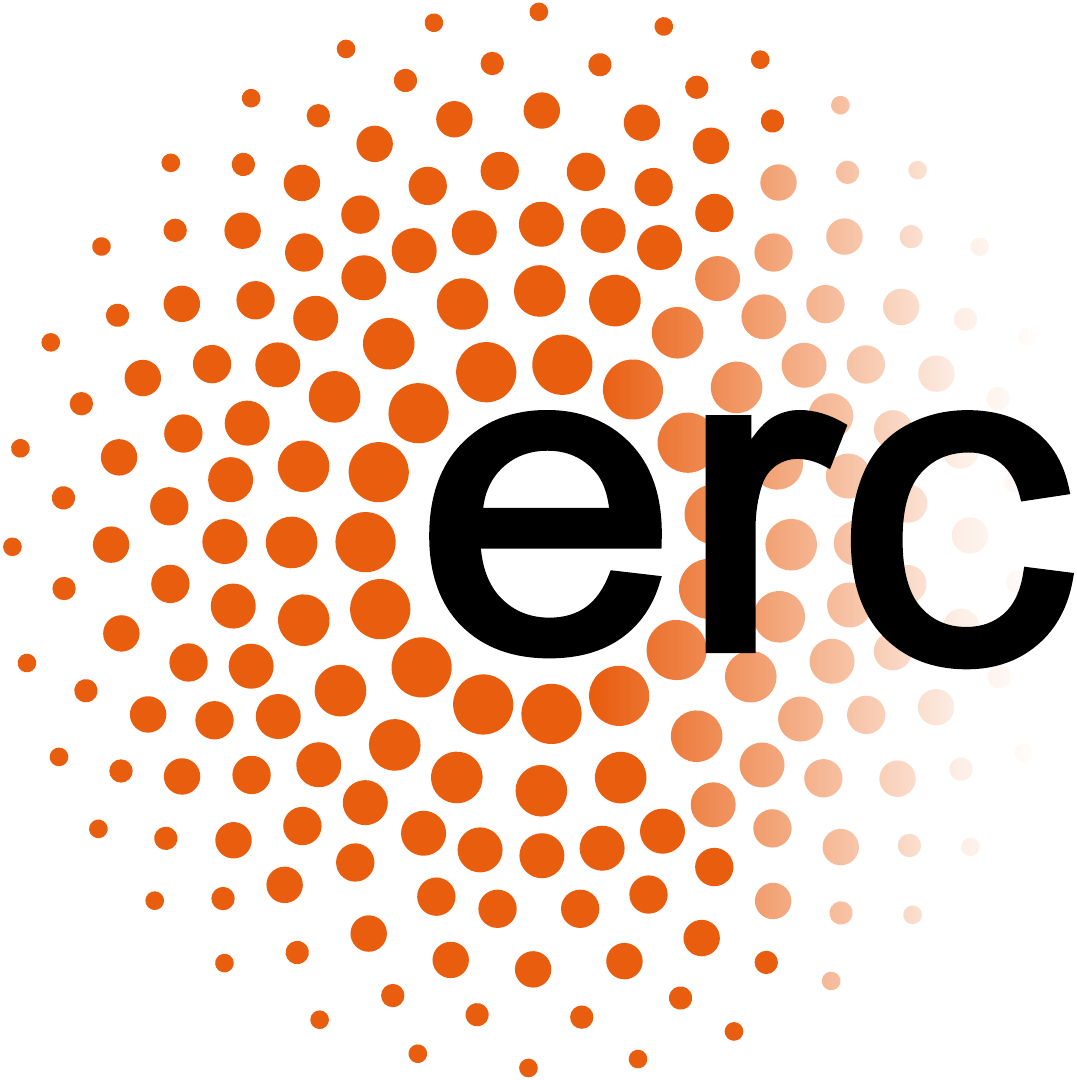}} (Project PSAP,
No.~101002047), Huawei, MS Research through its Swiss Joint Research Centre, and the European High-Performance Computing Joint Undertaking (JU) under grant agreement No.~955513 (MAELSTROM).


\fi

{
  \footnotesize
\bibliographystyle{ACM-Reference-Format}
\bibliography{references}


\newcommand{\SortNoop}[1]{}
\begin{thebibliography}{194}


\ifx \showCODEN    \undefined \def \showCODEN     #1{\unskip}     \fi
\ifx \showDOI      \undefined \def \showDOI       #1{#1}\fi
\ifx \showISBNx    \undefined \def \showISBNx     #1{\unskip}     \fi
\ifx \showISBNxiii \undefined \def \showISBNxiii  #1{\unskip}     \fi
\ifx \showISSN     \undefined \def \showISSN      #1{\unskip}     \fi
\ifx \showLCCN     \undefined \def \showLCCN      #1{\unskip}     \fi
\ifx \shownote     \undefined \def \shownote      #1{#1}          \fi
\ifx \showarticletitle \undefined \def \showarticletitle #1{#1}   \fi
\ifx \showURL      \undefined \def \showURL       {\relax}        \fi
\providecommand\bibfield[2]{#2}
\providecommand\bibinfo[2]{#2}
\providecommand\natexlab[1]{#1}
\providecommand\showeprint[2][]{arXiv:#2}

\bibitem[\protect\citeauthoryear{??}{tes}{2013}]%
        {tesoriero2013getting}
 \bibinfo{year}{2013}\natexlab{}.
\newblock \bibinfo{booktitle}{\emph{{Getting Started with OrientDB}}}.
\newblock \bibinfo{publisher}{Packt Publishing Ltd}.
\newblock


\bibitem[\protect\citeauthoryear{??}{wea}{2018}]%
        {weaver_links}
 \bibinfo{year}{2018}\natexlab{}.
\newblock \bibinfo{title}{{Weaver}}.
\newblock \bibinfo{howpublished}{Available at \url{http://weaver.systems/}}.
\newblock


\bibitem[\protect\citeauthoryear{??}{tig}{2022}]%
        {tigergraph2022ldbc}
 \bibinfo{year}{2022}\natexlab{}.
\newblock \bibinfo{booktitle}{\emph{Using the Linked Data Benchmark Council
  Social Network Benchmark Methodology to Evaluate TigerGraph at 36
  Terabytes}}.
\newblock \bibinfo{type}{White Paper}. \bibinfo{institution}{TigerGraph, Inc.}
\newblock


\bibitem[\protect\citeauthoryear{{Abu-El-Haija}, Perozzi, Kapoor, Alipourfard,
  Lerman, Harutyunyan, Steeg, and Galstyan}{{Abu-El-Haija}
  et~al\mbox{.}}{2019}]%
        {abu-el-haijaMixHopHigherOrderGraph2019}
\bibfield{author}{\bibinfo{person}{Sami {Abu-El-Haija}}, \bibinfo{person}{Bryan
  Perozzi}, \bibinfo{person}{Amol Kapoor}, \bibinfo{person}{Nazanin
  Alipourfard}, \bibinfo{person}{Kristina Lerman}, \bibinfo{person}{Hrayr
  Harutyunyan}, \bibinfo{person}{Greg~Ver Steeg}, {and} \bibinfo{person}{Aram
  Galstyan}.} \bibinfo{year}{2019}\natexlab{}.
\newblock \showarticletitle{{{MixHop}}: {{Higher-Order Graph Convolutional
  Architectures}} via {{Sparsified Neighborhood Mixing}}}.
\newblock \bibinfo{journal}{\emph{arXiv:1905.00067 [cs, stat]}}
  (\bibinfo{date}{June} \bibinfo{year}{2019}).
\newblock
\showeprint[arxiv]{cs, stat/1905.00067}


\bibitem[\protect\citeauthoryear{Abu-El-Haija, Perozzi, Kapoor, Alipourfard,
  Lerman, Harutyunyan, Ver~Steeg, and Galstyan}{Abu-El-Haija
  et~al\mbox{.}}{2019}]%
        {abu2019mixhop}
\bibfield{author}{\bibinfo{person}{Sami Abu-El-Haija}, \bibinfo{person}{Bryan
  Perozzi}, \bibinfo{person}{Amol Kapoor}, \bibinfo{person}{Nazanin
  Alipourfard}, \bibinfo{person}{Kristina Lerman}, \bibinfo{person}{Hrayr
  Harutyunyan}, \bibinfo{person}{Greg Ver~Steeg}, {and} \bibinfo{person}{Aram
  Galstyan}.} \bibinfo{year}{2019}\natexlab{}.
\newblock \showarticletitle{Mixhop: Higher-order graph convolutional
  architectures via sparsified neighborhood mixing}. In
  \bibinfo{booktitle}{\emph{international conference on machine learning}}.
  PMLR, \bibinfo{pages}{21--29}.
\newblock


\bibitem[\protect\citeauthoryear{Akrami, Saeef, Zhang, Hu, and Li}{Akrami
  et~al\mbox{.}}{2020}]%
        {akrami2020realistic}
\bibfield{author}{\bibinfo{person}{Farahnaz Akrami},
  \bibinfo{person}{Mohammed~Samiul Saeef}, \bibinfo{person}{Qingheng Zhang},
  \bibinfo{person}{Wei Hu}, {and} \bibinfo{person}{Chengkai Li}.}
  \bibinfo{year}{2020}\natexlab{}.
\newblock \showarticletitle{Realistic re-evaluation of knowledge graph
  completion methods: An experimental study}. In
  \bibinfo{booktitle}{\emph{Proceedings of the 2020 ACM SIGMOD International
  Conference on Management of Data}}. \bibinfo{pages}{1995--2010}.
\newblock


\bibitem[\protect\citeauthoryear{{Amazon}}{{Amazon}}{2018}]%
        {amazon_neptune_links}
\bibfield{author}{\bibinfo{person}{{Amazon}}.} \bibinfo{year}{2018}\natexlab{}.
\newblock \bibinfo{title}{{Amazon Neptune}}.
\newblock \bibinfo{howpublished}{Available at
  \url{https://aws.amazon.com/neptune/}}.
\newblock


\bibitem[\protect\citeauthoryear{Angles, Arenas, Barcel\'{o}, Hogan, Reutter,
  and Vrgo\v{c}}{Angles et~al\mbox{.}}{2017}]%
        {gdb_query_language_Angles}
\bibfield{author}{\bibinfo{person}{Renzo Angles}, \bibinfo{person}{Marcelo
  Arenas}, \bibinfo{person}{Pablo Barcel\'{o}}, \bibinfo{person}{Aidan Hogan},
  \bibinfo{person}{Juan Reutter}, {and} \bibinfo{person}{Domagoj Vrgo\v{c}}.}
  \bibinfo{year}{2017}\natexlab{}.
\newblock \showarticletitle{{Foundations of Modern Query Languages for Graph
  Databases}}.
\newblock \bibinfo{journal}{\emph{in ACM Comput. Surv.}} \bibinfo{volume}{50},
  \bibinfo{number}{5}, Article \bibinfo{articleno}{68} (\bibinfo{year}{2017}),
  \bibinfo{numpages}{40}~pages.
\newblock
\showISSN{0360-0300}
\urldef\tempurl%
\url{https://doi.org/10.1145/3104031}
\showDOI{\tempurl}


\bibitem[\protect\citeauthoryear{Angles and Gutierrez}{Angles and
  Gutierrez}{2008}]%
        {gdb_survey_paper_Angles}
\bibfield{author}{\bibinfo{person}{Renzo Angles} {and} \bibinfo{person}{Claudio
  Gutierrez}.} \bibinfo{year}{2008}\natexlab{}.
\newblock \showarticletitle{{Survey of Graph Database Models}}.
\newblock \bibinfo{journal}{\emph{in ACM Comput. Surv.}} \bibinfo{volume}{40},
  \bibinfo{number}{1}, Article \bibinfo{articleno}{1} (\bibinfo{year}{2008}),
  \bibinfo{numpages}{39}~pages.
\newblock
\showISSN{0360-0300}
\urldef\tempurl%
\url{https://doi.org/10.1145/1322432.1322433}
\showDOI{\tempurl}


\bibitem[\protect\citeauthoryear{Angles and Gutierrez}{Angles and
  Gutierrez}{2018}]%
        {angles2018introduction}
\bibfield{author}{\bibinfo{person}{Renzo Angles} {and} \bibinfo{person}{Claudio
  Gutierrez}.} \bibinfo{year}{2018}\natexlab{}.
\newblock \showarticletitle{{An Introduction to Graph Data Management}}.
\newblock In \bibinfo{booktitle}{\emph{Graph Data Management, Fundamental
  Issues and Recent Developments}}. \bibinfo{pages}{1--32}.
\newblock


\bibitem[\protect\citeauthoryear{{Apache}}{{Apache}}{2018}]%
        {mormotta_links}
\bibfield{author}{\bibinfo{person}{{Apache}}.} \bibinfo{year}{2018}\natexlab{}.
\newblock \bibinfo{title}{{Apache Mormotta}}.
\newblock \bibinfo{howpublished}{Available at
  \url{http://marmotta.apache.org/}}.
\newblock


\bibitem[\protect\citeauthoryear{{ArangoDB Inc.}}{{ArangoDB Inc.}}{2018a}]%
        {arangodb_links}
\bibfield{author}{\bibinfo{person}{{ArangoDB Inc.}}}
  \bibinfo{year}{2018}\natexlab{a}.
\newblock \bibinfo{title}{{ArangoDB}}.
\newblock \bibinfo{howpublished}{Available at
  \url{https://docs.arangodb.com/3.3/Manual/DataModeling/Concepts.html}}.
\newblock


\bibitem[\protect\citeauthoryear{{ArangoDB Inc.}}{{ArangoDB Inc.}}{2018b}]%
        {arangodb_indexing_links}
\bibfield{author}{\bibinfo{person}{{ArangoDB Inc.}}}
  \bibinfo{year}{2018}\natexlab{b}.
\newblock \bibinfo{title}{{ArangoDB: Index Free Adjacency or Hybrid Indexes for
  Graph Databases}}.
\newblock \bibinfo{howpublished}{Available at
  \url{https://www.arangodb.com/2016/04/index-free-adjacency-hybrid-indexes-graph-databases/}}.
\newblock


\bibitem[\protect\citeauthoryear{{ArangoDB Inc.}}{{ArangoDB Inc.}}{2018c}]%
        {arangodb_starter}
\bibfield{author}{\bibinfo{person}{{ArangoDB Inc.}}}
  \bibinfo{year}{2018}\natexlab{c}.
\newblock \bibinfo{title}{{ArangoDB Starter Tool}}.
\newblock \bibinfo{howpublished}{Available at
  \url{https://docs.arangodb.com/devel/Manual/Tutorials/Starter/}}.
\newblock


\bibitem[\protect\citeauthoryear{Arora}{Arora}{2020}]%
        {arora2020survey}
\bibfield{author}{\bibinfo{person}{Siddhant Arora}.}
  \bibinfo{year}{2020}\natexlab{}.
\newblock \showarticletitle{A survey on graph neural networks for knowledge
  graph completion}.
\newblock \bibinfo{journal}{\emph{arXiv preprint arXiv:2007.12374}}
  (\bibinfo{year}{2020}).
\newblock


\bibitem[\protect\citeauthoryear{Battaglia, Hamrick, Bapst, Sanchez-Gonzalez,
  Zambaldi, Malinowski, Tacchetti, Raposo, Santoro, Faulkner,
  et~al\mbox{.}}{Battaglia et~al\mbox{.}}{2018}]%
        {battaglia2018relational}
\bibfield{author}{\bibinfo{person}{Peter~W Battaglia},
  \bibinfo{person}{Jessica~B Hamrick}, \bibinfo{person}{Victor Bapst},
  \bibinfo{person}{Alvaro Sanchez-Gonzalez}, \bibinfo{person}{Vinicius
  Zambaldi}, \bibinfo{person}{Mateusz Malinowski}, \bibinfo{person}{Andrea
  Tacchetti}, \bibinfo{person}{David Raposo}, \bibinfo{person}{Adam Santoro},
  \bibinfo{person}{Ryan Faulkner}, {et~al\mbox{.}}}
  \bibinfo{year}{2018}\natexlab{}.
\newblock \showarticletitle{Relational inductive biases, deep learning, and
  graph networks}.
\newblock \bibinfo{journal}{\emph{arXiv preprint arXiv:1806.01261}}
  (\bibinfo{year}{2018}).
\newblock


\bibitem[\protect\citeauthoryear{Benson et~al\mbox{.}}{Benson
  et~al\mbox{.}}{2018}]%
        {benson2018simplicial}
\bibfield{author}{\bibinfo{person}{Austin~R Benson} {et~al\mbox{.}}}
  \bibinfo{year}{2018}\natexlab{}.
\newblock \showarticletitle{Simplicial closure and higher-order link
  prediction}.
\newblock \bibinfo{journal}{\emph{Proceedings of the National Academy of
  Sciences}} \bibinfo{volume}{115}, \bibinfo{number}{48}
  (\bibinfo{year}{2018}), \bibinfo{pages}{E11221--E11230}.
\newblock


\bibitem[\protect\citeauthoryear{Besta}{Besta}{2021}]%
        {besta2021enabling}
\bibfield{author}{\bibinfo{person}{Maciej Besta}.}
  \bibinfo{year}{2021}\natexlab{}.
\newblock \emph{\bibinfo{title}{Enabling High-Performance Large-Scale Irregular
  Computations}}.
\newblock \bibinfo{thesistype}{Ph.D. Dissertation}. \bibinfo{school}{ETH
  Zurich}.
\newblock


\bibitem[\protect\citeauthoryear{Besta et~al\mbox{.}}{Besta
  et~al\mbox{.}}{2019a}]%
        {besta2019slim}
\bibfield{author}{\bibinfo{person}{Maciej Besta} {et~al\mbox{.}}}
  \bibinfo{year}{2019}\natexlab{a}.
\newblock \showarticletitle{Slim Graph: Practical Lossy Graph Compression for
  Approximate Graph Processing, Storage, and Analytics}.
\newblock , Article \bibinfo{articleno}{35} (\bibinfo{year}{2019}),
  \bibinfo{numpages}{25}~pages.
\newblock
\showISBNx{9781450362290}
\urldef\tempurl%
\url{https://doi.org/10.1145/3295500.3356182}
\showDOI{\tempurl}


\bibitem[\protect\citeauthoryear{Besta et~al\mbox{.}}{Besta
  et~al\mbox{.}}{2020a}]%
        {besta2020highcolor}
\bibfield{author}{\bibinfo{person}{Maciej Besta} {et~al\mbox{.}}}
  \bibinfo{year}{2020}\natexlab{a}.
\newblock \showarticletitle{{High-Performance Parallel Graph Coloring with
  Strong Guarantees on Work, Depth, and Quality}}. In
  \bibinfo{booktitle}{\emph{ACM/IEEE SC}} (Atlanta, Georgia). Article
  \bibinfo{articleno}{99}, \bibinfo{numpages}{17}~pages.
\newblock
\showISBNx{9781728199986}


\bibitem[\protect\citeauthoryear{Besta et~al\mbox{.}}{Besta
  et~al\mbox{.}}{2021a}]%
        {gms}
\bibfield{author}{\bibinfo{person}{Maciej Besta} {et~al\mbox{.}}}
  \bibinfo{year}{2021}\natexlab{a}.
\newblock \showarticletitle{GraphMineSuite: Enabling High-Performance and
  Programmable Graph Mining Algorithms with Set Algebra}.
\newblock \bibinfo{journal}{\emph{VLDB}}.
\newblock


\bibitem[\protect\citeauthoryear{Besta et~al\mbox{.}}{Besta
  et~al\mbox{.}}{2021b}]%
        {besta2021sisa}
\bibfield{author}{\bibinfo{person}{Maciej Besta} {et~al\mbox{.}}}
  \bibinfo{year}{2021}\natexlab{b}.
\newblock \showarticletitle{SISA: Set-Centric Instruction Set Architecture for
  Graph Mining on Processing-in-Memory Systems}.
\newblock \bibinfo{journal}{\emph{arXiv preprint arXiv:2104.07582}}
  (\bibinfo{year}{2021}).
\newblock


\bibitem[\protect\citeauthoryear{Besta et~al\mbox{.}}{Besta
  et~al\mbox{.}}{2022a}]%
        {besta2019practice}
\bibfield{author}{\bibinfo{person}{Maciej Besta} {et~al\mbox{.}}}
  \bibinfo{year}{2022}\natexlab{a}.
\newblock \showarticletitle{Practice of Streaming Processing of Dynamic Graphs:
  Concepts, Models, and Systems}.
\newblock \bibinfo{journal}{\emph{IEEE TPDS}} (\bibinfo{year}{2022}).
\newblock


\bibitem[\protect\citeauthoryear{Besta, Fischer, Ben-Nun, Stanojevic, Licht,
  and Hoefler}{Besta et~al\mbox{.}}{2020b}]%
        {besta2020substream}
\bibfield{author}{\bibinfo{person}{Maciej Besta}, \bibinfo{person}{Marc
  Fischer}, \bibinfo{person}{Tal Ben-Nun}, \bibinfo{person}{Dimitri
  Stanojevic}, \bibinfo{person}{Johannes De~Fine Licht}, {and}
  \bibinfo{person}{Torsten Hoefler}.} \bibinfo{year}{2020}\natexlab{b}.
\newblock \showarticletitle{Substream-Centric Maximum Matchings on FPGA}.
\newblock \bibinfo{journal}{\emph{ACM TRETS}} \bibinfo{volume}{13},
  \bibinfo{number}{2} (\bibinfo{year}{2020}), \bibinfo{pages}{1--33}.
\newblock


\bibitem[\protect\citeauthoryear{Besta, Grob, Miglioli, Bernold, Kwasniewski,
  Gjini, Kanakagiri, Ashkboos, Gianinazzi, Dryden, et~al\mbox{.}}{Besta
  et~al\mbox{.}}{2022b}]%
        {besta2021motif}
\bibfield{author}{\bibinfo{person}{Maciej Besta}, \bibinfo{person}{Raphael
  Grob}, \bibinfo{person}{Cesare Miglioli}, \bibinfo{person}{Nicola Bernold},
  \bibinfo{person}{Grzegorz Kwasniewski}, \bibinfo{person}{Gabriel Gjini},
  \bibinfo{person}{Raghavendra Kanakagiri}, \bibinfo{person}{Saleh Ashkboos},
  \bibinfo{person}{Lukas Gianinazzi}, \bibinfo{person}{Nikoli Dryden},
  {et~al\mbox{.}}} \bibinfo{year}{2022}\natexlab{b}.
\newblock \showarticletitle{Motif Prediction with Graph Neural Networks}, In
  \bibinfo{booktitle}{ACM KDD}.
\newblock \bibinfo{journal}{\emph{arXiv preprint arXiv:2106.00761}}.
\newblock


\bibitem[\protect\citeauthoryear{Besta and Hoefler}{Besta and Hoefler}{2015}]%
        {Besta:2015:AIC:2749246.2749263}
\bibfield{author}{\bibinfo{person}{Maciej Besta} {and} \bibinfo{person}{Torsten
  Hoefler}.} \bibinfo{year}{2015}\natexlab{}.
\newblock \showarticletitle{{Accelerating Irregular Computations with Hardware
  Transactional Memory and Active Messages}}. In
  \bibinfo{booktitle}{\emph{Proc. of the Intl. Symp. on High-Perf. Par. and
  Dist. Comp.}} \emph{(\bibinfo{series}{HPDC '15})}. \bibinfo{pages}{161--172}.
\newblock
\showISBNx{978-1-4503-3550-8}


\bibitem[\protect\citeauthoryear{Besta and Hoefler}{Besta and Hoefler}{2018}]%
        {besta2018survey}
\bibfield{author}{\bibinfo{person}{Maciej Besta} {and} \bibinfo{person}{Torsten
  Hoefler}.} \bibinfo{year}{2018}\natexlab{}.
\newblock \showarticletitle{Survey and taxonomy of lossless graph compression
  and space-efficient graph representations}.
\newblock \bibinfo{journal}{\emph{arXiv preprint arXiv:1806.01799}}
  (\bibinfo{year}{2018}).
\newblock


\bibitem[\protect\citeauthoryear{Besta and Hoefler}{Besta and Hoefler}{2022}]%
        {besta2022parallel}
\bibfield{author}{\bibinfo{person}{Maciej Besta} {and} \bibinfo{person}{Torsten
  Hoefler}.} \bibinfo{year}{2022}\natexlab{}.
\newblock \showarticletitle{Parallel and Distributed Graph Neural Networks: An
  In-Depth Concurrency Analysis}.
\newblock \bibinfo{journal}{\emph{arXiv preprint arXiv:2205.09702}}
  (\bibinfo{year}{2022}).
\newblock


\bibitem[\protect\citeauthoryear{Besta, Kanakagiri, Mustafa, Karasikov,
  R{\"a}tsch, Hoefler, and Solomonik}{Besta et~al\mbox{.}}{2019b}]%
        {besta2019communication}
\bibfield{author}{\bibinfo{person}{Maciej Besta}, \bibinfo{person}{Raghavendra
  Kanakagiri}, \bibinfo{person}{Harun Mustafa}, \bibinfo{person}{Mikhail
  Karasikov}, \bibinfo{person}{Gunnar R{\"a}tsch}, \bibinfo{person}{Torsten
  Hoefler}, {and} \bibinfo{person}{Edgar Solomonik}.}
  \bibinfo{year}{2019}\natexlab{b}.
\newblock \showarticletitle{Communication-Efficient Jaccard Similarity for
  High-Performance Distributed Genome Comparisons}.
\newblock \bibinfo{journal}{\emph{arXiv preprint arXiv:1911.04200}}
  (\bibinfo{year}{2019}).
\newblock


\bibitem[\protect\citeauthoryear{Besta, Marending, Solomonik, and
  Hoefler}{Besta et~al\mbox{.}}{2017a}]%
        {besta2017slimsell}
\bibfield{author}{\bibinfo{person}{Maciej Besta}, \bibinfo{person}{Florian
  Marending}, \bibinfo{person}{Edgar Solomonik}, {and} \bibinfo{person}{Torsten
  Hoefler}.} \bibinfo{year}{2017}\natexlab{a}.
\newblock \showarticletitle{Slimsell: A vectorizable graph representation for
  breadth-first search}. In \bibinfo{booktitle}{\emph{IEEE IPDPS}}. IEEE,
  \bibinfo{pages}{32--41}.
\newblock


\bibitem[\protect\citeauthoryear{Besta, Miglioli, Labini, T{\v{e}}tek, Iff,
  Kanakagiri, Ashkboos, Janda, Podstawski, Kwasniewski, et~al\mbox{.}}{Besta
  et~al\mbox{.}}{2022c}]%
        {besta2022probgraph}
\bibfield{author}{\bibinfo{person}{Maciej Besta}, \bibinfo{person}{Cesare
  Miglioli}, \bibinfo{person}{Paolo~Sylos Labini}, \bibinfo{person}{Jakub
  T{\v{e}}tek}, \bibinfo{person}{Patrick Iff}, \bibinfo{person}{Raghavendra
  Kanakagiri}, \bibinfo{person}{Saleh Ashkboos}, \bibinfo{person}{Kacper
  Janda}, \bibinfo{person}{Michal Podstawski}, \bibinfo{person}{Grzegorz
  Kwasniewski}, {et~al\mbox{.}}} \bibinfo{year}{2022}\natexlab{c}.
\newblock \showarticletitle{ProbGraph: High-Performance and High-Accuracy Graph
  Mining with Probabilistic Set Representations}.
\newblock \bibinfo{journal}{\emph{arXiv preprint arXiv:2208.11469}}
  (\bibinfo{year}{2022}).
\newblock


\bibitem[\protect\citeauthoryear{Besta, Peter, Gerstenberger, Fischer,
  Podstawski, Barthels, Alonso, and Hoefler}{Besta et~al\mbox{.}}{2019c}]%
        {besta2019demystifying}
\bibfield{author}{\bibinfo{person}{Maciej Besta}, \bibinfo{person}{Emanuel
  Peter}, \bibinfo{person}{Robert Gerstenberger}, \bibinfo{person}{Marc
  Fischer}, \bibinfo{person}{Micha{\l} Podstawski}, \bibinfo{person}{Claude
  Barthels}, \bibinfo{person}{Gustavo Alonso}, {and} \bibinfo{person}{Torsten
  Hoefler}.} \bibinfo{year}{2019}\natexlab{c}.
\newblock \showarticletitle{Demystifying Graph Databases: Analysis and Taxonomy
  of Data Organization, System Designs, and Graph Queries}.
\newblock \bibinfo{journal}{\emph{arXiv preprint arXiv:1910.09017}}
  (\bibinfo{year}{2019}).
\newblock


\bibitem[\protect\citeauthoryear{Besta, Podstawski, Groner, Solomonik, and
  Hoefler}{Besta et~al\mbox{.}}{2017b}]%
        {besta2017push}
\bibfield{author}{\bibinfo{person}{Maciej Besta}, \bibinfo{person}{Micha{\l}
  Podstawski}, \bibinfo{person}{Linus Groner}, \bibinfo{person}{Edgar
  Solomonik}, {and} \bibinfo{person}{Torsten Hoefler}.}
  \bibinfo{year}{2017}\natexlab{b}.
\newblock \showarticletitle{To push or to pull: On reducing communication and
  synchronization in graph computations}. In \bibinfo{booktitle}{\emph{ACM
  HPDC}}.
\newblock


\bibitem[\protect\citeauthoryear{Besta, Stanojevic, Zivic, Singh, Hoerold, and
  Hoefler}{Besta et~al\mbox{.}}{2018}]%
        {besta2018log}
\bibfield{author}{\bibinfo{person}{Maciej Besta}, \bibinfo{person}{Dimitri
  Stanojevic}, \bibinfo{person}{Tijana Zivic}, \bibinfo{person}{Jagpreet
  Singh}, \bibinfo{person}{Maurice Hoerold}, {and} \bibinfo{person}{Torsten
  Hoefler}.} \bibinfo{year}{2018}\natexlab{}.
\newblock \showarticletitle{Log (graph): a near-optimal high-performance graph
  representation.}. In \bibinfo{booktitle}{\emph{PACT}} (Limassol, Cyprus).
  ACM, Article \bibinfo{articleno}{7}, \bibinfo{numpages}{13}~pages.
\newblock
\showISBNx{9781450359863}
\urldef\tempurl%
\url{https://doi.org/10.1145/3243176.3243198}
\showDOI{\tempurl}


\bibitem[\protect\citeauthoryear{Bianchi, Grattarola, Livi, and Alippi}{Bianchi
  et~al\mbox{.}}{2021}]%
        {bianchi2021graph}
\bibfield{author}{\bibinfo{person}{Filippo~Maria Bianchi},
  \bibinfo{person}{Daniele Grattarola}, \bibinfo{person}{Lorenzo Livi}, {and}
  \bibinfo{person}{Cesare Alippi}.} \bibinfo{year}{2021}\natexlab{}.
\newblock \showarticletitle{Graph neural networks with convolutional arma
  filters}.
\newblock \bibinfo{journal}{\emph{IEEE TPAMI}} (\bibinfo{year}{2021}).
\newblock


\bibitem[\protect\citeauthoryear{Biewald}{Biewald}{2020}]%
        {wandb}
\bibfield{author}{\bibinfo{person}{Lukas Biewald}.}
  \bibinfo{year}{2020}\natexlab{}.
\newblock \bibinfo{title}{Experiment Tracking with Weights and Biases}.
\newblock
\newblock
\urldef\tempurl%
\url{https://www.wandb.com/}
\showURL{%
\tempurl}
\newblock
\shownote{Software available from wandb.com.}


\bibitem[\protect\citeauthoryear{{Blazegraph}}{{Blazegraph}}{2018}]%
        {blaze_graph_links}
\bibfield{author}{\bibinfo{person}{{Blazegraph}}.}
  \bibinfo{year}{2018}\natexlab{}.
\newblock \bibinfo{title}{{BlazeGraph DB}}.
\newblock \bibinfo{howpublished}{Available at
  \url{https://www.blazegraph.com/}}.
\newblock


\bibitem[\protect\citeauthoryear{Bodnar, Frasca, Wang, Otter, Montufar, Lio,
  and Bronstein}{Bodnar et~al\mbox{.}}{2021}]%
        {bodnar2021weisfeiler}
\bibfield{author}{\bibinfo{person}{Cristian Bodnar}, \bibinfo{person}{Fabrizio
  Frasca}, \bibinfo{person}{Yuguang Wang}, \bibinfo{person}{Nina Otter},
  \bibinfo{person}{Guido~F Montufar}, \bibinfo{person}{Pietro Lio}, {and}
  \bibinfo{person}{Michael Bronstein}.} \bibinfo{year}{2021}\natexlab{}.
\newblock \showarticletitle{Weisfeiler and lehman go topological: Message
  passing simplicial networks}. In \bibinfo{booktitle}{\emph{International
  Conference on Machine Learning}}. PMLR, \bibinfo{pages}{1026--1037}.
\newblock


\bibitem[\protect\citeauthoryear{Bojchevski, Klicpera, Perozzi, Kapoor, Blais,
  R{\'o}zemberczki, Lukasik, and G{\"u}nnemann}{Bojchevski
  et~al\mbox{.}}{2020}]%
        {bojchevski2020scaling}
\bibfield{author}{\bibinfo{person}{Aleksandar Bojchevski},
  \bibinfo{person}{Johannes Klicpera}, \bibinfo{person}{Bryan Perozzi},
  \bibinfo{person}{Amol Kapoor}, \bibinfo{person}{Martin Blais},
  \bibinfo{person}{Benedek R{\'o}zemberczki}, \bibinfo{person}{Michal Lukasik},
  {and} \bibinfo{person}{Stephan G{\"u}nnemann}.}
  \bibinfo{year}{2020}\natexlab{}.
\newblock \showarticletitle{Scaling graph neural networks with approximate
  pagerank}. In \bibinfo{booktitle}{\emph{ACM KDD}}.
\newblock


\bibitem[\protect\citeauthoryear{Bonifati, Fletcher, Voigt, and
  Yakovets}{Bonifati et~al\mbox{.}}{2018}]%
        {bonifati2018querying}
\bibfield{author}{\bibinfo{person}{Angela Bonifati}, \bibinfo{person}{George
  Fletcher}, \bibinfo{person}{Hannes Voigt}, {and} \bibinfo{person}{Nikolay
  Yakovets}.} \bibinfo{year}{2018}\natexlab{}.
\newblock \showarticletitle{Querying graphs}.
\newblock \bibinfo{journal}{\emph{Synthesis Lectures on Data Management}}
  \bibinfo{volume}{10}, \bibinfo{number}{3} (\bibinfo{year}{2018}),
  \bibinfo{pages}{1--184}.
\newblock


\bibitem[\protect\citeauthoryear{Bresson and Laurent}{Bresson and
  Laurent}{2017}]%
        {bresson2017residual}
\bibfield{author}{\bibinfo{person}{Xavier Bresson} {and}
  \bibinfo{person}{Thomas Laurent}.} \bibinfo{year}{2017}\natexlab{}.
\newblock \showarticletitle{Residual gated graph convnets}.
\newblock \bibinfo{journal}{\emph{arXiv preprint arXiv:1711.07553}}
  (\bibinfo{year}{2017}).
\newblock


\bibitem[\protect\citeauthoryear{Bronstein, Bruna, Cohen, and
  Veli{\v{c}}kovi{\'c}}{Bronstein et~al\mbox{.}}{2021}]%
        {bronstein2021geometric}
\bibfield{author}{\bibinfo{person}{Michael~M Bronstein}, \bibinfo{person}{Joan
  Bruna}, \bibinfo{person}{Taco Cohen}, {and} \bibinfo{person}{Petar
  Veli{\v{c}}kovi{\'c}}.} \bibinfo{year}{2021}\natexlab{}.
\newblock \showarticletitle{Geometric deep learning: Grids, groups, graphs,
  geodesics, and gauges}.
\newblock \bibinfo{journal}{\emph{arXiv preprint arXiv:2104.13478}}
  (\bibinfo{year}{2021}).
\newblock


\bibitem[\protect\citeauthoryear{Bronstein, Bruna, LeCun, Szlam, and
  Vandergheynst}{Bronstein et~al\mbox{.}}{2017}]%
        {bronstein2017geometric}
\bibfield{author}{\bibinfo{person}{Michael~M Bronstein}, \bibinfo{person}{Joan
  Bruna}, \bibinfo{person}{Yann LeCun}, \bibinfo{person}{Arthur Szlam}, {and}
  \bibinfo{person}{Pierre Vandergheynst}.} \bibinfo{year}{2017}\natexlab{}.
\newblock \showarticletitle{Geometric deep learning: going beyond euclidean
  data}.
\newblock \bibinfo{journal}{\emph{IEEE Signal Processing Magazine}}
  \bibinfo{volume}{34}, \bibinfo{number}{4} (\bibinfo{year}{2017}),
  \bibinfo{pages}{18--42}.
\newblock


\bibitem[\protect\citeauthoryear{Bulu{\c{c}} and Gilbert}{Bulu{\c{c}} and
  Gilbert}{2011}]%
        {bulucc2011combinatorial}
\bibfield{author}{\bibinfo{person}{Ayd{\i}n Bulu{\c{c}}} {and}
  \bibinfo{person}{John~R Gilbert}.} \bibinfo{year}{2011}\natexlab{}.
\newblock \showarticletitle{The Combinatorial BLAS: Design, implementation, and
  applications}.
\newblock \bibinfo{journal}{\emph{IJHPCA}} \bibinfo{volume}{25},
  \bibinfo{number}{4} (\bibinfo{year}{2011}), \bibinfo{pages}{496--509}.
\newblock


\bibitem[\protect\citeauthoryear{{Callidus Software Inc.}}{{Callidus Software
  Inc.}}{2018}]%
        {orientdb_lwedge_links}
\bibfield{author}{\bibinfo{person}{{Callidus Software Inc.}}}
  \bibinfo{year}{2018}\natexlab{}.
\newblock \bibinfo{title}{{OrientDB: Lightweight Edges.}}
\newblock \bibinfo{howpublished}{Available at
  \url{https://orientdb.com/docs/3.0.x/java/Lightweight-Edges.html}}.
\newblock


\bibitem[\protect\citeauthoryear{{Cambridge Semantics}}{{Cambridge
  Semantics}}{2018}]%
        {anzo_graph_links}
\bibfield{author}{\bibinfo{person}{{Cambridge Semantics}}.}
  \bibinfo{year}{2018}\natexlab{}.
\newblock \bibinfo{title}{{AnzoGraph}}.
\newblock \bibinfo{howpublished}{Available at
  \url{https://www.cambridgesemantics.com/product/anzograph/}}.
\newblock


\bibitem[\protect\citeauthoryear{Cao, Yan, He, and He}{Cao
  et~al\mbox{.}}{2020}]%
        {cao2020comprehensive}
\bibfield{author}{\bibinfo{person}{Wenming Cao}, \bibinfo{person}{Zhiyue Yan},
  \bibinfo{person}{Zhiquan He}, {and} \bibinfo{person}{Zhihai He}.}
  \bibinfo{year}{2020}\natexlab{}.
\newblock \showarticletitle{A comprehensive survey on geometric deep learning}.
\newblock \bibinfo{journal}{\emph{IEEE Access}}  \bibinfo{volume}{8}
  (\bibinfo{year}{2020}), \bibinfo{pages}{35929--35949}.
\newblock


\bibitem[\protect\citeauthoryear{{Cayley}}{{Cayley}}{2018}]%
        {cayley_links}
\bibfield{author}{\bibinfo{person}{{Cayley}}.} \bibinfo{year}{2018}\natexlab{}.
\newblock \bibinfo{title}{{CayleyGraph}}.
\newblock \bibinfo{howpublished}{Available at \url{https://cayley.io/} and
  \url{https://github.com/cayleygraph/cayley}}.
\newblock


\bibitem[\protect\citeauthoryear{Cen, Zou, Zhang, Yang, Zhou, and Tang}{Cen
  et~al\mbox{.}}{2019}]%
        {cen2019representation}
\bibfield{author}{\bibinfo{person}{Yukuo Cen}, \bibinfo{person}{Xu Zou},
  \bibinfo{person}{Jianwei Zhang}, \bibinfo{person}{Hongxia Yang},
  \bibinfo{person}{Jingren Zhou}, {and} \bibinfo{person}{Jie Tang}.}
  \bibinfo{year}{2019}\natexlab{}.
\newblock \showarticletitle{Representation learning for attributed multiplex
  heterogeneous network}. In \bibinfo{booktitle}{\emph{Proceedings of the 25th
  ACM SIGKDD International Conference on Knowledge Discovery \& Data Mining}}.
  \bibinfo{pages}{1358--1368}.
\newblock


\bibitem[\protect\citeauthoryear{Chami, Abu-El-Haija, Perozzi, R{\'e}, and
  Murphy}{Chami et~al\mbox{.}}{2020}]%
        {chami2020machine}
\bibfield{author}{\bibinfo{person}{Ines Chami}, \bibinfo{person}{Sami
  Abu-El-Haija}, \bibinfo{person}{Bryan Perozzi}, \bibinfo{person}{Christopher
  R{\'e}}, {and} \bibinfo{person}{Kevin Murphy}.}
  \bibinfo{year}{2020}\natexlab{}.
\newblock \showarticletitle{Machine learning on graphs: A model and
  comprehensive taxonomy}.
\newblock \bibinfo{journal}{\emph{arXiv preprint arXiv:2005.03675}}
  (\bibinfo{year}{2020}).
\newblock


\bibitem[\protect\citeauthoryear{Chen et~al\mbox{.}}{Chen
  et~al\mbox{.}}{2020a}]%
        {chen2020bridging}
\bibfield{author}{\bibinfo{person}{Zhiqian Chen} {et~al\mbox{.}}}
  \bibinfo{year}{2020}\natexlab{a}.
\newblock \showarticletitle{Bridging the gap between spatial and spectral
  domains: A survey on graph neural networks}.
\newblock \bibinfo{journal}{\emph{arXiv preprint arXiv:2002.11867}}
  (\bibinfo{year}{2020}).
\newblock


\bibitem[\protect\citeauthoryear{Chen, Wang, Zhao, Cheng, Zhao, and Duan}{Chen
  et~al\mbox{.}}{2020b}]%
        {chen2020knowledge}
\bibfield{author}{\bibinfo{person}{Zhe Chen}, \bibinfo{person}{Yuehan Wang},
  \bibinfo{person}{Bin Zhao}, \bibinfo{person}{Jing Cheng},
  \bibinfo{person}{Xin Zhao}, {and} \bibinfo{person}{Zongtao Duan}.}
  \bibinfo{year}{2020}\natexlab{b}.
\newblock \showarticletitle{Knowledge graph completion: A review}.
\newblock \bibinfo{journal}{\emph{Ieee Access}}  \bibinfo{volume}{8}
  (\bibinfo{year}{2020}), \bibinfo{pages}{192435--192456}.
\newblock


\bibitem[\protect\citeauthoryear{Choudhury, Agarwal, Purohit, Zhang, Pirrung,
  Smith, and Thomas}{Choudhury et~al\mbox{.}}{2017}]%
        {choudhury2017nous}
\bibfield{author}{\bibinfo{person}{Sutanay Choudhury}, \bibinfo{person}{Khushbu
  Agarwal}, \bibinfo{person}{Sumit Purohit}, \bibinfo{person}{Baichuan Zhang},
  \bibinfo{person}{Meg Pirrung}, \bibinfo{person}{Will Smith}, {and}
  \bibinfo{person}{Mathew Thomas}.} \bibinfo{year}{2017}\natexlab{}.
\newblock \showarticletitle{Nous: Construction and querying of dynamic
  knowledge graphs}. In \bibinfo{booktitle}{\emph{IEEE ICDE}}.
  \bibinfo{pages}{1563--1565}.
\newblock


\bibitem[\protect\citeauthoryear{Copik, Kwasniewski, Besta, Podstawski, and
  Hoefler}{Copik et~al\mbox{.}}{2020}]%
        {copik2020sebs}
\bibfield{author}{\bibinfo{person}{Marcin Copik}, \bibinfo{person}{Grzegorz
  Kwasniewski}, \bibinfo{person}{Maciej Besta}, \bibinfo{person}{Michal
  Podstawski}, {and} \bibinfo{person}{Torsten Hoefler}.}
  \bibinfo{year}{2020}\natexlab{}.
\newblock \showarticletitle{SeBS: A Serverless Benchmark Suite for
  Function-as-a-Service Computing}.
\newblock \bibinfo{journal}{\emph{arXiv preprint arXiv:2012.14132}}
  (\bibinfo{year}{2020}).
\newblock


\bibitem[\protect\citeauthoryear{Dai, Wang, Xiong, and Guo}{Dai
  et~al\mbox{.}}{2020}]%
        {dai2020survey}
\bibfield{author}{\bibinfo{person}{Yuanfei Dai}, \bibinfo{person}{Shiping
  Wang}, \bibinfo{person}{Neal~N Xiong}, {and} \bibinfo{person}{Wenzhong Guo}.}
  \bibinfo{year}{2020}\natexlab{}.
\newblock \showarticletitle{A survey on knowledge graph embedding: Approaches,
  applications and benchmarks}.
\newblock \bibinfo{journal}{\emph{Electronics}} \bibinfo{volume}{9},
  \bibinfo{number}{5} (\bibinfo{year}{2020}), \bibinfo{pages}{750}.
\newblock


\bibitem[\protect\citeauthoryear{{DataStax, Inc.}}{{DataStax, Inc.}}{2018}]%
        {datastax_links}
\bibfield{author}{\bibinfo{person}{{DataStax, Inc.}}}
  \bibinfo{year}{2018}\natexlab{}.
\newblock \bibinfo{title}{{DSE Graph (DataStax)}}.
\newblock \bibinfo{howpublished}{Available at \url{https://www.datastax.com/}}.
\newblock


\bibitem[\protect\citeauthoryear{Davies, Veli{\v{c}}kovi{\'c}, Buesing,
  Blackwell, Zheng, Toma{\v{s}}ev, Tanburn, Battaglia, Blundell, Juh{\'a}sz,
  et~al\mbox{.}}{Davies et~al\mbox{.}}{2021}]%
        {davies2021advancing}
\bibfield{author}{\bibinfo{person}{Alex Davies}, \bibinfo{person}{Petar
  Veli{\v{c}}kovi{\'c}}, \bibinfo{person}{Lars Buesing}, \bibinfo{person}{Sam
  Blackwell}, \bibinfo{person}{Daniel Zheng}, \bibinfo{person}{Nenad
  Toma{\v{s}}ev}, \bibinfo{person}{Richard Tanburn}, \bibinfo{person}{Peter
  Battaglia}, \bibinfo{person}{Charles Blundell}, \bibinfo{person}{Andr{\'a}s
  Juh{\'a}sz}, {et~al\mbox{.}}} \bibinfo{year}{2021}\natexlab{}.
\newblock \showarticletitle{Advancing mathematics by guiding human intuition
  with AI}.
\newblock \bibinfo{journal}{\emph{Nature}} \bibinfo{volume}{600},
  \bibinfo{number}{7887} (\bibinfo{year}{2021}), \bibinfo{pages}{70--74}.
\newblock


\bibitem[\protect\citeauthoryear{Davoudian, Chen, and Liu}{Davoudian
  et~al\mbox{.}}{2018}]%
        {davoudian2018survey}
\bibfield{author}{\bibinfo{person}{Ali Davoudian}, \bibinfo{person}{Liu Chen},
  {and} \bibinfo{person}{Mengchi Liu}.} \bibinfo{year}{2018}\natexlab{}.
\newblock \showarticletitle{{A survey on NoSQL stores}}.
\newblock \bibinfo{journal}{\emph{ACM Computing Surveys (CSUR)}}
  \bibinfo{volume}{51}, \bibinfo{number}{2}, Article \bibinfo{articleno}{40}
  (\bibinfo{year}{2018}), \bibinfo{numpages}{43}~pages.
\newblock
\showISSN{0360-0300}
\urldef\tempurl%
\url{https://doi.org/10.1145/3158661}
\showDOI{\tempurl}


\bibitem[\protect\citeauthoryear{Defferrard et~al\mbox{.}}{Defferrard
  et~al\mbox{.}}{2016}]%
        {defferrard2016convolutional}
\bibfield{author}{\bibinfo{person}{Micha{\"e}l Defferrard} {et~al\mbox{.}}}
  \bibinfo{year}{2016}\natexlab{}.
\newblock \showarticletitle{Convolutional neural networks on graphs with fast
  localized spectral filtering}.
\newblock \bibinfo{journal}{\emph{NeurIPS}} (\bibinfo{year}{2016}).
\newblock


\bibitem[\protect\citeauthoryear{Demetrescu, Eppstein, Galil, and
  Italiano}{Demetrescu et~al\mbox{.}}{2009}]%
        {demetrescu2009dynamic}
\bibfield{author}{\bibinfo{person}{Camil Demetrescu}, \bibinfo{person}{David
  Eppstein}, \bibinfo{person}{Zvi Galil}, {and} \bibinfo{person}{Giuseppe~F
  Italiano}.} \bibinfo{year}{2009}\natexlab{}.
\newblock \showarticletitle{Dynamic graph algorithms}.
\newblock In \bibinfo{booktitle}{\emph{Algorithms and Theory of Computation
  Handbook, Volume 1}}. \bibinfo{publisher}{Chapman and Hall/CRC},
  \bibinfo{pages}{235--262}.
\newblock


\bibitem[\protect\citeauthoryear{{Dgraph Labs, Inc.}}{{Dgraph Labs,
  Inc.}}{2018}]%
        {dgraph_links}
\bibfield{author}{\bibinfo{person}{{Dgraph Labs, Inc.}}}
  \bibinfo{year}{2018}\natexlab{}.
\newblock \bibinfo{title}{{DGraph}}.
\newblock \bibinfo{howpublished}{Available at \url{https://dgraph.io/},
  \url{https://docs.dgraph.io/design-concepts}}.
\newblock


\bibitem[\protect\citeauthoryear{Dubey, Hill, Escriva, and Sirer}{Dubey
  et~al\mbox{.}}{2016}]%
        {dubey2016weaver}
\bibfield{author}{\bibinfo{person}{Ayush Dubey}, \bibinfo{person}{Greg~D Hill},
  \bibinfo{person}{Robert Escriva}, {and} \bibinfo{person}{Emin~G{\"u}n
  Sirer}.} \bibinfo{year}{2016}\natexlab{}.
\newblock \showarticletitle{Weaver: a high-performance, transactional graph
  database based on refinable timestamps}.
\newblock \bibinfo{journal}{\emph{Proceedings of the VLDB Endowment}}
  \bibinfo{volume}{9}, \bibinfo{number}{11} (\bibinfo{year}{2016}),
  \bibinfo{pages}{852--863}.
\newblock


\bibitem[\protect\citeauthoryear{Dwivedi, Luu, Laurent, Bengio, and
  Bresson}{Dwivedi et~al\mbox{.}}{2021}]%
        {dwivedi2021graph}
\bibfield{author}{\bibinfo{person}{Vijay~Prakash Dwivedi},
  \bibinfo{person}{Anh~Tuan Luu}, \bibinfo{person}{Thomas Laurent},
  \bibinfo{person}{Yoshua Bengio}, {and} \bibinfo{person}{Xavier Bresson}.}
  \bibinfo{year}{2021}\natexlab{}.
\newblock \showarticletitle{Graph neural networks with learnable structural and
  positional representations}.
\newblock \bibinfo{journal}{\emph{arXiv preprint arXiv:2110.07875}}
  (\bibinfo{year}{2021}).
\newblock


\bibitem[\protect\citeauthoryear{Eppstein, Galil, and Italiano}{Eppstein
  et~al\mbox{.}}{1999}]%
        {eppstein1999dynamic}
\bibfield{author}{\bibinfo{person}{David Eppstein}, \bibinfo{person}{Zvi
  Galil}, {and} \bibinfo{person}{Giuseppe~F Italiano}.}
  \bibinfo{year}{1999}\natexlab{}.
\newblock \showarticletitle{Dynamic graph algorithms}.
\newblock \bibinfo{journal}{\emph{Algorithms and theory of computation
  handbook}}  \bibinfo{volume}{1} (\bibinfo{year}{1999}),
  \bibinfo{pages}{9--1}.
\newblock


\bibitem[\protect\citeauthoryear{{FactNexus}}{{FactNexus}}{2018}]%
        {graphbase_links}
\bibfield{author}{\bibinfo{person}{{FactNexus}}.}
  \bibinfo{year}{2018}\natexlab{}.
\newblock \bibinfo{title}{{GraphBase}}.
\newblock \bibinfo{howpublished}{Available at \url{https://graphbase.ai/}}.
\newblock


\bibitem[\protect\citeauthoryear{F{\"{a}}rber}{F{\"{a}}rber}{2019}]%
        {DBLP:conf/semweb/Farber19}
\bibfield{author}{\bibinfo{person}{Michael F{\"{a}}rber}.}
  \bibinfo{year}{2019}\natexlab{}.
\newblock \showarticletitle{{The Microsoft Academic Knowledge Graph: {A} Linked
  Data Source with 8 Billion Triples of Scholarly Data}}. In
  \bibinfo{booktitle}{\emph{{Proceedings of the 18th International Semantic Web
  Conference}}} ({Auckland, New Zealand}) \emph{(\bibinfo{series}{{ISWC'19}})}.
  \bibinfo{pages}{113--129}.
\newblock
\urldef\tempurl%
\url{https://doi.org/10.1007/978-3-030-30796-7\_8}
\showDOI{\tempurl}


\bibitem[\protect\citeauthoryear{Fey and Lenssen}{Fey and Lenssen}{2019}]%
        {fey2019fast}
\bibfield{author}{\bibinfo{person}{Matthias Fey} {and}
  \bibinfo{person}{Jan~Eric Lenssen}.} \bibinfo{year}{2019}\natexlab{}.
\newblock \showarticletitle{Fast graph representation learning with PyTorch
  Geometric}.
\newblock \bibinfo{journal}{\emph{arXiv preprint arXiv:1903.02428}}
  (\bibinfo{year}{2019}).
\newblock


\bibitem[\protect\citeauthoryear{{Franz Inc.}}{{Franz Inc.}}{2018}]%
        {allegro_graph_links}
\bibfield{author}{\bibinfo{person}{{Franz Inc.}}}
  \bibinfo{year}{2018}\natexlab{}.
\newblock \bibinfo{title}{{AllegroGraph}}.
\newblock \bibinfo{howpublished}{Available at
  \url{https://franz.com/agraph/allegrograph/}}.
\newblock


\bibitem[\protect\citeauthoryear{Fu, Zhang, Meng, and King}{Fu
  et~al\mbox{.}}{2020}]%
        {fu2020magnn}
\bibfield{author}{\bibinfo{person}{Xinyu Fu}, \bibinfo{person}{Jiani Zhang},
  \bibinfo{person}{Ziqiao Meng}, {and} \bibinfo{person}{Irwin King}.}
  \bibinfo{year}{2020}\natexlab{}.
\newblock \showarticletitle{Magnn: Metapath aggregated graph neural network for
  heterogeneous graph embedding}. In \bibinfo{booktitle}{\emph{Proceedings of
  The Web Conference 2020}}. \bibinfo{pages}{2331--2341}.
\newblock


\bibitem[\protect\citeauthoryear{Gajendran}{Gajendran}{2012}]%
        {gajendran2012survey}
\bibfield{author}{\bibinfo{person}{Santhosh~Kumar Gajendran}.}
  \bibinfo{year}{2012}\natexlab{}.
\newblock \showarticletitle{{A survey on NoSQL databases}}.
\newblock \bibinfo{journal}{\emph{University of Illinois}}
  (\bibinfo{year}{2012}).
\newblock


\bibitem[\protect\citeauthoryear{Geng, Li, Shi, Wu, Wang, Li, Haghi, Tumeo,
  Che, Reinhardt, et~al\mbox{.}}{Geng et~al\mbox{.}}{2020}]%
        {geng2020awb}
\bibfield{author}{\bibinfo{person}{Tong Geng}, \bibinfo{person}{Ang Li},
  \bibinfo{person}{Runbin Shi}, \bibinfo{person}{Chunshu Wu},
  \bibinfo{person}{Tianqi Wang}, \bibinfo{person}{Yanfei Li},
  \bibinfo{person}{Pouya Haghi}, \bibinfo{person}{Antonino Tumeo},
  \bibinfo{person}{Shuai Che}, \bibinfo{person}{Steve Reinhardt},
  {et~al\mbox{.}}} \bibinfo{year}{2020}\natexlab{}.
\newblock \showarticletitle{AWB-GCN: A graph convolutional network accelerator
  with runtime workload rebalancing}. In \bibinfo{booktitle}{\emph{IEEE/ACM
  MICRO}}.
\newblock


\bibitem[\protect\citeauthoryear{Gianinazzi, Besta, Schaffner, and
  Hoefler}{Gianinazzi et~al\mbox{.}}{2021a}]%
        {gianinazzi2021parallel}
\bibfield{author}{\bibinfo{person}{Lukas Gianinazzi}, \bibinfo{person}{Maciej
  Besta}, \bibinfo{person}{Yannick Schaffner}, {and} \bibinfo{person}{Torsten
  Hoefler}.} \bibinfo{year}{2021}\natexlab{a}.
\newblock \showarticletitle{Parallel Algorithms for Finding Large Cliques in
  Sparse Graphs}. In \bibinfo{booktitle}{\emph{Proceedings of the 33rd ACM
  Symposium on Parallelism in Algorithms and Architectures}}.
  \bibinfo{pages}{243--253}.
\newblock


\bibitem[\protect\citeauthoryear{Gianinazzi, Fries, Dryden, Ben-Nun, and
  Hoefler}{Gianinazzi et~al\mbox{.}}{2021b}]%
        {gianinazzi2021learning}
\bibfield{author}{\bibinfo{person}{Lukas Gianinazzi},
  \bibinfo{person}{Maximilian Fries}, \bibinfo{person}{Nikoli Dryden},
  \bibinfo{person}{Tal Ben-Nun}, {and} \bibinfo{person}{Torsten Hoefler}.}
  \bibinfo{year}{2021}\natexlab{b}.
\newblock \showarticletitle{Learning Combinatorial Node Labeling Algorithms}.
\newblock \bibinfo{journal}{\emph{arXiv preprint arXiv:2106.03594}}
  (\bibinfo{year}{2021}).
\newblock


\bibitem[\protect\citeauthoryear{Gianinazzi, Kalvoda, De~Palma, Besta, and
  Hoefler}{Gianinazzi et~al\mbox{.}}{2018}]%
        {gianinazzi2018communication}
\bibfield{author}{\bibinfo{person}{Lukas Gianinazzi}, \bibinfo{person}{Pavel
  Kalvoda}, \bibinfo{person}{Alessandro De~Palma}, \bibinfo{person}{Maciej
  Besta}, {and} \bibinfo{person}{Torsten Hoefler}.}
  \bibinfo{year}{2018}\natexlab{}.
\newblock \showarticletitle{Communication-avoiding parallel minimum cuts and
  connected components}, In \bibinfo{booktitle}{ACM SIGPLAN Notices}.
\newblock \bibinfo{journal}{\emph{ACM SIGPLAN Notices}} \bibinfo{volume}{53},
  \bibinfo{number}{1}, \bibinfo{pages}{219--232}.
\newblock
\showISSN{0362-1340}
\urldef\tempurl%
\url{https://doi.org/10.1145/3200691.3178504}
\showDOI{\tempurl}


\bibitem[\protect\citeauthoryear{Gilmer, Schoenholz, Riley, Vinyals, and
  Dahl}{Gilmer et~al\mbox{.}}{2017}]%
        {gilmer2017neural}
\bibfield{author}{\bibinfo{person}{Justin Gilmer}, \bibinfo{person}{Samuel~S
  Schoenholz}, \bibinfo{person}{Patrick~F Riley}, \bibinfo{person}{Oriol
  Vinyals}, {and} \bibinfo{person}{George~E Dahl}.}
  \bibinfo{year}{2017}\natexlab{}.
\newblock \showarticletitle{Neural message passing for quantum chemistry}. In
  \bibinfo{booktitle}{\emph{International Conference on Machine Learning}}.
  PMLR, \bibinfo{pages}{1263--1272}.
\newblock


\bibitem[\protect\citeauthoryear{Gleinig, Besta, and Hoefler}{Gleinig
  et~al\mbox{.}}{2022}]%
        {gleinig2022io}
\bibfield{author}{\bibinfo{person}{Niels Gleinig}, \bibinfo{person}{Maciej
  Besta}, {and} \bibinfo{person}{Torsten Hoefler}.}
  \bibinfo{year}{2022}\natexlab{}.
\newblock \showarticletitle{I/O-optimal Cache-oblivious Sparse Matrix-Sparse
  Matrix Multiplication}.
\newblock \bibinfo{journal}{\emph{IEEE IPDPS}}.
\newblock


\bibitem[\protect\citeauthoryear{Grover and Leskovec}{Grover and
  Leskovec}{2016}]%
        {grover2016node2vec}
\bibfield{author}{\bibinfo{person}{Aditya Grover} {and} \bibinfo{person}{Jure
  Leskovec}.} \bibinfo{year}{2016}\natexlab{}.
\newblock \showarticletitle{node2vec: Scalable feature learning for networks}.
  In \bibinfo{booktitle}{\emph{KDD}}. \bibinfo{pages}{855--864}.
\newblock


\bibitem[\protect\citeauthoryear{Hamilton}{Hamilton}{2020}]%
        {hamilton2020graph}
\bibfield{author}{\bibinfo{person}{William~L Hamilton}.}
  \bibinfo{year}{2020}\natexlab{}.
\newblock \showarticletitle{Graph representation learning}.
\newblock \bibinfo{journal}{\emph{Synthesis Lectures on Artifical Intelligence
  and Machine Learning}} \bibinfo{volume}{14}, \bibinfo{number}{3}
  (\bibinfo{year}{2020}), \bibinfo{pages}{1--159}.
\newblock


\bibitem[\protect\citeauthoryear{Hamilton et~al\mbox{.}}{Hamilton
  et~al\mbox{.}}{2017a}]%
        {hamilton2017representation}
\bibfield{author}{\bibinfo{person}{William~L Hamilton} {et~al\mbox{.}}}
  \bibinfo{year}{2017}\natexlab{a}.
\newblock \showarticletitle{Representation learning on graphs: Methods and
  applications}.
\newblock \bibinfo{journal}{\emph{arXiv preprint arXiv:1709.05584}}
  (\bibinfo{year}{2017}).
\newblock


\bibitem[\protect\citeauthoryear{Hamilton, Ying, and Leskovec}{Hamilton
  et~al\mbox{.}}{2017b}]%
        {hamilton2017inductive}
\bibfield{author}{\bibinfo{person}{William~L Hamilton}, \bibinfo{person}{Rex
  Ying}, {and} \bibinfo{person}{Jure Leskovec}.}
  \bibinfo{year}{2017}\natexlab{b}.
\newblock \showarticletitle{Inductive representation learning on large graphs}.
  In \bibinfo{booktitle}{\emph{NeurIPS}}.
\newblock


\bibitem[\protect\citeauthoryear{Han, Haihong, Le, and Du}{Han
  et~al\mbox{.}}{2011}]%
        {han2011survey}
\bibfield{author}{\bibinfo{person}{Jing Han}, \bibinfo{person}{E Haihong},
  \bibinfo{person}{Guan Le}, {and} \bibinfo{person}{Jian Du}.}
  \bibinfo{year}{2011}\natexlab{}.
\newblock \showarticletitle{{Survey on NoSQL database}}. In
  \bibinfo{booktitle}{\emph{2011 6th international conference on pervasive
  computing and applications}}. IEEE, \bibinfo{pages}{363--366}.
\newblock


\bibitem[\protect\citeauthoryear{Harris, Lamb, Shadbolt, et~al\mbox{.}}{Harris
  et~al\mbox{.}}{2009}]%
        {harris20094store}
\bibfield{author}{\bibinfo{person}{Steve Harris}, \bibinfo{person}{Nick Lamb},
  \bibinfo{person}{Nigel Shadbolt}, {et~al\mbox{.}}}
  \bibinfo{year}{2009}\natexlab{}.
\newblock \showarticletitle{4store: The design and implementation of a
  clustered RDF store}. In \bibinfo{booktitle}{\emph{5th International Workshop
  on Scalable Semantic Web Knowledge Base Systems (SSWS2009)}},
  Vol.~\bibinfo{volume}{94}.
\newblock


\bibitem[\protect\citeauthoryear{Henzinger and King}{Henzinger and
  King}{1999}]%
        {henzinger1999randomized}
\bibfield{author}{\bibinfo{person}{Monika~R Henzinger} {and}
  \bibinfo{person}{Valerie King}.} \bibinfo{year}{1999}\natexlab{}.
\newblock \showarticletitle{Randomized fully dynamic graph algorithms with
  polylogarithmic time per operation}.
\newblock \bibinfo{journal}{\emph{Journal of the ACM (JACM)}}
  \bibinfo{volume}{46}, \bibinfo{number}{4} (\bibinfo{year}{1999}),
  \bibinfo{pages}{502--516}.
\newblock


\bibitem[\protect\citeauthoryear{Hodler and Needham}{Hodler and
  Needham}{2022}]%
        {hodlergraph}
\bibfield{author}{\bibinfo{person}{Amy~E Hodler} {and} \bibinfo{person}{Mark
  Needham}.} \bibinfo{year}{2022}\natexlab{}.
\newblock \showarticletitle{Graph Data Science Using Neo4j}.
\newblock In \bibinfo{booktitle}{\emph{Massive Graph Analytics}}.
  \bibinfo{publisher}{Chapman and Hall/CRC}, \bibinfo{pages}{433--457}.
\newblock


\bibitem[\protect\citeauthoryear{Hu et~al\mbox{.}}{Hu et~al\mbox{.}}{2020b}]%
        {hu2020featgraph}
\bibfield{author}{\bibinfo{person}{Yuwei Hu} {et~al\mbox{.}}}
  \bibinfo{year}{2020}\natexlab{b}.
\newblock \showarticletitle{Featgraph: A flexible and efficient backend for
  graph neural network systems}.
\newblock \bibinfo{journal}{\emph{arXiv preprint arXiv:2008.11359}}
  (\bibinfo{year}{2020}).
\newblock


\bibitem[\protect\citeauthoryear{Hu, Dong, Wang, and Sun}{Hu
  et~al\mbox{.}}{2020a}]%
        {hu2020heterogeneous}
\bibfield{author}{\bibinfo{person}{Ziniu Hu}, \bibinfo{person}{Yuxiao Dong},
  \bibinfo{person}{Kuansan Wang}, {and} \bibinfo{person}{Yizhou Sun}.}
  \bibinfo{year}{2020}\natexlab{a}.
\newblock \showarticletitle{Heterogeneous graph transformer}. In
  \bibinfo{booktitle}{\emph{Proceedings of The Web Conference 2020}}.
  \bibinfo{pages}{2704--2710}.
\newblock


\bibitem[\protect\citeauthoryear{Ivkovi{\'c} and Lloyd}{Ivkovi{\'c} and
  Lloyd}{1993}]%
        {ivkovic1993fully}
\bibfield{author}{\bibinfo{person}{Zoran Ivkovi{\'c}} {and}
  \bibinfo{person}{Errol~L Lloyd}.} \bibinfo{year}{1993}\natexlab{}.
\newblock \showarticletitle{Fully dynamic maintenance of vertex cover}. In
  \bibinfo{booktitle}{\emph{International Workshop on Graph-Theoretic Concepts
  in Computer Science}}. Springer, \bibinfo{pages}{99--111}.
\newblock


\bibitem[\protect\citeauthoryear{Jia et~al\mbox{.}}{Jia et~al\mbox{.}}{2020}]%
        {jia2020improving}
\bibfield{author}{\bibinfo{person}{Zhihao Jia} {et~al\mbox{.}}}
  \bibinfo{year}{2020}\natexlab{}.
\newblock \showarticletitle{Improving the accuracy, scalability, and
  performance of graph neural networks with roc}.
\newblock \bibinfo{journal}{\emph{MLSys}} (\bibinfo{year}{2020}).
\newblock


\bibitem[\protect\citeauthoryear{Jumper, Evans, Pritzel, Green, Figurnov,
  Ronneberger, Tunyasuvunakool, Bates, {\v{Z}}{\'\i}dek, Potapenko,
  et~al\mbox{.}}{Jumper et~al\mbox{.}}{2021}]%
        {jumper2021highly}
\bibfield{author}{\bibinfo{person}{John Jumper}, \bibinfo{person}{Richard
  Evans}, \bibinfo{person}{Alexander Pritzel}, \bibinfo{person}{Tim Green},
  \bibinfo{person}{Michael Figurnov}, \bibinfo{person}{Olaf Ronneberger},
  \bibinfo{person}{Kathryn Tunyasuvunakool}, \bibinfo{person}{Russ Bates},
  \bibinfo{person}{Augustin {\v{Z}}{\'\i}dek}, \bibinfo{person}{Anna
  Potapenko}, {et~al\mbox{.}}} \bibinfo{year}{2021}\natexlab{}.
\newblock \showarticletitle{Highly accurate protein structure prediction with
  AlphaFold}.
\newblock \bibinfo{journal}{\emph{Nature}} \bibinfo{volume}{596},
  \bibinfo{number}{7873} (\bibinfo{year}{2021}), \bibinfo{pages}{583--589}.
\newblock


\bibitem[\protect\citeauthoryear{Junghanns, Petermann, Neumann, and
  Rahm}{Junghanns et~al\mbox{.}}{2017}]%
        {junghanns2017management}
\bibfield{author}{\bibinfo{person}{Martin Junghanns},
  \bibinfo{person}{Andr{\'e} Petermann}, \bibinfo{person}{Martin Neumann},
  {and} \bibinfo{person}{Erhard Rahm}.} \bibinfo{year}{2017}\natexlab{}.
\newblock \showarticletitle{Management and analysis of big graph data: current
  systems and open challenges}.
\newblock In \bibinfo{booktitle}{\emph{Handbook of Big Data Technologies}}.
  \bibinfo{publisher}{Springer}, \bibinfo{pages}{457--505}.
\newblock


\bibitem[\protect\citeauthoryear{Jurisch and Igler}{Jurisch and Igler}{2018}]%
        {jurisch2018rdf2vec}
\bibfield{author}{\bibinfo{person}{Matthias Jurisch} {and}
  \bibinfo{person}{Bodo Igler}.} \bibinfo{year}{2018}\natexlab{}.
\newblock \showarticletitle{RDF2Vec-based classification of ontology alignment
  changes}.
\newblock \bibinfo{journal}{\emph{arXiv preprint arXiv:1805.09145}}
  (\bibinfo{year}{2018}).
\newblock


\bibitem[\protect\citeauthoryear{Kaliyar}{Kaliyar}{2015}]%
        {gdb_survey_paper_Kaliyar}
\bibfield{author}{\bibinfo{person}{R.~Kumar Kaliyar}.}
  \bibinfo{year}{2015}\natexlab{}.
\newblock \showarticletitle{Graph databases: A survey}. In
  \bibinfo{booktitle}{\emph{ICCCA}}. \bibinfo{pages}{785--790}.
\newblock


\bibitem[\protect\citeauthoryear{Kang, Tong, Sun, Lin, and Faloutsos}{Kang
  et~al\mbox{.}}{2012}]%
        {gbase_paper}
\bibfield{author}{\bibinfo{person}{U. Kang}, \bibinfo{person}{Hanghang Tong},
  \bibinfo{person}{Jimeng Sun}, \bibinfo{person}{Ching-Yung Lin}, {and}
  \bibinfo{person}{Christos Faloutsos}.} \bibinfo{year}{2012}\natexlab{}.
\newblock \showarticletitle{Gbase: An Efficient Analysis Platform for Large
  Graphs}.
\newblock \bibinfo{journal}{\emph{In PVLDB}} \bibinfo{volume}{21},
  \bibinfo{number}{5} (\bibinfo{year}{2012}), \bibinfo{pages}{637--650}.
\newblock
\showISSN{1066-8888}
\urldef\tempurl%
\url{https://doi.org/10.1007/s00778-012-0283-9}
\showDOI{\tempurl}


\bibitem[\protect\citeauthoryear{Kankanamge, Sahu, Mhedbhi, Chen,
  et~al\mbox{.}}{Kankanamge et~al\mbox{.}}{2017}]%
        {graphflow}
\bibfield{author}{\bibinfo{person}{Chathura Kankanamge},
  \bibinfo{person}{Siddhartha Sahu}, \bibinfo{person}{Amine Mhedbhi},
  \bibinfo{person}{Jeremy Chen}, {et~al\mbox{.}}}
  \bibinfo{year}{2017}\natexlab{}.
\newblock \showarticletitle{{Graphflow: An Active Graph Database}}. In
  \bibinfo{booktitle}{\emph{ACM SIGMOD}} (Chicago, Illinois, USA).
  \bibinfo{pages}{1695--1698}.
\newblock
\showISBNx{9781450341974}
\urldef\tempurl%
\url{https://doi.org/10.1145/3035918.3056445}
\showDOI{\tempurl}


\bibitem[\protect\citeauthoryear{Kepner, Aaltonen, Bader, Bulu{\c{c}},
  Franchetti, Gilbert, Hutchison, Kumar, Lumsdaine, Meyerhenke,
  et~al\mbox{.}}{Kepner et~al\mbox{.}}{2016}]%
        {kepner2016mathematical}
\bibfield{author}{\bibinfo{person}{Jeremy Kepner}, \bibinfo{person}{Peter
  Aaltonen}, \bibinfo{person}{David Bader}, \bibinfo{person}{Aydin
  Bulu{\c{c}}}, \bibinfo{person}{Franz Franchetti}, \bibinfo{person}{John
  Gilbert}, \bibinfo{person}{Dylan Hutchison}, \bibinfo{person}{Manoj Kumar},
  \bibinfo{person}{Andrew Lumsdaine}, \bibinfo{person}{Henning Meyerhenke},
  {et~al\mbox{.}}} \bibinfo{year}{2016}\natexlab{}.
\newblock \showarticletitle{Mathematical foundations of the GraphBLAS}. In
  \bibinfo{booktitle}{\emph{2016 IEEE High Performance Extreme Computing
  Conference (HPEC)}}. IEEE, \bibinfo{pages}{1--9}.
\newblock


\bibitem[\protect\citeauthoryear{Kingma and Ba}{Kingma and Ba}{2014}]%
        {kingma2014adam}
\bibfield{author}{\bibinfo{person}{Diederik~P Kingma} {and}
  \bibinfo{person}{Jimmy Ba}.} \bibinfo{year}{2014}\natexlab{}.
\newblock \showarticletitle{Adam: A method for stochastic optimization}.
\newblock \bibinfo{journal}{\emph{arXiv preprint arXiv:1412.6980}}
  (\bibinfo{year}{2014}).
\newblock


\bibitem[\protect\citeauthoryear{Kiningham, Levis, and R{\'e}}{Kiningham
  et~al\mbox{.}}{2020a}]%
        {kiningham2020greta}
\bibfield{author}{\bibinfo{person}{Kevin Kiningham}, \bibinfo{person}{Philip
  Levis}, {and} \bibinfo{person}{Christopher R{\'e}}.}
  \bibinfo{year}{2020}\natexlab{a}.
\newblock \showarticletitle{GReTA: Hardware Optimized Graph Processing for
  GNNs}. In \bibinfo{booktitle}{\emph{ReCoML}}.
\newblock


\bibitem[\protect\citeauthoryear{Kiningham, Re, and Levis}{Kiningham
  et~al\mbox{.}}{2020b}]%
        {kiningham2020grip}
\bibfield{author}{\bibinfo{person}{Kevin Kiningham},
  \bibinfo{person}{Christopher Re}, {and} \bibinfo{person}{Philip Levis}.}
  \bibinfo{year}{2020}\natexlab{b}.
\newblock \showarticletitle{GRIP: a graph neural network accelerator
  architecture}.
\newblock \bibinfo{journal}{\emph{arXiv preprint arXiv:2007.13828}}
  (\bibinfo{year}{2020}).
\newblock


\bibitem[\protect\citeauthoryear{Kipf and Welling}{Kipf and Welling}{2016}]%
        {kipf2016semi}
\bibfield{author}{\bibinfo{person}{Thomas~N Kipf} {and} \bibinfo{person}{Max
  Welling}.} \bibinfo{year}{2016}\natexlab{}.
\newblock \showarticletitle{Semi-supervised classification with graph
  convolutional networks}.
\newblock \bibinfo{journal}{\emph{arXiv preprint arXiv:1609.02907}}
  (\bibinfo{year}{2016}).
\newblock


\bibitem[\protect\citeauthoryear{Klicpera et~al\mbox{.}}{Klicpera
  et~al\mbox{.}}{2018}]%
        {klicpera2018predict}
\bibfield{author}{\bibinfo{person}{Johannes Klicpera} {et~al\mbox{.}}}
  \bibinfo{year}{2018}\natexlab{}.
\newblock \showarticletitle{Predict then propagate: Graph neural networks meet
  personalized pagerank}.
\newblock \bibinfo{journal}{\emph{arXiv preprint arXiv:1810.05997}}
  (\bibinfo{year}{2018}).
\newblock


\bibitem[\protect\citeauthoryear{Kumar and Babu}{Kumar and Babu}{2015}]%
        {kumar2015domain}
\bibfield{author}{\bibinfo{person}{Vijay Kumar} {and} \bibinfo{person}{Anjan
  Babu}.} \bibinfo{year}{2015}\natexlab{}.
\newblock \showarticletitle{Domain Suitable Graph Database Selection: A
  Preliminary Report}. In \bibinfo{booktitle}{\emph{3rd International
  Conference on Advances in Engineering Sciences \& Applied Mathematics,
  London, UK}}. \bibinfo{pages}{26--29}.
\newblock


\bibitem[\protect\citeauthoryear{Lassila, Swick, et~al\mbox{.}}{Lassila
  et~al\mbox{.}}{1998}]%
        {lassila1998resource}
\bibfield{author}{\bibinfo{person}{Ora Lassila}, \bibinfo{person}{Ralph~R
  Swick}, {et~al\mbox{.}}} \bibinfo{year}{1998}\natexlab{}.
\newblock \showarticletitle{Resource description framework (RDF) model and
  syntax specification}.
\newblock  (\bibinfo{year}{1998}).
\newblock


\bibitem[\protect\citeauthoryear{Lee, Choi, and Chung}{Lee
  et~al\mbox{.}}{2016}]%
        {lee2016efficient}
\bibfield{author}{\bibinfo{person}{Min-Joong Lee}, \bibinfo{person}{Sunghee
  Choi}, {and} \bibinfo{person}{Chin-Wan Chung}.}
  \bibinfo{year}{2016}\natexlab{}.
\newblock \showarticletitle{Efficient algorithms for updating betweenness
  centrality in fully dynamic graphs}.
\newblock \bibinfo{journal}{\emph{Information Sciences}}  \bibinfo{volume}{326}
  (\bibinfo{year}{2016}), \bibinfo{pages}{278--296}.
\newblock


\bibitem[\protect\citeauthoryear{Li et~al\mbox{.}}{Li et~al\mbox{.}}{2020}]%
        {li2020pytorch}
\bibfield{author}{\bibinfo{person}{Shen Li} {et~al\mbox{.}}}
  \bibinfo{year}{2020}\natexlab{}.
\newblock \showarticletitle{Pytorch distributed: Experiences on accelerating
  data parallel training}.
\newblock \bibinfo{journal}{\emph{arXiv preprint arXiv:2006.15704}}
  (\bibinfo{year}{2020}).
\newblock


\bibitem[\protect\citeauthoryear{Liang, Wang, Liu, He, Huawei, Xu, and
  Li}{Liang et~al\mbox{.}}{2020}]%
        {liang2020engn}
\bibfield{author}{\bibinfo{person}{Shengwen Liang}, \bibinfo{person}{Ying
  Wang}, \bibinfo{person}{Cheng Liu}, \bibinfo{person}{Lei He},
  \bibinfo{person}{LI Huawei}, \bibinfo{person}{Dawen Xu}, {and}
  \bibinfo{person}{Xiaowei Li}.} \bibinfo{year}{2020}\natexlab{}.
\newblock \showarticletitle{Engn: A high-throughput and energy-efficient
  accelerator for large graph neural networks}.
\newblock \bibinfo{journal}{\emph{IEEE TOC}} (\bibinfo{year}{2020}).
\newblock


\bibitem[\protect\citeauthoryear{Lin, Liu, Sun, Liu, and Zhu}{Lin
  et~al\mbox{.}}{2015}]%
        {lin2015learning}
\bibfield{author}{\bibinfo{person}{Yankai Lin}, \bibinfo{person}{Zhiyuan Liu},
  \bibinfo{person}{Maosong Sun}, \bibinfo{person}{Yang Liu}, {and}
  \bibinfo{person}{Xuan Zhu}.} \bibinfo{year}{2015}\natexlab{}.
\newblock \showarticletitle{Learning entity and relation embeddings for
  knowledge graph completion}. In \bibinfo{booktitle}{\emph{AAAI}}.
\newblock


\bibitem[\protect\citeauthoryear{Lv, Ding, Liu, Chen, Feng, He, Zhou, Jiang,
  Dong, and Tang}{Lv et~al\mbox{.}}{2021}]%
        {lv2021we}
\bibfield{author}{\bibinfo{person}{Qingsong Lv}, \bibinfo{person}{Ming Ding},
  \bibinfo{person}{Qiang Liu}, \bibinfo{person}{Yuxiang Chen},
  \bibinfo{person}{Wenzheng Feng}, \bibinfo{person}{Siming He},
  \bibinfo{person}{Chang Zhou}, \bibinfo{person}{Jianguo Jiang},
  \bibinfo{person}{Yuxiao Dong}, {and} \bibinfo{person}{Jie Tang}.}
  \bibinfo{year}{2021}\natexlab{}.
\newblock \showarticletitle{Are we really making much progress? Revisiting,
  benchmarking and refining heterogeneous graph neural networks}. In
  \bibinfo{booktitle}{\emph{Proceedings of the 27th ACM SIGKDD Conference on
  Knowledge Discovery \& Data Mining}}. \bibinfo{pages}{1150--1160}.
\newblock


\bibitem[\protect\citeauthoryear{Lyu, Qin, Lin, Chang, and Yu}{Lyu
  et~al\mbox{.}}{2016}]%
        {lyu2016scalable}
\bibfield{author}{\bibinfo{person}{Bingqing Lyu}, \bibinfo{person}{Lu Qin},
  \bibinfo{person}{Xuemin Lin}, \bibinfo{person}{Lijun Chang}, {and}
  \bibinfo{person}{Jeffrey~Xu Yu}.} \bibinfo{year}{2016}\natexlab{}.
\newblock \showarticletitle{Scalable supergraph search in large graph
  databases}. In \bibinfo{booktitle}{\emph{2016 IEEE 32nd International
  Conference on Data Engineering (ICDE)}}. IEEE, \bibinfo{pages}{157--168}.
\newblock


\bibitem[\protect\citeauthoryear{Ma, Yang, Miao, Xue, Wu, Zhou, and Dai}{Ma
  et~al\mbox{.}}{2019}]%
        {ma2019neugraph}
\bibfield{author}{\bibinfo{person}{Lingxiao Ma}, \bibinfo{person}{Zhi Yang},
  \bibinfo{person}{Youshan Miao}, \bibinfo{person}{Jilong Xue},
  \bibinfo{person}{Ming Wu}, \bibinfo{person}{Lidong Zhou}, {and}
  \bibinfo{person}{Yafei Dai}.} \bibinfo{year}{2019}\natexlab{}.
\newblock \showarticletitle{Neugraph: parallel deep neural network computation
  on large graphs}. In \bibinfo{booktitle}{\emph{USENIX ATC}}.
\newblock


\bibitem[\protect\citeauthoryear{Ma, Li, Hu, Lin, and Huai}{Ma
  et~al\mbox{.}}{2016}]%
        {ma2016big}
\bibfield{author}{\bibinfo{person}{Shuai Ma}, \bibinfo{person}{Jia Li},
  \bibinfo{person}{Chunming Hu}, \bibinfo{person}{Xuelian Lin}, {and}
  \bibinfo{person}{Jinpeng Huai}.} \bibinfo{year}{2016}\natexlab{}.
\newblock \showarticletitle{Big graph search: challenges and techniques}.
\newblock \bibinfo{journal}{\emph{Frontiers of Computer Science}}
  \bibinfo{volume}{10}, \bibinfo{number}{3} (\bibinfo{year}{2016}),
  \bibinfo{pages}{387--398}.
\newblock


\bibitem[\protect\citeauthoryear{Madduri, Ediger, Jiang, Bader, and
  Chavarria-Miranda}{Madduri et~al\mbox{.}}{2009}]%
        {madduri2009faster}
\bibfield{author}{\bibinfo{person}{Kamesh Madduri}, \bibinfo{person}{David
  Ediger}, \bibinfo{person}{Karl Jiang}, \bibinfo{person}{David~A Bader}, {and}
  \bibinfo{person}{Daniel Chavarria-Miranda}.} \bibinfo{year}{2009}\natexlab{}.
\newblock \showarticletitle{A faster parallel algorithm and efficient
  multithreaded implementations for evaluating betweenness centrality on
  massive datasets}. In \bibinfo{booktitle}{\emph{Parallel \& Distributed
  Processing, 2009. IPDPS 2009. IEEE International Symposium on}}. IEEE,
  \bibinfo{pages}{1--8}.
\newblock


\bibitem[\protect\citeauthoryear{Mao, Wang, Li, Huang, Zhang, Liao, and Ma}{Mao
  et~al\mbox{.}}{2022}]%
        {mao2022ermer}
\bibfield{author}{\bibinfo{person}{Zhitao Mao}, \bibinfo{person}{Ruoyu Wang},
  \bibinfo{person}{Haoran Li}, \bibinfo{person}{Yixin Huang},
  \bibinfo{person}{Qiang Zhang}, \bibinfo{person}{Xiaoping Liao}, {and}
  \bibinfo{person}{Hongwu Ma}.} \bibinfo{year}{2022}\natexlab{}.
\newblock \showarticletitle{ERMer: a serverless platform for navigating,
  analyzing, and visualizing Escherichia coli regulatory landscape through
  graph database}.
\newblock \bibinfo{journal}{\emph{Nucleic Acids Research}}
  (\bibinfo{year}{2022}).
\newblock


\bibitem[\protect\citeauthoryear{Martínez-Bazan, Muntés-Mulero,
  Gómez-Villamor, Águila Lorente, Dominguez-Sal, and
  Larriba-Pey}{Martínez-Bazan et~al\mbox{.}}{2012}]%
        {sparksee_paper}
\bibfield{author}{\bibinfo{person}{Norbert Martínez-Bazan},
  \bibinfo{person}{Victor Muntés-Mulero}, \bibinfo{person}{Sergio
  Gómez-Villamor}, \bibinfo{person}{M.Ángel Águila Lorente},
  \bibinfo{person}{David Dominguez-Sal}, {and} \bibinfo{person}{Josep-L.
  Larriba-Pey}.} \bibinfo{year}{2012}\natexlab{}.
\newblock \showarticletitle{{Efficient Graph Management Based On Bitmap
  Indices}}.
\newblock \bibinfo{journal}{\emph{In IDEAS}} (\bibinfo{year}{2012}),
  \bibinfo{pages}{110--119}.
\newblock
\showISBNx{9781450312349}
\urldef\tempurl%
\url{https://doi.org/10.1145/2351476.2351489}
\showDOI{\tempurl}


\bibitem[\protect\citeauthoryear{{Memgraph Ltd.}}{{Memgraph Ltd.}}{2018}]%
        {memgraph_links}
\bibfield{author}{\bibinfo{person}{{Memgraph Ltd.}}}
  \bibinfo{year}{2018}\natexlab{}.
\newblock \bibinfo{title}{{Memgraph}}.
\newblock \bibinfo{howpublished}{Available at \url{https://memgraph.com/}}.
\newblock


\bibitem[\protect\citeauthoryear{{Microsoft}}{{Microsoft}}{2018}]%
        {azure_cosmosdb_links}
\bibfield{author}{\bibinfo{person}{{Microsoft}}.}
  \bibinfo{year}{2018}\natexlab{}.
\newblock \bibinfo{title}{{Azure Cosmos DB}}.
\newblock \bibinfo{howpublished}{Available at
  \url{https://azure.microsoft.com/en-us/services/cosmos-db/}}.
\newblock


\bibitem[\protect\citeauthoryear{Miller}{Miller}{2013}]%
        {miller2013graph}
\bibfield{author}{\bibinfo{person}{Justin~J Miller}.}
  \bibinfo{year}{2013}\natexlab{}.
\newblock \showarticletitle{{Graph Database Applications and Concepts with
  Neo4j}}. In \bibinfo{booktitle}{\emph{{Proceedings of the Southern
  Association for Information Systems Conference}}},
  Vol.~\bibinfo{volume}{2324}.
\newblock


\bibitem[\protect\citeauthoryear{Mirhoseini, Goldie, Yazgan, Jiang, Songhori,
  Wang, Lee, Johnson, Pathak, Nazi, et~al\mbox{.}}{Mirhoseini
  et~al\mbox{.}}{2021}]%
        {mirhoseini2021graph}
\bibfield{author}{\bibinfo{person}{Azalia Mirhoseini}, \bibinfo{person}{Anna
  Goldie}, \bibinfo{person}{Mustafa Yazgan}, \bibinfo{person}{Joe~Wenjie
  Jiang}, \bibinfo{person}{Ebrahim Songhori}, \bibinfo{person}{Shen Wang},
  \bibinfo{person}{Young-Joon Lee}, \bibinfo{person}{Eric Johnson},
  \bibinfo{person}{Omkar Pathak}, \bibinfo{person}{Azade Nazi},
  {et~al\mbox{.}}} \bibinfo{year}{2021}\natexlab{}.
\newblock \showarticletitle{A graph placement methodology for fast chip
  design}.
\newblock \bibinfo{journal}{\emph{Nature}} \bibinfo{volume}{594},
  \bibinfo{number}{7862} (\bibinfo{year}{2021}), \bibinfo{pages}{207--212}.
\newblock


\bibitem[\protect\citeauthoryear{Modoni, Sacco, and Terkaj}{Modoni
  et~al\mbox{.}}{2014}]%
        {modoni2014survey}
\bibfield{author}{\bibinfo{person}{Gianfranco~E Modoni}, \bibinfo{person}{Marco
  Sacco}, {and} \bibinfo{person}{Walter Terkaj}.}
  \bibinfo{year}{2014}\natexlab{}.
\newblock \showarticletitle{A survey of RDF store solutions}. In
  \bibinfo{booktitle}{\emph{2014 International Conference on Engineering,
  Technology and Innovation (ICE)}}. IEEE, \bibinfo{pages}{1--7}.
\newblock


\bibitem[\protect\citeauthoryear{Monti, Boscaini, Masci, Rodola, Svoboda, and
  Bronstein}{Monti et~al\mbox{.}}{2017}]%
        {monti2017geometric}
\bibfield{author}{\bibinfo{person}{Federico Monti}, \bibinfo{person}{Davide
  Boscaini}, \bibinfo{person}{Jonathan Masci}, \bibinfo{person}{Emanuele
  Rodola}, \bibinfo{person}{Jan Svoboda}, {and} \bibinfo{person}{Michael~M
  Bronstein}.} \bibinfo{year}{2017}\natexlab{}.
\newblock \showarticletitle{Geometric deep learning on graphs and manifolds
  using mixture model cnns}. In \bibinfo{booktitle}{\emph{IEEE CVPR}}.
\newblock


\bibitem[\protect\citeauthoryear{Morris, Ritzert, Fey, Hamilton, Lenssen,
  Rattan, and Grohe}{Morris et~al\mbox{.}}{2019}]%
        {morris2019weisfeiler}
\bibfield{author}{\bibinfo{person}{Christopher Morris}, \bibinfo{person}{Martin
  Ritzert}, \bibinfo{person}{Matthias Fey}, \bibinfo{person}{William~L
  Hamilton}, \bibinfo{person}{Jan~Eric Lenssen}, \bibinfo{person}{Gaurav
  Rattan}, {and} \bibinfo{person}{Martin Grohe}.}
  \bibinfo{year}{2019}\natexlab{}.
\newblock \showarticletitle{Weisfeiler and leman go neural: Higher-order graph
  neural networks}. In \bibinfo{booktitle}{\emph{Proceedings of the AAAI
  conference on artificial intelligence}}, Vol.~\bibinfo{volume}{33}.
  \bibinfo{pages}{4602--4609}.
\newblock


\bibitem[\protect\citeauthoryear{Nabti and Seba}{Nabti and Seba}{2017}]%
        {nabti2017querying}
\bibfield{author}{\bibinfo{person}{Chemseddine Nabti} {and}
  \bibinfo{person}{Hamida Seba}.} \bibinfo{year}{2017}\natexlab{}.
\newblock \showarticletitle{{Querying massive graph data: A compress and search
  approach}}.
\newblock \bibinfo{journal}{\emph{Future Generation Computer Systems}}
  \bibinfo{volume}{74} (\bibinfo{year}{2017}), \bibinfo{pages}{63--75}.
\newblock


\bibitem[\protect\citeauthoryear{{Neo4j, Inc.}}{{Neo4j, Inc.}}{2018}]%
        {neo3j_release}
\bibfield{author}{\bibinfo{person}{{Neo4j, Inc.}}}
  \bibinfo{year}{2018}\natexlab{}.
\newblock \bibinfo{title}{Neo4j (3.0 Release)}.
\newblock \bibinfo{howpublished}{\newline Available at
  \url{https://neo4j.com/blog/neo4j-3-0-massive-scale-developer-productivity/}}.
\newblock


\bibitem[\protect\citeauthoryear{{Networked Planet Limited}}{{Networked Planet
  Limited}}{2018}]%
        {brightstardb_links}
\bibfield{author}{\bibinfo{person}{{Networked Planet Limited}}.}
  \bibinfo{year}{2018}\natexlab{}.
\newblock \bibinfo{title}{{BrightstarDB}}.
\newblock \bibinfo{howpublished}{Available at \url{http://brightstardb.com/}}.
\newblock


\bibitem[\protect\citeauthoryear{{Objectivity Inc.}}{{Objectivity
  Inc.}}{2018}]%
        {infinite_graph_links}
\bibfield{author}{\bibinfo{person}{{Objectivity Inc.}}}
  \bibinfo{year}{2018}\natexlab{}.
\newblock \bibinfo{title}{{InfiniteGraph}}.
\newblock \bibinfo{howpublished}{Available at
  \url{https://www.objectivity.com/products/infinitegraph/}}.
\newblock


\bibitem[\protect\citeauthoryear{{Ontotext}}{{Ontotext}}{2018}]%
        {graphdb_links}
\bibfield{author}{\bibinfo{person}{{Ontotext}}.}
  \bibinfo{year}{2018}\natexlab{}.
\newblock \bibinfo{title}{{GraphDB}}.
\newblock \bibinfo{howpublished}{Available at
  \url{https://www.ontotext.com/products/graphdb/}}.
\newblock


\bibitem[\protect\citeauthoryear{{OpenLink}}{{OpenLink}}{2018}]%
        {virtuoso_links}
\bibfield{author}{\bibinfo{person}{{OpenLink}}.}
  \bibinfo{year}{2018}\natexlab{}.
\newblock \bibinfo{title}{{Virtuoso}}.
\newblock \bibinfo{howpublished}{Available at
  \url{https://virtuoso.openlinksw.com/}}.
\newblock


\bibitem[\protect\citeauthoryear{{Oracle}}{{Oracle}}{2018}]%
        {oracle_spatial}
\bibfield{author}{\bibinfo{person}{{Oracle}}.} \bibinfo{year}{2018}\natexlab{}.
\newblock \bibinfo{title}{{Oracle Spatial and Graph}}.
\newblock \bibinfo{howpublished}{Available at
  \url{https://www.oracle.com/database/technologies/spatialandgraph.html}}.
\newblock


\bibitem[\protect\citeauthoryear{Papailiou, Konstantinou, Tsoumakos, and
  Koziris}{Papailiou et~al\mbox{.}}{2012}]%
        {papailiou2012h2rdf}
\bibfield{author}{\bibinfo{person}{Nikolaos Papailiou},
  \bibinfo{person}{Ioannis Konstantinou}, \bibinfo{person}{Dimitrios
  Tsoumakos}, {and} \bibinfo{person}{Nectarios Koziris}.}
  \bibinfo{year}{2012}\natexlab{}.
\newblock \showarticletitle{H2RDF: adaptive query processing on RDF data in the
  cloud.}. In \bibinfo{booktitle}{\emph{Proceedings of the 21st International
  Conference on World Wide Web}}. \bibinfo{pages}{397--400}.
\newblock


\bibitem[\protect\citeauthoryear{Patil, Kiran, Kavya, and Naresh~Patel}{Patil
  et~al\mbox{.}}{2018}]%
        {gdb_management_huge_unstr_data}
\bibfield{author}{\bibinfo{person}{N.S. Patil}, \bibinfo{person}{P Kiran},
  \bibinfo{person}{N.P. Kavya}, {and} \bibinfo{person}{K.M. Naresh~Patel}.}
  \bibinfo{year}{2018}\natexlab{}.
\newblock \showarticletitle{{A Survey on Graph Database Management Techniques
  for Huge Unstructured Data}}.
\newblock \bibinfo{journal}{\emph{{International Journal of Electrical and
  Computer Engineering}}} \bibinfo{volume}{81}, \bibinfo{number}{2}
  (\bibinfo{year}{2018}), \bibinfo{pages}{1140--1149}.
\newblock


\bibitem[\protect\citeauthoryear{Perozzi, Al-Rfou, and Skiena}{Perozzi
  et~al\mbox{.}}{2014}]%
        {perozzi2014deepwalk}
\bibfield{author}{\bibinfo{person}{Bryan Perozzi}, \bibinfo{person}{Rami
  Al-Rfou}, {and} \bibinfo{person}{Steven Skiena}.}
  \bibinfo{year}{2014}\natexlab{}.
\newblock \showarticletitle{Deepwalk: Online learning of social
  representations}. In \bibinfo{booktitle}{\emph{KDD}}.
  \bibinfo{pages}{701--710}.
\newblock


\bibitem[\protect\citeauthoryear{Petersen, Pedersen, et~al\mbox{.}}{Petersen
  et~al\mbox{.}}{2008}]%
        {petersen2008matrix}
\bibfield{author}{\bibinfo{person}{Kaare~Brandt Petersen},
  \bibinfo{person}{Michael~Syskind Pedersen}, {et~al\mbox{.}}}
  \bibinfo{year}{2008}\natexlab{}.
\newblock \showarticletitle{The matrix cookbook}.
\newblock \bibinfo{journal}{\emph{Technical University of Denmark}}
  \bibinfo{volume}{7}, \bibinfo{number}{15} (\bibinfo{year}{2008}),
  \bibinfo{pages}{510}.
\newblock


\bibitem[\protect\citeauthoryear{Pfaff, Fortunato, Sanchez-Gonzalez, and
  Battaglia}{Pfaff et~al\mbox{.}}{2020}]%
        {pfaff2020learning}
\bibfield{author}{\bibinfo{person}{Tobias Pfaff}, \bibinfo{person}{Meire
  Fortunato}, \bibinfo{person}{Alvaro Sanchez-Gonzalez}, {and}
  \bibinfo{person}{Peter~W Battaglia}.} \bibinfo{year}{2020}\natexlab{}.
\newblock \showarticletitle{Learning mesh-based simulation with graph
  networks}.
\newblock \bibinfo{journal}{\emph{arXiv preprint arXiv:2010.03409}}
  (\bibinfo{year}{2020}).
\newblock


\bibitem[\protect\citeauthoryear{Pokorny}{Pokorny}{2015}]%
        {pokorny2015graph}
\bibfield{author}{\bibinfo{person}{Jaroslav Pokorny}.}
  \bibinfo{year}{2015}\natexlab{}.
\newblock \showarticletitle{Graph databases: their power and limitations}. In
  \bibinfo{booktitle}{\emph{IFIP International Conference on Computer
  Information Systems and Industrial Management}}. Springer,
  \bibinfo{pages}{58--69}.
\newblock


\bibitem[\protect\citeauthoryear{Polikoff}{Polikoff}{2018}]%
        {tdan}
\bibfield{author}{\bibinfo{person}{Irene Polikoff}.}
  \bibinfo{year}{2018}\natexlab{}.
\newblock \bibinfo{title}{Knowledge Graphs vs. Property Graphs - Part I}.
\newblock \bibinfo{howpublished}{\newline Available at
  \url{https://tdan.com/knowledge-graphs-vs-property-graphs-part-1/27140}}.
\newblock


\bibitem[\protect\citeauthoryear{Portisch, Hladik, and Paulheim}{Portisch
  et~al\mbox{.}}{2020}]%
        {portisch2020rdf2vec}
\bibfield{author}{\bibinfo{person}{Jan Portisch}, \bibinfo{person}{Michael
  Hladik}, {and} \bibinfo{person}{Heiko Paulheim}.}
  \bibinfo{year}{2020}\natexlab{}.
\newblock \showarticletitle{RDF2Vec Light--A Lightweight Approach for Knowledge
  Graph Embeddings}.
\newblock \bibinfo{journal}{\emph{arXiv preprint arXiv:2009.07659}}
  (\bibinfo{year}{2020}).
\newblock


\bibitem[\protect\citeauthoryear{{Profium}}{{Profium}}{2018}]%
        {profium_sense_links}
\bibfield{author}{\bibinfo{person}{{Profium}}.}
  \bibinfo{year}{2018}\natexlab{}.
\newblock \bibinfo{title}{{Profium Sense}}.
\newblock \bibinfo{howpublished}{Available at
  \url{https://www.profium.com/en/}}.
\newblock


\bibitem[\protect\citeauthoryear{{Redis Labs}}{{Redis Labs}}{2018}]%
        {redisgraph_links}
\bibfield{author}{\bibinfo{person}{{Redis Labs}}.}
  \bibinfo{year}{2018}\natexlab{}.
\newblock \bibinfo{title}{{RedisGraph}}.
\newblock \bibinfo{howpublished}{Available at
  \url{https://oss.redislabs.com/redisgraph/}}.
\newblock


\bibitem[\protect\citeauthoryear{Reimers and Gurevych}{Reimers and
  Gurevych}{2019}]%
        {reimers2019sentence}
\bibfield{author}{\bibinfo{person}{Nils Reimers} {and} \bibinfo{person}{Iryna
  Gurevych}.} \bibinfo{year}{2019}\natexlab{}.
\newblock \showarticletitle{Sentence-bert: Sentence embeddings using siamese
  bert-networks}.
\newblock \bibinfo{journal}{\emph{arXiv preprint arXiv:1908.10084}}
  (\bibinfo{year}{2019}).
\newblock


\bibitem[\protect\citeauthoryear{Rickett, Haus, Maltby, and Maschhoff}{Rickett
  et~al\mbox{.}}{2018}]%
        {cge_paper}
\bibfield{author}{\bibinfo{person}{Christopher~D. Rickett},
  \bibinfo{person}{Utz-Uwe Haus}, \bibinfo{person}{James Maltby}, {and}
  \bibinfo{person}{Kristyn~J. Maschhoff}.} \bibinfo{year}{2018}\natexlab{}.
\newblock \showarticletitle{{Loading and Querying a Trillion RDF triples with
  Cray Graph Engine on the Cray XC}}. In \bibinfo{booktitle}{\emph{CUG}}. Cray
  Users Group.
\newblock


\bibitem[\protect\citeauthoryear{Ristoski and Paulheim}{Ristoski and
  Paulheim}{2016}]%
        {ristoski2016rdf2vec}
\bibfield{author}{\bibinfo{person}{Petar Ristoski} {and} \bibinfo{person}{Heiko
  Paulheim}.} \bibinfo{year}{2016}\natexlab{}.
\newblock \showarticletitle{Rdf2vec: Rdf graph embeddings for data mining}. In
  \bibinfo{booktitle}{\emph{International Semantic Web Conference}}. Springer,
  \bibinfo{pages}{498--514}.
\newblock


\bibitem[\protect\citeauthoryear{Ristoski, Rosati, Di~Noia, De~Leone, and
  Paulheim}{Ristoski et~al\mbox{.}}{2019}]%
        {ristoski2019rdf2vec}
\bibfield{author}{\bibinfo{person}{Petar Ristoski}, \bibinfo{person}{Jessica
  Rosati}, \bibinfo{person}{Tommaso Di~Noia}, \bibinfo{person}{Renato
  De~Leone}, {and} \bibinfo{person}{Heiko Paulheim}.}
  \bibinfo{year}{2019}\natexlab{}.
\newblock \showarticletitle{RDF2Vec: RDF graph embeddings and their
  applications}.
\newblock \bibinfo{journal}{\emph{Semantic Web}} \bibinfo{volume}{10},
  \bibinfo{number}{4} (\bibinfo{year}{2019}), \bibinfo{pages}{721--752}.
\newblock


\bibitem[\protect\citeauthoryear{Rossi, Ahmed, and Koh}{Rossi
  et~al\mbox{.}}{2018a}]%
        {rossiHigherorderNetworkRepresentation2018}
\bibfield{author}{\bibinfo{person}{Ryan~A. Rossi}, \bibinfo{person}{Nesreen~K.
  Ahmed}, {and} \bibinfo{person}{Eunyee Koh}.}
  \bibinfo{year}{2018}\natexlab{a}.
\newblock \showarticletitle{Higher-Order {{Network Representation Learning}}}.
  In \bibinfo{booktitle}{\emph{Companion {{Proceedings}} of the {{The Web
  Conference}} 2018}} \emph{(\bibinfo{series}{{{WWW}} '18})}.
  \bibinfo{publisher}{{International World Wide Web Conferences Steering
  Committee}}, \bibinfo{address}{{Republic and Canton of Geneva, CHE}},
  \bibinfo{pages}{3--4}.
\newblock
\showISBNx{978-1-4503-5640-4}
\urldef\tempurl%
\url{https://doi.org/10.1145/3184558.3186900}
\showDOI{\tempurl}


\bibitem[\protect\citeauthoryear{Rossi, Ahmed, Koh, Kim, Rao, and
  Yadkori}{Rossi et~al\mbox{.}}{2018b}]%
        {rossiHONEHigherOrderNetwork2018}
\bibfield{author}{\bibinfo{person}{Ryan~A. Rossi}, \bibinfo{person}{Nesreen~K.
  Ahmed}, \bibinfo{person}{Eunyee Koh}, \bibinfo{person}{Sungchul Kim},
  \bibinfo{person}{Anup Rao}, {and} \bibinfo{person}{Yasin~Abbasi Yadkori}.}
  \bibinfo{year}{2018}\natexlab{b}.
\newblock \showarticletitle{{{HONE}}: {{Higher-Order Network Embeddings}}}.
\newblock \bibinfo{journal}{\emph{arXiv:1801.09303 [cs, stat]}}
  (\bibinfo{date}{May} \bibinfo{year}{2018}).
\newblock
\showeprint[arxiv]{cs, stat/1801.09303}


\bibitem[\protect\citeauthoryear{Sakr et~al\mbox{.}}{Sakr
  et~al\mbox{.}}{2020}]%
        {sakr2020future}
\bibfield{author}{\bibinfo{person}{Sherif Sakr} {et~al\mbox{.}}}
  \bibinfo{year}{2020}\natexlab{}.
\newblock \showarticletitle{The Future is Big Graphs! A Community View on Graph
  Processing Systems}.
\newblock \bibinfo{journal}{\emph{arXiv preprint arXiv:2012.06171}}
  (\bibinfo{year}{2020}).
\newblock


\bibitem[\protect\citeauthoryear{Sanchez-Gonzalez, Godwin, Pfaff, Ying,
  Leskovec, and Battaglia}{Sanchez-Gonzalez et~al\mbox{.}}{2020}]%
        {sanchez2020learning}
\bibfield{author}{\bibinfo{person}{Alvaro Sanchez-Gonzalez},
  \bibinfo{person}{Jonathan Godwin}, \bibinfo{person}{Tobias Pfaff},
  \bibinfo{person}{Rex Ying}, \bibinfo{person}{Jure Leskovec}, {and}
  \bibinfo{person}{Peter Battaglia}.} \bibinfo{year}{2020}\natexlab{}.
\newblock \showarticletitle{Learning to simulate complex physics with graph
  networks}. In \bibinfo{booktitle}{\emph{ICML}}.
\newblock


\bibitem[\protect\citeauthoryear{Sato}{Sato}{2020}]%
        {sato2020survey}
\bibfield{author}{\bibinfo{person}{Ryoma Sato}.}
  \bibinfo{year}{2020}\natexlab{}.
\newblock \showarticletitle{A survey on the expressive power of graph neural
  networks}.
\newblock \bibinfo{journal}{\emph{arXiv preprint arXiv:2003.04078}}
  (\bibinfo{year}{2020}).
\newblock


\bibitem[\protect\citeauthoryear{Scarselli, Gori, Tsoi, Hagenbuchner, and
  Monfardini}{Scarselli et~al\mbox{.}}{2008}]%
        {scarselli2008graph}
\bibfield{author}{\bibinfo{person}{Franco Scarselli}, \bibinfo{person}{Marco
  Gori}, \bibinfo{person}{Ah~Chung Tsoi}, \bibinfo{person}{Markus
  Hagenbuchner}, {and} \bibinfo{person}{Gabriele Monfardini}.}
  \bibinfo{year}{2008}\natexlab{}.
\newblock \showarticletitle{The graph neural network model}.
\newblock \bibinfo{journal}{\emph{IEEE transactions on neural networks}}
  \bibinfo{volume}{20}, \bibinfo{number}{1} (\bibinfo{year}{2008}),
  \bibinfo{pages}{61--80}.
\newblock


\bibitem[\protect\citeauthoryear{Shi and Weninger}{Shi and Weninger}{2017}]%
        {shi2017proje}
\bibfield{author}{\bibinfo{person}{Baoxu Shi} {and} \bibinfo{person}{Tim
  Weninger}.} \bibinfo{year}{2017}\natexlab{}.
\newblock \showarticletitle{Proje: Embedding projection for knowledge graph
  completion}. In \bibinfo{booktitle}{\emph{Proceedings of the AAAI Conference
  on Artificial Intelligence}}, Vol.~\bibinfo{volume}{31}.
\newblock


\bibitem[\protect\citeauthoryear{Shi and Weninger}{Shi and Weninger}{2018}]%
        {shi2018open}
\bibfield{author}{\bibinfo{person}{Baoxu Shi} {and} \bibinfo{person}{Tim
  Weninger}.} \bibinfo{year}{2018}\natexlab{}.
\newblock \showarticletitle{Open-world knowledge graph completion}. In
  \bibinfo{booktitle}{\emph{Proceedings of the AAAI conference on artificial
  intelligence}}, Vol.~\bibinfo{volume}{32}.
\newblock


\bibitem[\protect\citeauthoryear{Solomonik, Besta, Vella, and
  Hoefler}{Solomonik et~al\mbox{.}}{2017}]%
        {solomonik2017scaling}
\bibfield{author}{\bibinfo{person}{Edgar Solomonik}, \bibinfo{person}{Maciej
  Besta}, \bibinfo{person}{Flavio Vella}, {and} \bibinfo{person}{Torsten
  Hoefler}.} \bibinfo{year}{2017}\natexlab{}.
\newblock \showarticletitle{Scaling betweenness centrality using
  communication-efficient sparse matrix multiplication}. In
  \bibinfo{booktitle}{\emph{ACM/IEEE Supercomputing}}.
\newblock


\bibitem[\protect\citeauthoryear{{Stardog Union}}{{Stardog Union}}{2018}]%
        {stardog_links}
\bibfield{author}{\bibinfo{person}{{Stardog Union}}.}
  \bibinfo{year}{2018}\natexlab{}.
\newblock \bibinfo{title}{{Stardog}}.
\newblock \bibinfo{howpublished}{Available at \url{https://www.stardog.com/}}.
\newblock


\bibitem[\protect\citeauthoryear{Sukhbaatar, Fergus, et~al\mbox{.}}{Sukhbaatar
  et~al\mbox{.}}{2016}]%
        {sukhbaatar2016learning}
\bibfield{author}{\bibinfo{person}{Sainbayar Sukhbaatar}, \bibinfo{person}{Rob
  Fergus}, {et~al\mbox{.}}} \bibinfo{year}{2016}\natexlab{}.
\newblock \showarticletitle{Learning multiagent communication with
  backpropagation}.
\newblock \bibinfo{journal}{\emph{NeurIPS}} (\bibinfo{year}{2016}).
\newblock


\bibitem[\protect\citeauthoryear{Sun, Vashishth, Sanyal, Talukdar, and
  Yang}{Sun et~al\mbox{.}}{2019}]%
        {sun2019re}
\bibfield{author}{\bibinfo{person}{Zhiqing Sun}, \bibinfo{person}{Shikhar
  Vashishth}, \bibinfo{person}{Soumya Sanyal}, \bibinfo{person}{Partha
  Talukdar}, {and} \bibinfo{person}{Yiming Yang}.}
  \bibinfo{year}{2019}\natexlab{}.
\newblock \showarticletitle{A re-evaluation of knowledge graph completion
  methods}.
\newblock \bibinfo{journal}{\emph{arXiv preprint arXiv:1911.03903}}
  (\bibinfo{year}{2019}).
\newblock


\bibitem[\protect\citeauthoryear{{The Apache Software Foundation}}{{The Apache
  Software Foundation}}{2021}]%
        {apache_jena_tbd_links}
\bibfield{author}{\bibinfo{person}{{The Apache Software Foundation}}.}
  \bibinfo{year}{2021}\natexlab{}.
\newblock \bibinfo{title}{{Apache Jena TBD}}.
\newblock \bibinfo{howpublished}{Available at
  \url{https://jena.apache.org/documentation/tdb/index.html}}.
\newblock


\bibitem[\protect\citeauthoryear{{The Linux Foundation}}{{The Linux
  Foundation}}{2018}]%
        {janus_graph_links}
\bibfield{author}{\bibinfo{person}{{The Linux Foundation}}.}
  \bibinfo{year}{2018}\natexlab{}.
\newblock \bibinfo{title}{{JanusGraph}}.
\newblock \bibinfo{howpublished}{Available at \url{http://janusgraph.org/}}.
\newblock


\bibitem[\protect\citeauthoryear{Thekumparampil, Wang, Oh, and
  Li}{Thekumparampil et~al\mbox{.}}{2018}]%
        {thekumparampil2018attention}
\bibfield{author}{\bibinfo{person}{Kiran~K Thekumparampil},
  \bibinfo{person}{Chong Wang}, \bibinfo{person}{Sewoong Oh}, {and}
  \bibinfo{person}{Li-Jia Li}.} \bibinfo{year}{2018}\natexlab{}.
\newblock \showarticletitle{Attention-based graph neural network for
  semi-supervised learning}.
\newblock \bibinfo{journal}{\emph{arXiv preprint arXiv:1803.03735}}
  (\bibinfo{year}{2018}).
\newblock


\bibitem[\protect\citeauthoryear{Thorup}{Thorup}{2000}]%
        {thorup2000near}
\bibfield{author}{\bibinfo{person}{Mikkel Thorup}.}
  \bibinfo{year}{2000}\natexlab{}.
\newblock \showarticletitle{Near-optimal fully-dynamic graph connectivity}. In
  \bibinfo{booktitle}{\emph{Proceedings of the thirty-second annual ACM
  symposium on Theory of computing}}. \bibinfo{pages}{343--350}.
\newblock


\bibitem[\protect\citeauthoryear{{TigerGraph}}{{TigerGraph}}{2018}]%
        {tiger_graph_links}
\bibfield{author}{\bibinfo{person}{{TigerGraph}}.}
  \bibinfo{year}{2018}\natexlab{}.
\newblock \bibinfo{title}{{TigerGraph}}.
\newblock \bibinfo{howpublished}{Available at
  \url{https://www.tigergraph.com/}}.
\newblock


\bibitem[\protect\citeauthoryear{Toader, Uta, Musaafir, and Iosup}{Toader
  et~al\mbox{.}}{2019}]%
        {toader2019graphless}
\bibfield{author}{\bibinfo{person}{Lucian Toader}, \bibinfo{person}{Alexandru
  Uta}, \bibinfo{person}{Ahmed Musaafir}, {and} \bibinfo{person}{Alexandru
  Iosup}.} \bibinfo{year}{2019}\natexlab{}.
\newblock \showarticletitle{Graphless: Toward serverless graph processing}. In
  \bibinfo{booktitle}{\emph{2019 18th International Symposium on Parallel and
  Distributed Computing (ISPDC)}}. IEEE, \bibinfo{pages}{66--73}.
\newblock


\bibitem[\protect\citeauthoryear{Tripathy, Yelick, and Bulu{\c{c}}}{Tripathy
  et~al\mbox{.}}{2020}]%
        {tripathy2020reducing}
\bibfield{author}{\bibinfo{person}{Alok Tripathy}, \bibinfo{person}{Katherine
  Yelick}, {and} \bibinfo{person}{Ayd{\i}n Bulu{\c{c}}}.}
  \bibinfo{year}{2020}\natexlab{}.
\newblock \showarticletitle{Reducing communication in graph neural network
  training}. In \bibinfo{booktitle}{\emph{ACM/IEEE Supercomputing}}.
\newblock


\bibitem[\protect\citeauthoryear{Trouillon, Dance, Welbl, Riedel, Gaussier, and
  Bouchard}{Trouillon et~al\mbox{.}}{2017}]%
        {trouillon2017knowledge}
\bibfield{author}{\bibinfo{person}{Th{\'e}o Trouillon},
  \bibinfo{person}{Christopher~R Dance}, \bibinfo{person}{Johannes Welbl},
  \bibinfo{person}{Sebastian Riedel}, \bibinfo{person}{{\'E}ric Gaussier},
  {and} \bibinfo{person}{Guillaume Bouchard}.} \bibinfo{year}{2017}\natexlab{}.
\newblock \showarticletitle{Knowledge graph completion via complex tensor
  factorization}.
\newblock \bibinfo{journal}{\emph{arXiv preprint arXiv:1702.06879}}
  (\bibinfo{year}{2017}).
\newblock


\bibitem[\protect\citeauthoryear{Urbani, Dutta, Gurajada, and Weikum}{Urbani
  et~al\mbox{.}}{2016}]%
        {urbani2016kognac}
\bibfield{author}{\bibinfo{person}{Jacopo Urbani}, \bibinfo{person}{Sourav
  Dutta}, \bibinfo{person}{Sairam Gurajada}, {and} \bibinfo{person}{Gerhard
  Weikum}.} \bibinfo{year}{2016}\natexlab{}.
\newblock \showarticletitle{{KOGNAC: efficient encoding of large knowledge
  graphs}}.
\newblock \bibinfo{journal}{\emph{arXiv preprint arXiv:1604.04795}}
  (\bibinfo{year}{2016}).
\newblock


\bibitem[\protect\citeauthoryear{Veli{\v{c}}kovi{\'c}, Cucurull, Casanova,
  Romero, Lio, and Bengio}{Veli{\v{c}}kovi{\'c} et~al\mbox{.}}{2017}]%
        {velivckovic2017graph}
\bibfield{author}{\bibinfo{person}{Petar Veli{\v{c}}kovi{\'c}},
  \bibinfo{person}{Guillem Cucurull}, \bibinfo{person}{Arantxa Casanova},
  \bibinfo{person}{Adriana Romero}, \bibinfo{person}{Pietro Lio}, {and}
  \bibinfo{person}{Yoshua Bengio}.} \bibinfo{year}{2017}\natexlab{}.
\newblock \showarticletitle{Graph attention networks}.
\newblock \bibinfo{journal}{\emph{arXiv preprint arXiv:1710.10903}}
  (\bibinfo{year}{2017}).
\newblock


\bibitem[\protect\citeauthoryear{Waleffe, Mohoney, Rekatsinas, and
  Venkataraman}{Waleffe et~al\mbox{.}}{2022}]%
        {waleffe2022marius++}
\bibfield{author}{\bibinfo{person}{Roger Waleffe}, \bibinfo{person}{Jason
  Mohoney}, \bibinfo{person}{Theodoros Rekatsinas}, {and}
  \bibinfo{person}{Shivaram Venkataraman}.} \bibinfo{year}{2022}\natexlab{}.
\newblock \showarticletitle{Marius++: Large-Scale Training of Graph Neural
  Networks on a Single Machine}.
\newblock \bibinfo{journal}{\emph{arXiv preprint arXiv:2202.02365}}
  (\bibinfo{year}{2022}).
\newblock


\bibitem[\protect\citeauthoryear{Wan, Li, Li, Kim, and Lin}{Wan
  et~al\mbox{.}}{2022a}]%
        {wan2022bns}
\bibfield{author}{\bibinfo{person}{Cheng Wan}, \bibinfo{person}{Youjie Li},
  \bibinfo{person}{Ang Li}, \bibinfo{person}{Nam~Sung Kim}, {and}
  \bibinfo{person}{Yingyan Lin}.} \bibinfo{year}{2022}\natexlab{a}.
\newblock \showarticletitle{BNS-GCN: Efficient Full-Graph Training of Graph
  Convolutional Networks with Partition-Parallelism and Random Boundary Node
  Sampling Sampling}.
\newblock \bibinfo{journal}{\emph{MLSys}} (\bibinfo{year}{2022}).
\newblock


\bibitem[\protect\citeauthoryear{Wan, Li, Wolfe, Kyrillidis, Kim, and Lin}{Wan
  et~al\mbox{.}}{2022b}]%
        {wan2022pipegcn}
\bibfield{author}{\bibinfo{person}{Cheng Wan}, \bibinfo{person}{Youjie Li},
  \bibinfo{person}{Cameron~R Wolfe}, \bibinfo{person}{Anastasios Kyrillidis},
  \bibinfo{person}{Nam~Sung Kim}, {and} \bibinfo{person}{Yingyan Lin}.}
  \bibinfo{year}{2022}\natexlab{b}.
\newblock \showarticletitle{PipeGCN: Efficient full-graph training of graph
  convolutional networks with pipelined feature communication}.
\newblock \bibinfo{journal}{\emph{arXiv preprint arXiv:2203.10428}}
  (\bibinfo{year}{2022}).
\newblock


\bibitem[\protect\citeauthoryear{Wang, Cui, and Zhu}{Wang
  et~al\mbox{.}}{2016}]%
        {wang2016structural}
\bibfield{author}{\bibinfo{person}{Daixin Wang}, \bibinfo{person}{Peng Cui},
  {and} \bibinfo{person}{Wenwu Zhu}.} \bibinfo{year}{2016}\natexlab{}.
\newblock \showarticletitle{Structural deep network embedding}. In
  \bibinfo{booktitle}{\emph{ACM KDD}}.
\newblock


\bibitem[\protect\citeauthoryear{Wang, Zheng, Ye, Gan, Li, Song, Zhou, Ma, Yu,
  Gai, et~al\mbox{.}}{Wang et~al\mbox{.}}{2019c}]%
        {wang2019deep}
\bibfield{author}{\bibinfo{person}{Minjie Wang}, \bibinfo{person}{Da Zheng},
  \bibinfo{person}{Zihao Ye}, \bibinfo{person}{Quan Gan},
  \bibinfo{person}{Mufei Li}, \bibinfo{person}{Xiang Song},
  \bibinfo{person}{Jinjing Zhou}, \bibinfo{person}{Chao Ma},
  \bibinfo{person}{Lingfan Yu}, \bibinfo{person}{Yu Gai}, {et~al\mbox{.}}}
  \bibinfo{year}{2019}\natexlab{c}.
\newblock \showarticletitle{Deep graph library: A graph-centric,
  highly-performant package for graph neural networks}.
\newblock \bibinfo{journal}{\emph{arXiv:1909.01315}} (\bibinfo{year}{2019}).
\newblock


\bibitem[\protect\citeauthoryear{Wang, Mao, Wang, and Guo}{Wang
  et~al\mbox{.}}{2017}]%
        {wang2017knowledge}
\bibfield{author}{\bibinfo{person}{Quan Wang}, \bibinfo{person}{Zhendong Mao},
  \bibinfo{person}{Bin Wang}, {and} \bibinfo{person}{Li Guo}.}
  \bibinfo{year}{2017}\natexlab{}.
\newblock \showarticletitle{Knowledge graph embedding: A survey of approaches
  and applications}.
\newblock \bibinfo{journal}{\emph{IEEE TKDE}} (\bibinfo{year}{2017}).
\newblock


\bibitem[\protect\citeauthoryear{Wang, Bo, Shi, Fan, Ye, and Yu}{Wang
  et~al\mbox{.}}{2020a}]%
        {wang2020survey}
\bibfield{author}{\bibinfo{person}{Xiao Wang}, \bibinfo{person}{Deyu Bo},
  \bibinfo{person}{Chuan Shi}, \bibinfo{person}{Shaohua Fan},
  \bibinfo{person}{Yanfang Ye}, {and} \bibinfo{person}{Philip~S Yu}.}
  \bibinfo{year}{2020}\natexlab{a}.
\newblock \showarticletitle{A survey on heterogeneous graph embedding: methods,
  techniques, applications and sources}.
\newblock \bibinfo{journal}{\emph{arXiv:2011.14867}} (\bibinfo{year}{2020}).
\newblock


\bibitem[\protect\citeauthoryear{Wang, Ji, Shi, Wang, Ye, Cui, and Yu}{Wang
  et~al\mbox{.}}{2019a}]%
        {wang2019heterogeneous}
\bibfield{author}{\bibinfo{person}{Xiao Wang}, \bibinfo{person}{Houye Ji},
  \bibinfo{person}{Chuan Shi}, \bibinfo{person}{Bai Wang},
  \bibinfo{person}{Yanfang Ye}, \bibinfo{person}{Peng Cui}, {and}
  \bibinfo{person}{Philip~S Yu}.} \bibinfo{year}{2019}\natexlab{a}.
\newblock \showarticletitle{Heterogeneous graph attention network}. In
  \bibinfo{booktitle}{\emph{The world wide web conference}}.
  \bibinfo{pages}{2022--2032}.
\newblock


\bibitem[\protect\citeauthoryear{Wang, Feng, Li, Li, Deng, Xie, and Ding}{Wang
  et~al\mbox{.}}{2020b}]%
        {wang2020gnnadvisor}
\bibfield{author}{\bibinfo{person}{Yuke Wang}, \bibinfo{person}{Boyuan Feng},
  \bibinfo{person}{Gushu Li}, \bibinfo{person}{Shuangchen Li},
  \bibinfo{person}{Lei Deng}, \bibinfo{person}{Yuan Xie}, {and}
  \bibinfo{person}{Yufei Ding}.} \bibinfo{year}{2020}\natexlab{b}.
\newblock \showarticletitle{GNNAdvisor: An Efficient Runtime System for GNN
  Acceleration on GPUs}.
\newblock \bibinfo{journal}{\emph{arXiv preprint arXiv:2006.06608}}
  (\bibinfo{year}{2020}).
\newblock


\bibitem[\protect\citeauthoryear{Wang, Sun, Liu, Sarma, Bronstein, and
  Solomon}{Wang et~al\mbox{.}}{2019b}]%
        {wang2019dynamic}
\bibfield{author}{\bibinfo{person}{Yue Wang}, \bibinfo{person}{Yongbin Sun},
  \bibinfo{person}{Ziwei Liu}, \bibinfo{person}{Sanjay~E Sarma},
  \bibinfo{person}{Michael~M Bronstein}, {and} \bibinfo{person}{Justin~M
  Solomon}.} \bibinfo{year}{2019}\natexlab{b}.
\newblock \showarticletitle{Dynamic graph cnn for learning on point clouds}.
\newblock \bibinfo{journal}{\emph{Acm Transactions On Graphics (tog)}}
  \bibinfo{volume}{38}, \bibinfo{number}{5} (\bibinfo{year}{2019}),
  \bibinfo{pages}{1--12}.
\newblock


\bibitem[\protect\citeauthoryear{Wu, Souza, Zhang, Fifty, Yu, and
  Weinberger}{Wu et~al\mbox{.}}{2019}]%
        {wu2019simplifying}
\bibfield{author}{\bibinfo{person}{Felix Wu}, \bibinfo{person}{Amauri Souza},
  \bibinfo{person}{Tianyi Zhang}, \bibinfo{person}{Christopher Fifty},
  \bibinfo{person}{Tao Yu}, {and} \bibinfo{person}{Kilian Weinberger}.}
  \bibinfo{year}{2019}\natexlab{}.
\newblock \showarticletitle{Simplifying graph convolutional networks}. In
  \bibinfo{booktitle}{\emph{International conference on machine learning}}.
  PMLR, \bibinfo{pages}{6861--6871}.
\newblock


\bibitem[\protect\citeauthoryear{Wu, Sun, Zhang, and Cui}{Wu
  et~al\mbox{.}}{2020b}]%
        {wu2020graph}
\bibfield{author}{\bibinfo{person}{Shiwen Wu}, \bibinfo{person}{Fei Sun},
  \bibinfo{person}{Wentao Zhang}, {and} \bibinfo{person}{Bin Cui}.}
  \bibinfo{year}{2020}\natexlab{b}.
\newblock \showarticletitle{Graph neural networks in recommender systems: a
  survey}.
\newblock \bibinfo{journal}{\emph{arXiv preprint arXiv:2011.02260}}
  (\bibinfo{year}{2020}).
\newblock


\bibitem[\protect\citeauthoryear{Wu, Ma, Cai, Jin, Li, Zheng, Cheng, and Yu}{Wu
  et~al\mbox{.}}{2021}]%
        {wu2021seastar}
\bibfield{author}{\bibinfo{person}{Yidi Wu}, \bibinfo{person}{Kaihao Ma},
  \bibinfo{person}{Zhenkun Cai}, \bibinfo{person}{Tatiana Jin},
  \bibinfo{person}{Boyang Li}, \bibinfo{person}{Chenguang Zheng},
  \bibinfo{person}{James Cheng}, {and} \bibinfo{person}{Fan Yu}.}
  \bibinfo{year}{2021}\natexlab{}.
\newblock \showarticletitle{Seastar: vertex-centric programming for graph
  neural networks}. In \bibinfo{booktitle}{\emph{EuroSys}}.
\newblock


\bibitem[\protect\citeauthoryear{Wu et~al\mbox{.}}{Wu et~al\mbox{.}}{2020a}]%
        {wu2020comprehensive}
\bibfield{author}{\bibinfo{person}{Zonghan Wu} {et~al\mbox{.}}}
  \bibinfo{year}{2020}\natexlab{a}.
\newblock \showarticletitle{A comprehensive survey on graph neural networks}.
\newblock \bibinfo{journal}{\emph{IEEE Transactions on Neural Networks and
  Learning Systems}} (\bibinfo{year}{2020}).
\newblock


\bibitem[\protect\citeauthoryear{Xie, Yu, Lv, Zhang, Wang, and Gong}{Xie
  et~al\mbox{.}}{2021}]%
        {xie2021survey}
\bibfield{author}{\bibinfo{person}{Yu Xie}, \bibinfo{person}{Bin Yu},
  \bibinfo{person}{Shengze Lv}, \bibinfo{person}{Chen Zhang},
  \bibinfo{person}{Guodong Wang}, {and} \bibinfo{person}{Maoguo Gong}.}
  \bibinfo{year}{2021}\natexlab{}.
\newblock \showarticletitle{A survey on heterogeneous network representation
  learning}.
\newblock \bibinfo{journal}{\emph{Pattern Recognition}}  \bibinfo{volume}{116}
  (\bibinfo{year}{2021}), \bibinfo{pages}{107936}.
\newblock


\bibitem[\protect\citeauthoryear{Xu, Hu, Leskovec, and Jegelka}{Xu
  et~al\mbox{.}}{2018}]%
        {xu2018powerful}
\bibfield{author}{\bibinfo{person}{Keyulu Xu}, \bibinfo{person}{Weihua Hu},
  \bibinfo{person}{Jure Leskovec}, {and} \bibinfo{person}{Stefanie Jegelka}.}
  \bibinfo{year}{2018}\natexlab{}.
\newblock \showarticletitle{How powerful are graph neural networks?}
\newblock \bibinfo{journal}{\emph{arXiv preprint arXiv:1810.00826}}
  (\bibinfo{year}{2018}).
\newblock


\bibitem[\protect\citeauthoryear{Yan, Deng, Hu, Liang, Feng, Ye, Zhang, Fan,
  and Xie}{Yan et~al\mbox{.}}{2020}]%
        {yan2020hygcn}
\bibfield{author}{\bibinfo{person}{Mingyu Yan}, \bibinfo{person}{Lei Deng},
  \bibinfo{person}{Xing Hu}, \bibinfo{person}{Ling Liang},
  \bibinfo{person}{Yujing Feng}, \bibinfo{person}{Xiaochun Ye},
  \bibinfo{person}{Zhimin Zhang}, \bibinfo{person}{Dongrui Fan}, {and}
  \bibinfo{person}{Yuan Xie}.} \bibinfo{year}{2020}\natexlab{}.
\newblock \showarticletitle{Hygcn: A gcn accelerator with hybrid architecture}.
  In \bibinfo{booktitle}{\emph{IEEE HPCA}}. IEEE, \bibinfo{pages}{15--29}.
\newblock


\bibitem[\protect\citeauthoryear{Yang, Xiao, Zhang, Sun, and Han}{Yang
  et~al\mbox{.}}{2020}]%
        {yang2020heterogeneous}
\bibfield{author}{\bibinfo{person}{Carl Yang}, \bibinfo{person}{Yuxin Xiao},
  \bibinfo{person}{Yu Zhang}, \bibinfo{person}{Yizhou Sun}, {and}
  \bibinfo{person}{Jiawei Han}.} \bibinfo{year}{2020}\natexlab{}.
\newblock \showarticletitle{Heterogeneous network representation learning: A
  unified framework with survey and benchmark}.
\newblock \bibinfo{journal}{\emph{IEEE TKDE}} (\bibinfo{year}{2020}).
\newblock


\bibitem[\protect\citeauthoryear{Yao, Mao, and Luo}{Yao et~al\mbox{.}}{2019}]%
        {yao2019kg}
\bibfield{author}{\bibinfo{person}{Liang Yao}, \bibinfo{person}{Chengsheng
  Mao}, {and} \bibinfo{person}{Yuan Luo}.} \bibinfo{year}{2019}\natexlab{}.
\newblock \showarticletitle{KG-BERT: BERT for knowledge graph completion}.
\newblock \bibinfo{journal}{\emph{arXiv preprint arXiv:1909.03193}}
  (\bibinfo{year}{2019}).
\newblock


\bibitem[\protect\citeauthoryear{Ying, He, Chen, Eksombatchai, Hamilton, and
  Leskovec}{Ying et~al\mbox{.}}{2018}]%
        {ying2018graph}
\bibfield{author}{\bibinfo{person}{Rex Ying}, \bibinfo{person}{Ruining He},
  \bibinfo{person}{Kaifeng Chen}, \bibinfo{person}{Pong Eksombatchai},
  \bibinfo{person}{William~L Hamilton}, {and} \bibinfo{person}{Jure Leskovec}.}
  \bibinfo{year}{2018}\natexlab{}.
\newblock \showarticletitle{Graph convolutional neural networks for web-scale
  recommender systems}. In \bibinfo{booktitle}{\emph{ACM KDD}}.
\newblock


\bibitem[\protect\citeauthoryear{You, Ying, and Leskovec}{You
  et~al\mbox{.}}{2020}]%
        {you2020design}
\bibfield{author}{\bibinfo{person}{Jiaxuan You}, \bibinfo{person}{Zhitao Ying},
  {and} \bibinfo{person}{Jure Leskovec}.} \bibinfo{year}{2020}\natexlab{}.
\newblock \showarticletitle{Design space for graph neural networks}.
\newblock \bibinfo{journal}{\emph{Advances in Neural Information Processing
  Systems}}  \bibinfo{volume}{33} (\bibinfo{year}{2020}),
  \bibinfo{pages}{17009--17021}.
\newblock


\bibitem[\protect\citeauthoryear{Yuan, Liu, Wu, Jin, Zhang, and Liu}{Yuan
  et~al\mbox{.}}{2013}]%
        {triplebit_links}
\bibfield{author}{\bibinfo{person}{Pingpeng Yuan}, \bibinfo{person}{Pu Liu},
  \bibinfo{person}{Buwen Wu}, \bibinfo{person}{Hai Jin}, \bibinfo{person}{Wenya
  Zhang}, {and} \bibinfo{person}{Ling Liu}.} \bibinfo{year}{2013}\natexlab{}.
\newblock \showarticletitle{{TripleBit: a fast and compact system for large
  scale RDF data}}.
\newblock \bibinfo{journal}{\emph{Proceedings of the VLDB Endowment}}
  \bibinfo{volume}{6}, \bibinfo{number}{7} (\bibinfo{year}{2013}),
  \bibinfo{pages}{517--528}.
\newblock
\showISSN{2150-8097}
\urldef\tempurl%
\url{https://doi.org/10.14778/2536349.2536352}
\showDOI{\tempurl}


\bibitem[\protect\citeauthoryear{Zeng, Zhou, Srivastava, Kannan, and
  Prasanna}{Zeng et~al\mbox{.}}{2019}]%
        {zeng2019graphsaint}
\bibfield{author}{\bibinfo{person}{Hanqing Zeng}, \bibinfo{person}{Hongkuan
  Zhou}, \bibinfo{person}{Ajitesh Srivastava}, \bibinfo{person}{Rajgopal
  Kannan}, {and} \bibinfo{person}{Viktor Prasanna}.}
  \bibinfo{year}{2019}\natexlab{}.
\newblock \showarticletitle{Graphsaint: Graph sampling based inductive learning
  method}.
\newblock \bibinfo{journal}{\emph{arXiv preprint arXiv:1907.04931}}
  (\bibinfo{year}{2019}).
\newblock


\bibitem[\protect\citeauthoryear{Zhang, Song, Huang, Swami, and Chawla}{Zhang
  et~al\mbox{.}}{2019}]%
        {zhang2019heterogeneous}
\bibfield{author}{\bibinfo{person}{Chuxu Zhang}, \bibinfo{person}{Dongjin
  Song}, \bibinfo{person}{Chao Huang}, \bibinfo{person}{Ananthram Swami}, {and}
  \bibinfo{person}{Nitesh~V Chawla}.} \bibinfo{year}{2019}\natexlab{}.
\newblock \showarticletitle{Heterogeneous graph neural network}. In
  \bibinfo{booktitle}{\emph{KDD}}. \bibinfo{pages}{793--803}.
\newblock


\bibitem[\protect\citeauthoryear{Zhang et~al\mbox{.}}{Zhang
  et~al\mbox{.}}{2020b}]%
        {zhang2020agl}
\bibfield{author}{\bibinfo{person}{Dalong Zhang} {et~al\mbox{.}}}
  \bibinfo{year}{2020}\natexlab{b}.
\newblock \showarticletitle{Agl: a scalable system for industrial-purpose graph
  machine learning}.
\newblock \bibinfo{journal}{\emph{arXiv preprint arXiv:2003.02454}}
  (\bibinfo{year}{2020}).
\newblock


\bibitem[\protect\citeauthoryear{Zhang, Cui, and Zhu}{Zhang
  et~al\mbox{.}}{2020a}]%
        {zhang2020deep}
\bibfield{author}{\bibinfo{person}{Ziwei Zhang}, \bibinfo{person}{Peng Cui},
  {and} \bibinfo{person}{Wenwu Zhu}.} \bibinfo{year}{2020}\natexlab{a}.
\newblock \showarticletitle{Deep learning on graphs: A survey}.
\newblock \bibinfo{journal}{\emph{IEEE Transactions on Knowledge and Data
  Engineering}} (\bibinfo{year}{2020}).
\newblock


\bibitem[\protect\citeauthoryear{Zheng, Song, Yang, LaSalle, Su, Wang, Ma, and
  Karypis}{Zheng et~al\mbox{.}}{2021}]%
        {zheng2021distributed}
\bibfield{author}{\bibinfo{person}{Da Zheng}, \bibinfo{person}{Xiang Song},
  \bibinfo{person}{Chengru Yang}, \bibinfo{person}{Dominique LaSalle},
  \bibinfo{person}{Qidong Su}, \bibinfo{person}{Minjie Wang},
  \bibinfo{person}{Chao Ma}, {and} \bibinfo{person}{George Karypis}.}
  \bibinfo{year}{2021}\natexlab{}.
\newblock \showarticletitle{Distributed Hybrid CPU and GPU training for Graph
  Neural Networks on Billion-Scale Graphs}.
\newblock \bibinfo{journal}{\emph{arXiv:2112.15345}} (\bibinfo{year}{2021}).
\newblock


\bibitem[\protect\citeauthoryear{Zhou et~al\mbox{.}}{Zhou
  et~al\mbox{.}}{2020}]%
        {zhou2020graph}
\bibfield{author}{\bibinfo{person}{Jie Zhou} {et~al\mbox{.}}}
  \bibinfo{year}{2020}\natexlab{}.
\newblock \showarticletitle{Graph neural networks: A review of methods and
  applications}.
\newblock \bibinfo{journal}{\emph{AI Open}}  \bibinfo{volume}{1}
  (\bibinfo{year}{2020}), \bibinfo{pages}{57--81}.
\newblock


\bibitem[\protect\citeauthoryear{Zhu et~al\mbox{.}}{Zhu et~al\mbox{.}}{2019}]%
        {zhu2019aligraph}
\bibfield{author}{\bibinfo{person}{Rong Zhu} {et~al\mbox{.}}}
  \bibinfo{year}{2019}\natexlab{}.
\newblock \showarticletitle{Aligraph: A comprehensive graph neural network
  platform}.
\newblock \bibinfo{journal}{\emph{arXiv preprint arXiv:1902.08730}}
  (\bibinfo{year}{2019}).
\newblock


\bibitem[\protect\citeauthoryear{Zhu et~al\mbox{.}}{Zhu et~al\mbox{.}}{2020}]%
        {livegraph}
\bibfield{author}{\bibinfo{person}{Xiaowei Zhu} {et~al\mbox{.}}}
  \bibinfo{year}{2020}\natexlab{}.
\newblock \showarticletitle{{LiveGraph: A Transactional Graph Storage System
  with Purely Sequential Adjacency List Scans}}.
\newblock \bibinfo{journal}{\emph{VLDB}} \bibinfo{volume}{13},
  \bibinfo{number}{7} (\bibinfo{year}{2020}), \bibinfo{pages}{1020--1034}.
\newblock
\showISSN{2150-8097}
\urldef\tempurl%
\url{https://doi.org/10.14778/3384345.3384351}
\showDOI{\tempurl}


\bibitem[\protect\citeauthoryear{Zou, \"{O}zsu, Chen, Shen, Huang, and
  Zhao}{Zou et~al\mbox{.}}{2014}]%
        {gstore}
\bibfield{author}{\bibinfo{person}{Lei Zou}, \bibinfo{person}{M.~Tamer
  \"{O}zsu}, \bibinfo{person}{Lei Chen}, \bibinfo{person}{Xuchuan Shen},
  \bibinfo{person}{Ruizhe Huang}, {and} \bibinfo{person}{Dongyan Zhao}.}
  \bibinfo{year}{2014}\natexlab{}.
\newblock \showarticletitle{{GStore: A Graph-Based SPARQL Query Engine}}.
\newblock \bibinfo{journal}{\emph{VLDB Journal}} \bibinfo{volume}{23},
  \bibinfo{number}{4} (\bibinfo{year}{2014}), \bibinfo{pages}{565--590}.
\newblock
\showISSN{1066-8888}
\urldef\tempurl%
\url{https://doi.org/10.1007/s00778-013-0337-7}
\showDOI{\tempurl}


\end{thebibliography}
}



\appendix

\section*{Appendix}

\section{Dataset Specification}

We present the details about the used datasets.

\subsection{MAKG}

The dataset \textit{MAKG} (small) consists of $3'066'782$ vertices and
$12'314'398$ edges.  Each vertex is labeled with either \textit{author}
($55\%$), \textit{paper} ($44\%$), \textit{affiliation} ($<1\%$),
\textit{conferenceseries} ($<1\%$), \textit{conferenceinstance} ($<1\%$),
\textit{fieldofstudy} ($<1\%$), or \textit{journal} ($<1\%$). Vertices with the
label \textit{paper} are further subdivided into \textit{book},
\textit{bookchapter}, \textit{conferencepaper}, \textit{journalpaper},
\textit{patentdocument} or \textit{others}.  Vertices labeled with
\textit{affiliation}, \textit{author}, \textit{conferenceseries},
\textit{conferenceinstance}, \textit{fieldofstudy} and \textit{journal} do all
have the properties \textit{rank}, \textit{name}, \textit{papercount},
\textit{citationcount}, and \textit{created}. Some of them have additional
properties, e.g., \textit{homepage}. All vertices with the label \textit{paper}
have the properties \textit{rank}, \textit{citationcount}, \textit{created},
\textit{title}, \textit{publicationdate}, \textit{referencecount}, and
\textit{estimatedcitationcount}. Some of them have the additional properties
\textit{publisher}, \textit{volume}, \textit{issueidentifier},
\textit{startingpage}, \textit{endingpage}, or \textit{doi}.  Edges do not have
properties but each edge has a label which is either \textit{cites} ($40\%$),
\textit{creator} ($36\%$), \textit{hasdiscipline} ($11\%$),
\textit{apreasinjournal} ($6\%$), \textit{memberof} ($5.7\%$),
\textit{appearsinconferenceinstance} ($<1\%$),
\textit{appearsinconferenceseries} ($<1\%$), or \textit{ispartof} ($<1\%$).

\iftr
\subsection{IMDB}

The dataset \textit{IMDB} consists of $13'148'047$ vertices and $23'243'638$ edges. In this 
dataset, each vertex can have multiple labels. There are a total of $53$ different labels.
The most frequent labels are \textit{person} ($90\%$), \textit{actor} ($21\%$), 
\textit{actress} ($13\%$), \textit{motionpicture} ($10\%$), \textit{miscellaneous} ($9\%$), 
 \textit{producer} ($8\%$), and \textit{director} ($7\%$). The remaining labels are either 
other jobs, departments of a movie production or a more fine-grained classification of movies.
All vertices labeled with \textit{motionpicture} have properties \textit{primarytitle},
 \textit{originaltitle}, \textit{isadult}, \textit{averagerating} and \textit{numvotes}.
Most of them have the additional properties \textit{startyear}, \textit{runtimeminutes}, and 
 \textit{genres} (a list containing multiple genres per movie). All vertices with the
label \textit{person} have the property \textit{primaryname}, a small subset of them have
the property \textit{birthyear} and an even smaller subset the property \textit{deathyear}.
Each edge has only one of a total of $14$ possible labels. The most frequent labels are
\textit{knownfortitle} ($52\%$), \textit{actor} ($12\%$), \textit{actress} ($7\%$),
and \textit{writer} ($7\%$). Most edges do not have any properties but $21\%$ of them have
the property \textit{character}, $10\%$ of them have the property \textit{job} and only
$2.5\%$ of the have the properties \textit{seasonnumber} or \textit{episodenumber}.

\subsection{BALB}

The \textit{BALB} dataset contains $3'538'495$ vertices and $5'409'812$ edges.
Vertices do not have labels, but they all have the following properties:
\textit{posx}, \textit{posy}, \textit{posz} (describing a position in the 3D
space), \textit{degree} (number of edges attached to a given vertex), and
\textit{isatsampleborder} which contains the value \texttt{true} for $9$
vertices and \texttt{false} for the remaining ones. Edges are not labeled
either, but all of them contain the properties \textit{length},
\textit{distance}, \textit{curveness}, \textit{colume},
\textit{avgcrosssection}, \textit{minradiusavg}, \textit{minradiusstd},
\textit{avgradiusavg}, \textit{avgradiusstd}, \textit{maxradiusavg},
\textit{maxradiusstd}, \textit{roundnessavg}, \textit{roundnessstd},
\textit{node1degree}, \textit{node2degree}, \textit{numvoxels}, and
\textit{hasnodeatsampleborder}.
\fi

\subsection{Citations}

The \textit{citations} dataset contains $132'259$ vertices and $221'237$ edges;
we use it mostly for debugging purposes.  Each vertex has one label which is
either \textit{author} ($61\%$), \textit{article} ($39\%$), or \textit{venue}
($<1\%$).  All \textit{article}-vertices have the properties \textit{index} (a
32-digit HEX number), \textit{title} and \textit{year} (the year in which the
article was published).  $85\%$ of the articles have the property
\textit{abstract} and $72\%$ of them have the property \textit{ncitations} (the
article's citation count).  Vertices labeled with \textit{author} or
\textit{venue} have only one property called \textit{name}. Edges do not have
properties but each edge has a label which is either \textit{author} ($64\%$),
\textit{venue} ($23\%$), or \textit{cited} ($13\%$).

\subsection{Twitter}

The dataset \textit{TwitterTrolls} contains $281'136$ vertices and $493'160$ edges.
Vertices are labeled with \textit{tweet} ($82\%$), \textit{url} ($8\%$), 
\textit{hashtag} ($5\%$), \textit{user} ($5\%$), \textit{trolluser} ($<1\%$), or 
\textit{source} ($<1\%$). Vertices wit labels \textit{hashtag}, \textit{source}, 
\textit{user}, and \textit{url} have a single property each, namely \textit{tag},  
\textit{name}, \textit{userkey}, and \textit{expandedurl} respectively. 
Vertices labeled with \textit{trolluser} do all have the properties \textit{sourcename} and
\textit{userkey}. Most of them ($>80\%$) have additional properties
\textit{lang} (language), \textit{verified} (true or false),
\textit{name}, \textit{description}, \textit{location}, \textit{timezone}, \textit{createdat}, 
\textit{favoritescount}, \textit{followerscount}, \textit{friendscount},
\textit{listedcount}, and \textit{statusescount}.
Most vertices labeled with \textit{tweet} ($>80\%$) have properties \textit{createdat},
\textit{createdstr} and \textit{text}. About $25\%$ of them have additional properties
\textit{favoritecount}, \textit{retweetcount}, and \textit{retweeted}.
Edge do not have properties but each edge has a label which can be
\textit{posted} ($41\%$), 
\textit{hastag} ($22\%$), 
\textit{postedvia} ($12\%$), 
\textit{mentions} ($11\%$), 
\textit{retweeted} ($8\%$), 
\textit{haslink} ($6\%$), or
\textit{inreplyto} ($<1\%$).

\marginpar{\Large\vspace{1em}\colorbox{yellow}{\textbf{Cmds}}}

\subsection{\colorbox{yellow}{Differences to Traditional GNN Datasets}}

\hl{The main difference between LPG graphs and traditional GNN datasets such as Citeseer or Cora is
that the latter usually do not have extensive sets of labels. Instead, these
datasets often have vertices from different classes, which may be interpreted as a single label
(that would encode such different classes). Moreover, these datasets often do not have rich sets
of attached \emph{different} properties. Instead, they may come with extensive feature vectors that
encode a single large additional piece of information, for example a whole abstract. 
Finally, in the graph database setting, it is less common to process graphs such as PROTEINS, where
the dataset consists of a very large number of relatively small graph. Instead, it is more common
to focus on one large graph dataset.}

\section{Results for Additional Labels, Properties, and Datasets}

\begin{figure}[h]
\vspaceSQ{-0.5em}
    \centering
    \includegraphics[width=1.0\textwidth]{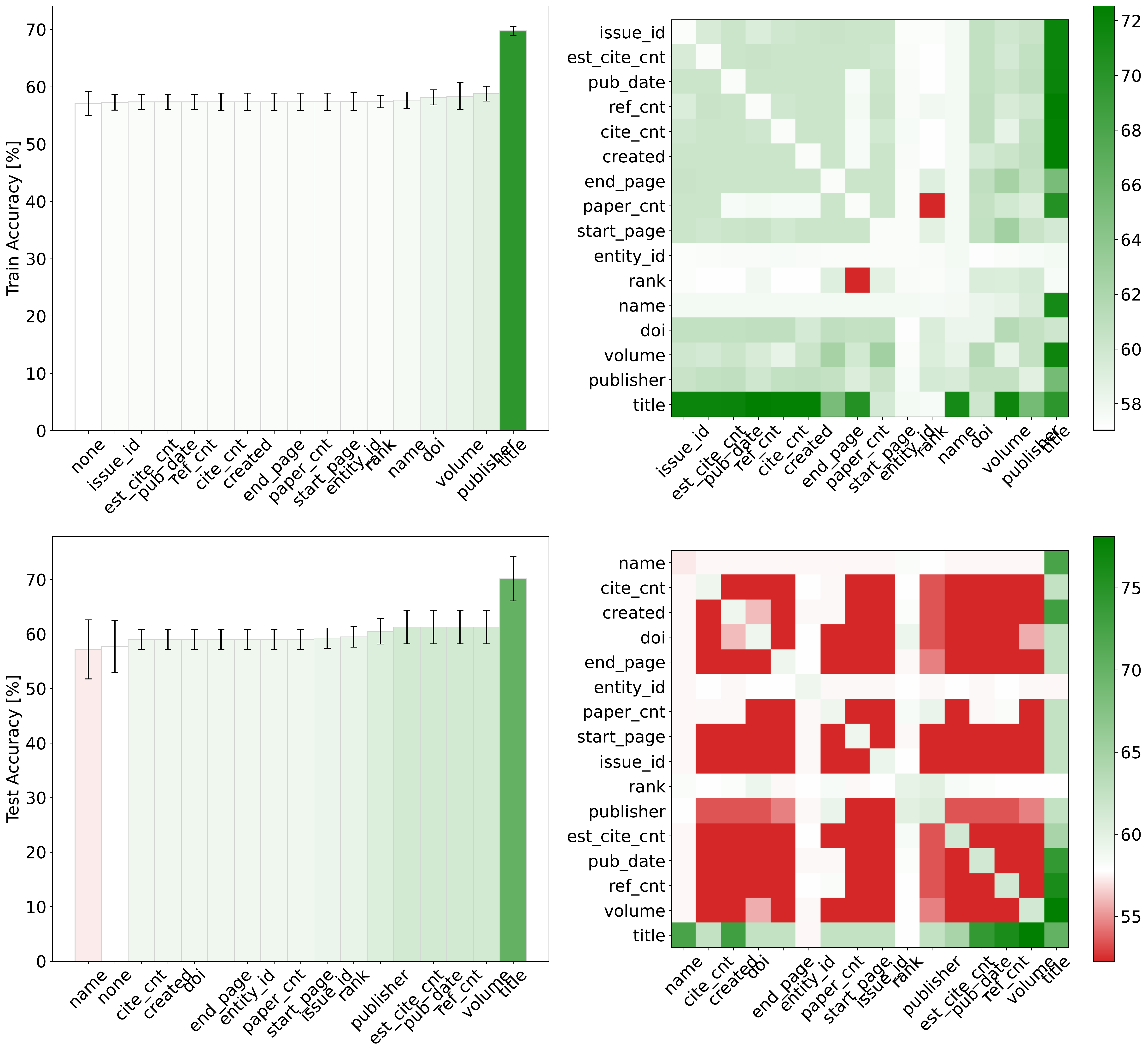}
    \vspaceSQ{-0.25em}
    \caption{\textbf{MAKG small (node classification, 4 classes, results aggregated over all three models)}. Impact from different properties and their combinations on the accuracy.
    Green: accuracy is better than that of a graph with
    no labels/properties; red: the accuracy is worse than that of a graph with
            no labels/properties.}
    \label{fig:makg-small-hm}
\end{figure}

\begin{figure}[h]
\vspaceSQ{-1.5em}
    \centering
    \includegraphics[width=1.0\textwidth]{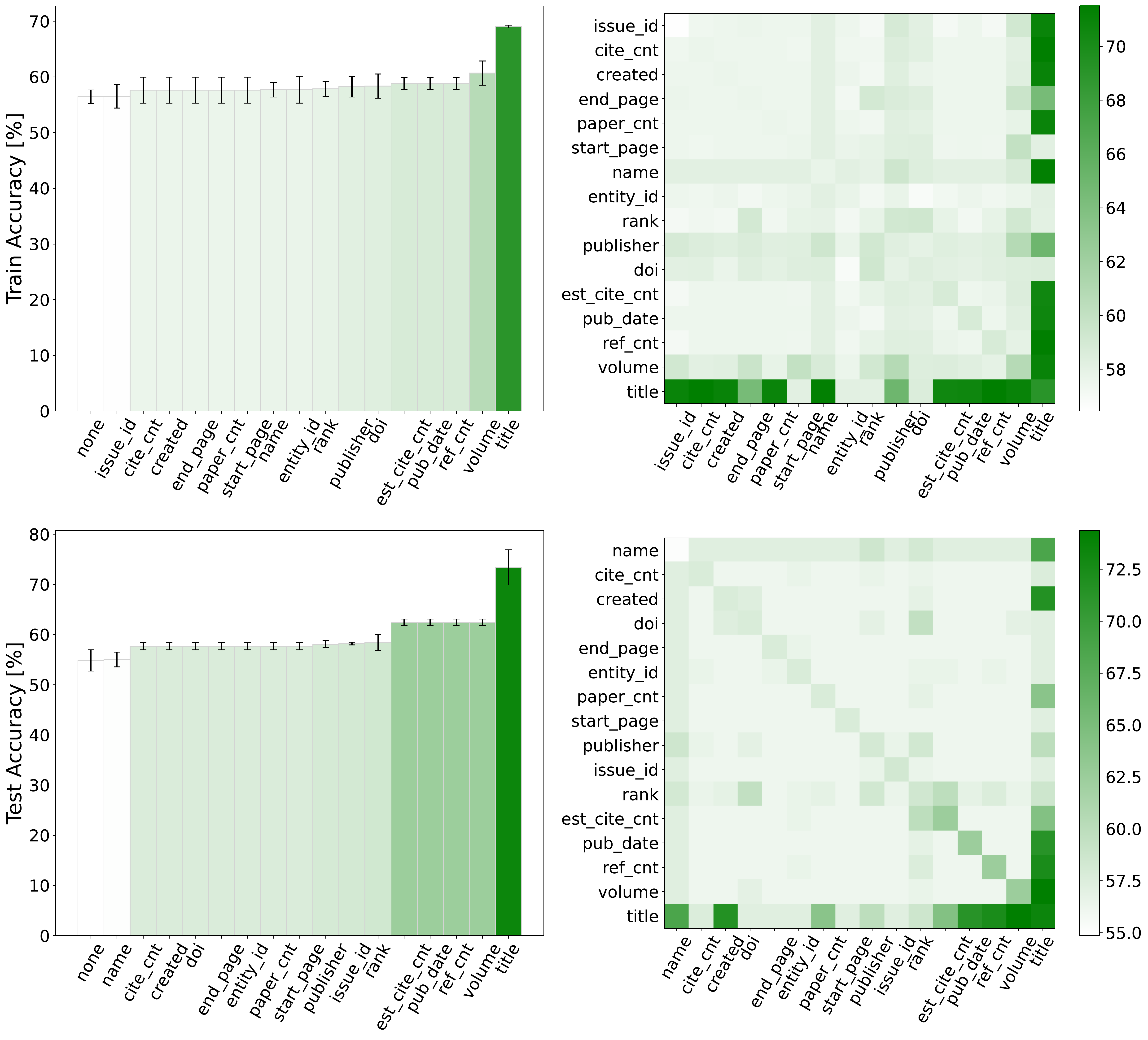}
    \vspaceSQ{-0.25em}
    \caption{\textbf{MAKG small (node classification, 4 classes, GCN-only results)}. Impact from different properties and their combinations on the accuracy.
    Green: the accuracy is better than that of a graph with
    no labels/properties; red: the accuracy is worse than that of a graph with
        no labels/properties.}
    \label{fig:makg-small-hm-gcn}
\end{figure}

\begin{figure}[h]
\vspaceSQ{-1.5em}
    \centering
    \includegraphics[width=1.0\textwidth]{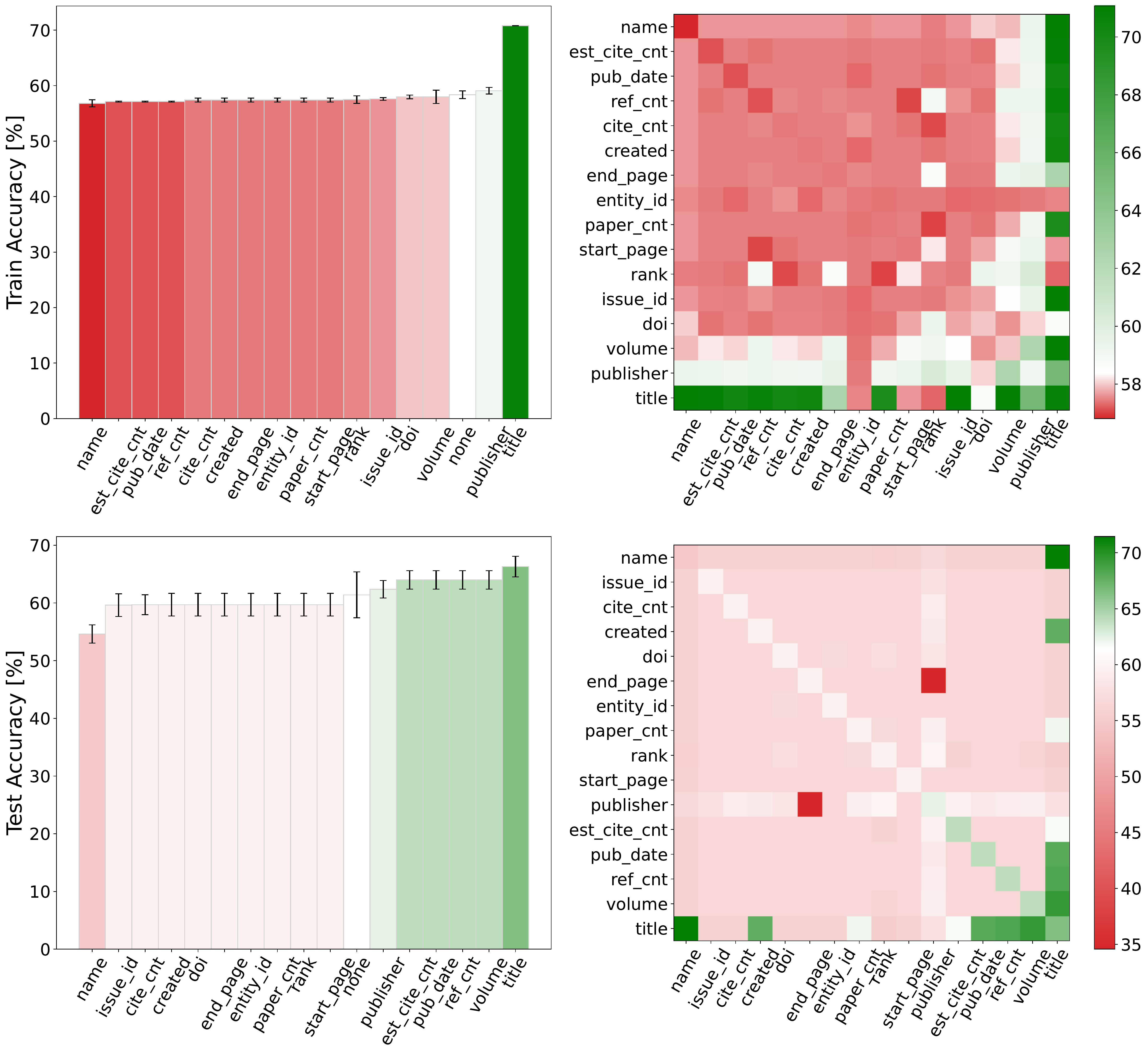}
    \vspaceSQ{-0.25em}
    \caption{\textbf{MAKG small (node classification, 4 classes, GAT-only results)}. Impact from different properties and their combinations on the accuracy.
    Green: accuracy is better than that of a graph with
    no labels/properties; red: the accuracy is worse than that of a graph with
            no labels/properties.}
    \label{fig:makg-small-hm-gat}
\end{figure}

\begin{figure}[h]
\vspaceSQ{-1.5em}
    \centering
    \includegraphics[width=1.0\textwidth]{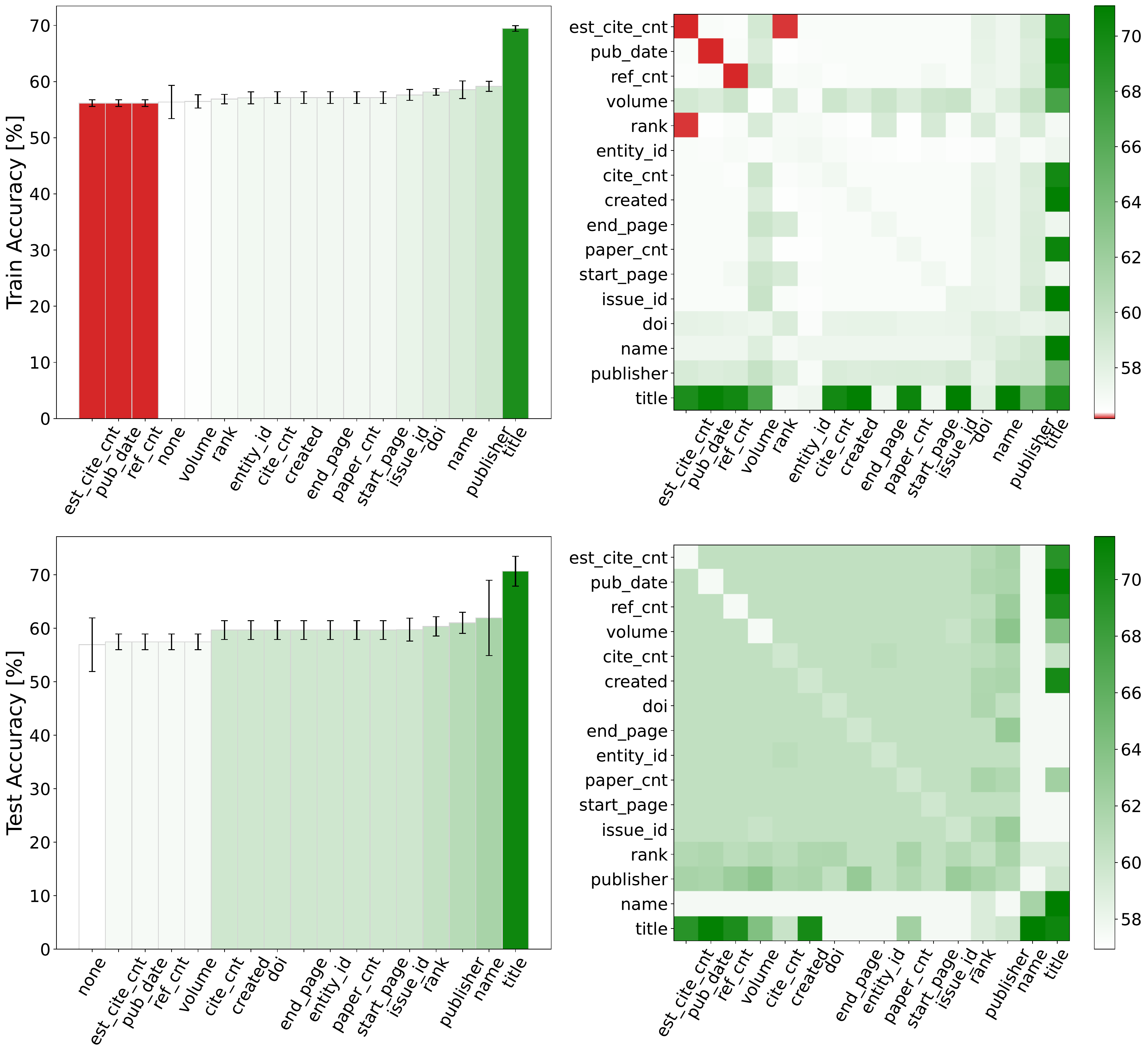}
    \vspaceSQ{-0.25em}
    \caption{\textbf{MAKG small (node classification, 4 classes, GIN-only results)}. Impact from different properties and their combinations on the accuracy.
    Green: accuracy is better than that of a graph with
    no labels/properties; red: the accuracy is worse than that of a graph with
            no labels/properties.}
    \label{fig:makg-small-hm-gin}
\end{figure}

\begin{figure}[h]
\vspaceSQ{-1.5em}
    \centering
    \includegraphics[width=1.0\textwidth]{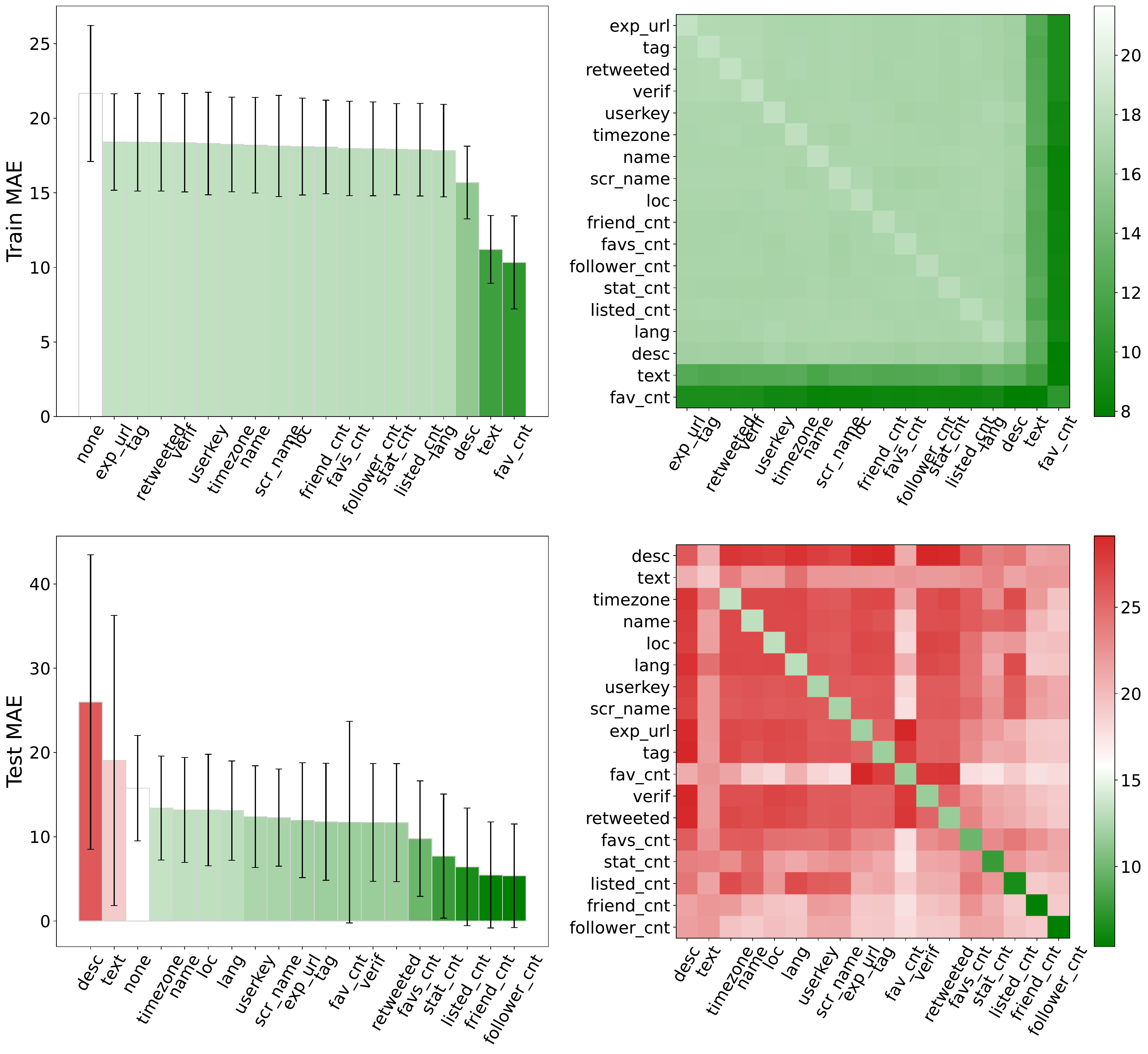}
    \vspaceSQ{-0.25em}
    \caption{\textbf{Neo4j Twitter trolls (node regression, results aggregated over all three models)}. Impact from different properties and their combinations on the MAE.
    Green: MAE is better than that of a graph with
    no labels/properties; red: the MAE is worse than that of a graph with
            no labels/properties.}
    \label{fig:tweeter-hm}
\end{figure}

\begin{figure}[h]
\vspaceSQ{-1.5em}
    \centering
    \includegraphics[width=1.0\textwidth]{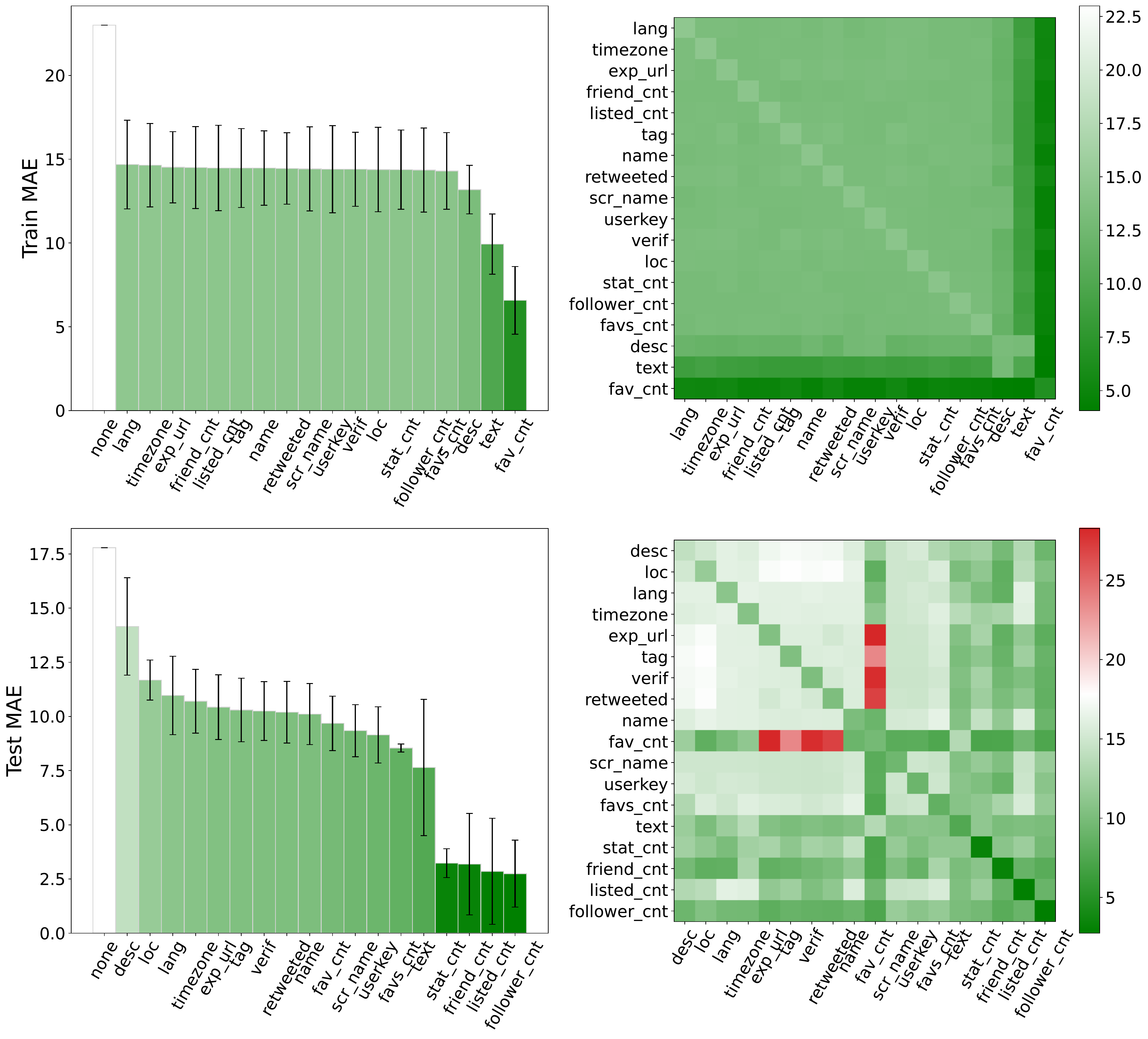}
    \vspaceSQ{-0.25em}
    \caption{\textbf{Neo4j Twitter trolls (node regression, GCN-only results)}. Impact from different properties and their combinations on the MAE.
    Green: MAE is better than that of a graph with
    no labels/properties; red: the MAE is worse than that of a graph with
            no labels/properties.}
    \label{fig:tweeter-hm-gcn}
\end{figure}

\begin{figure}[h]
\vspaceSQ{-1.5em}
    \centering
    \includegraphics[width=1.0\textwidth]{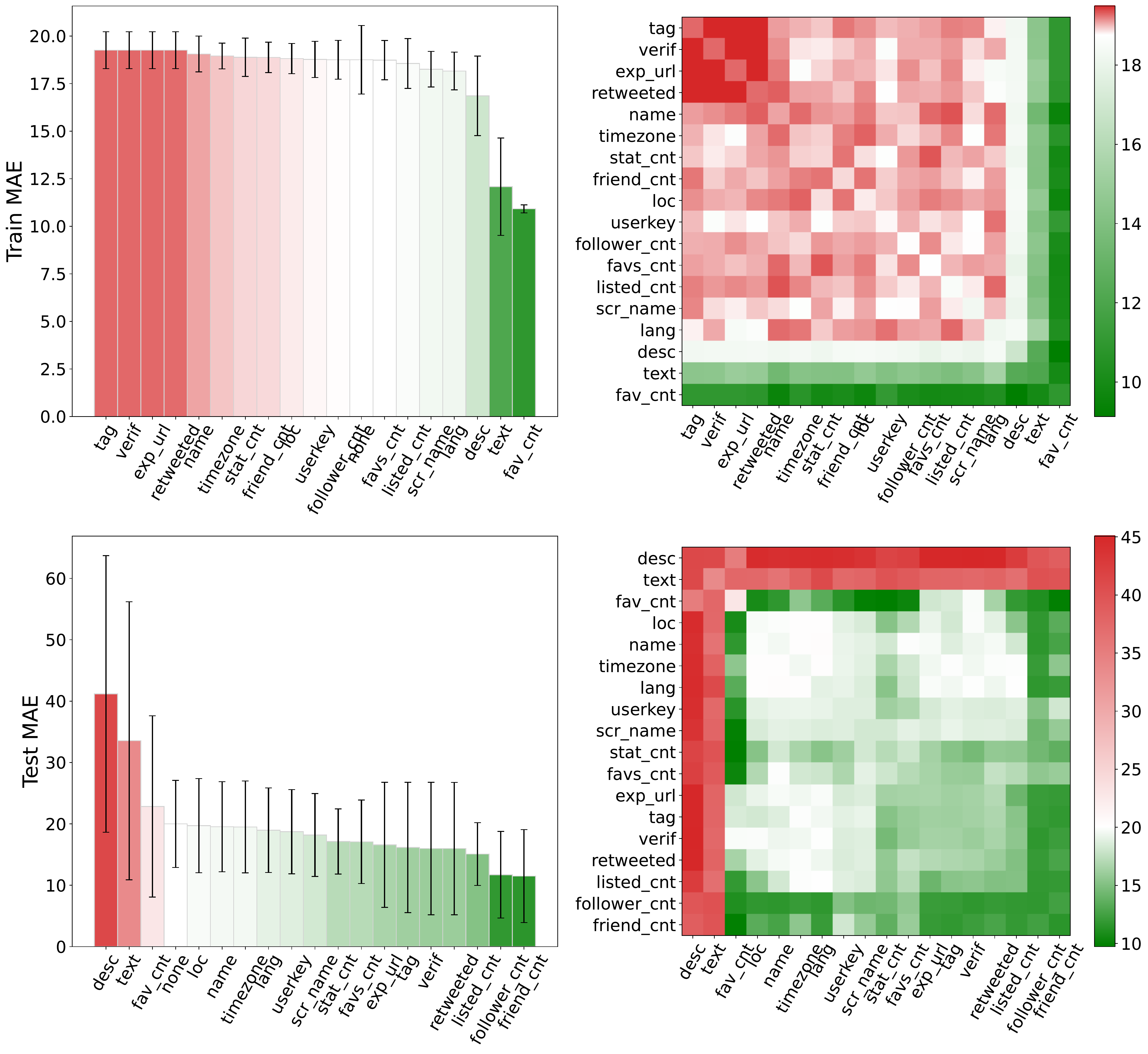}
    \vspaceSQ{-0.25em}
    \caption{\textbf{Neo4j Twitter trolls (node regression, GAT-only results)}. Impact from different properties and their combinations on the MAE.
    Green: MAE is better than that of a graph with
    no labels/properties; red: the MAE is worse than that of a graph with
            no labels/properties.}
    \label{fig:tweeter-hm-gat}
\end{figure}

\begin{figure}[h]
\vspaceSQ{-1.5em}
    \centering
    \includegraphics[width=1.0\textwidth]{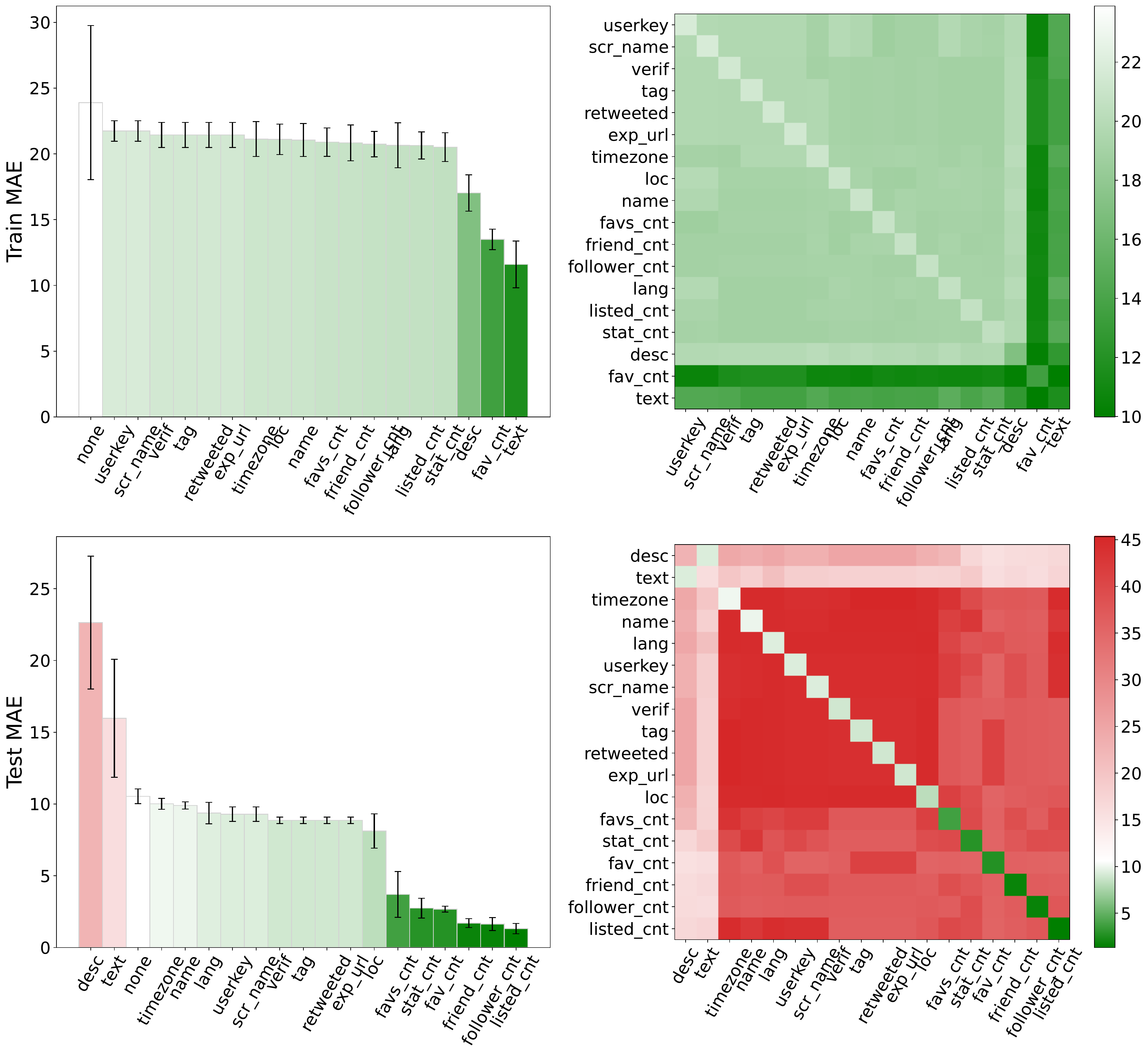}
    \vspaceSQ{-0.25em}
    \caption{\textbf{Neo4j Twitter trolls (node regression, GIN-only results)}. Impact from different properties and their combinations on the MAE.
    Green: MAE is better than that of a graph with
    no labels/properties; red: the MAE is worse than that of a graph with
            no labels/properties.}
    \label{fig:tweeter-hm-gin}
\end{figure}

\begin{figure}[h]
\vspaceSQ{-1.5em}
    \centering
    \includegraphics[width=0.8\textwidth]{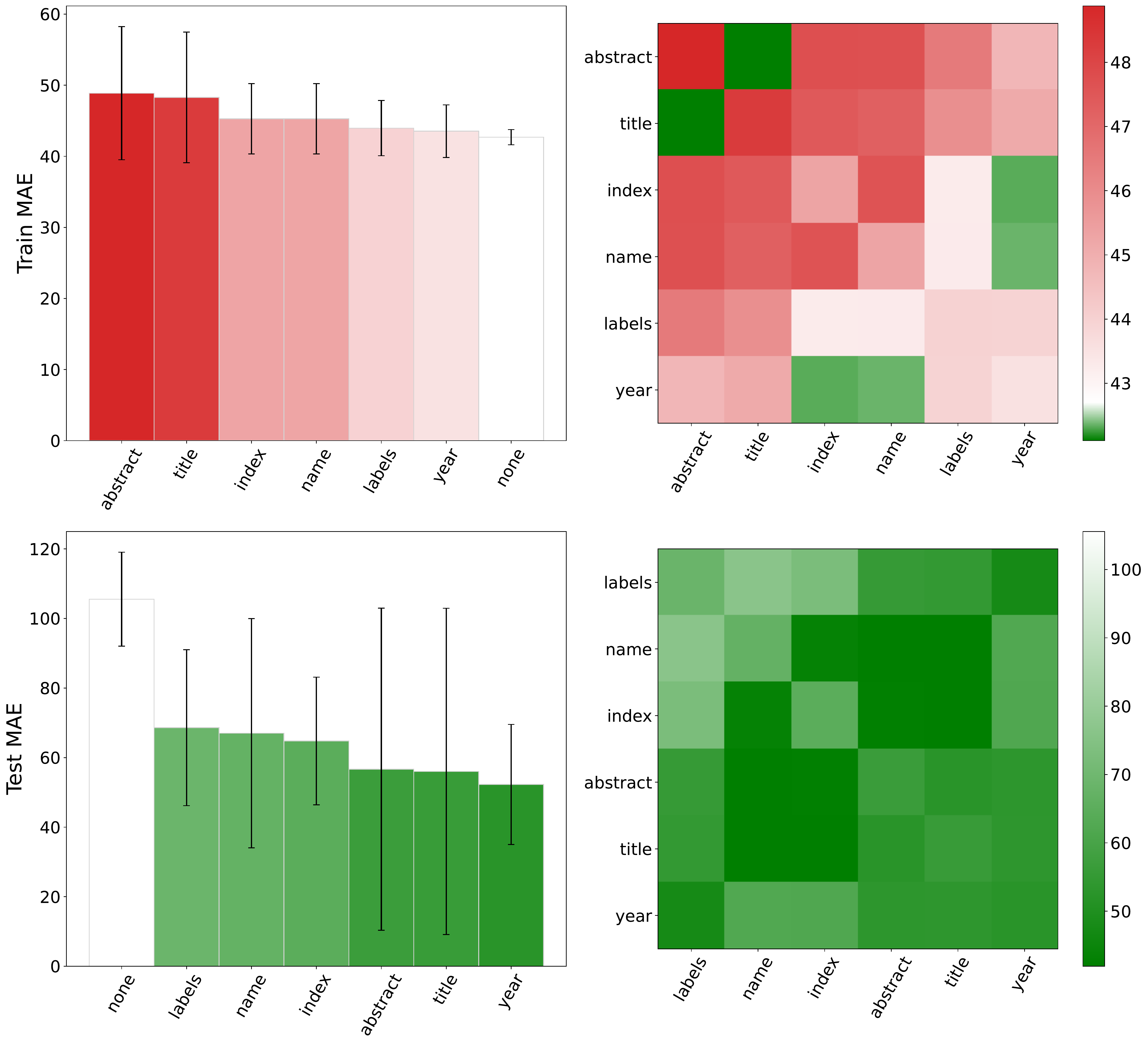}
    \vspaceSQ{-0.25em}
    \caption{\textbf{Neo4j citations (node regression, results aggregated over all three models)}. Impact from different properties and their combinations on the MAE.
    Green: MAE is better than that of a graph with
    no labels/properties; red: the MAE is worse than that of a graph with
            no labels/properties.}
    \label{fig:citations-hm}
\end{figure}

\begin{figure}[h]
\vspaceSQ{-1.5em}
    \centering
    \includegraphics[width=0.8\textwidth]{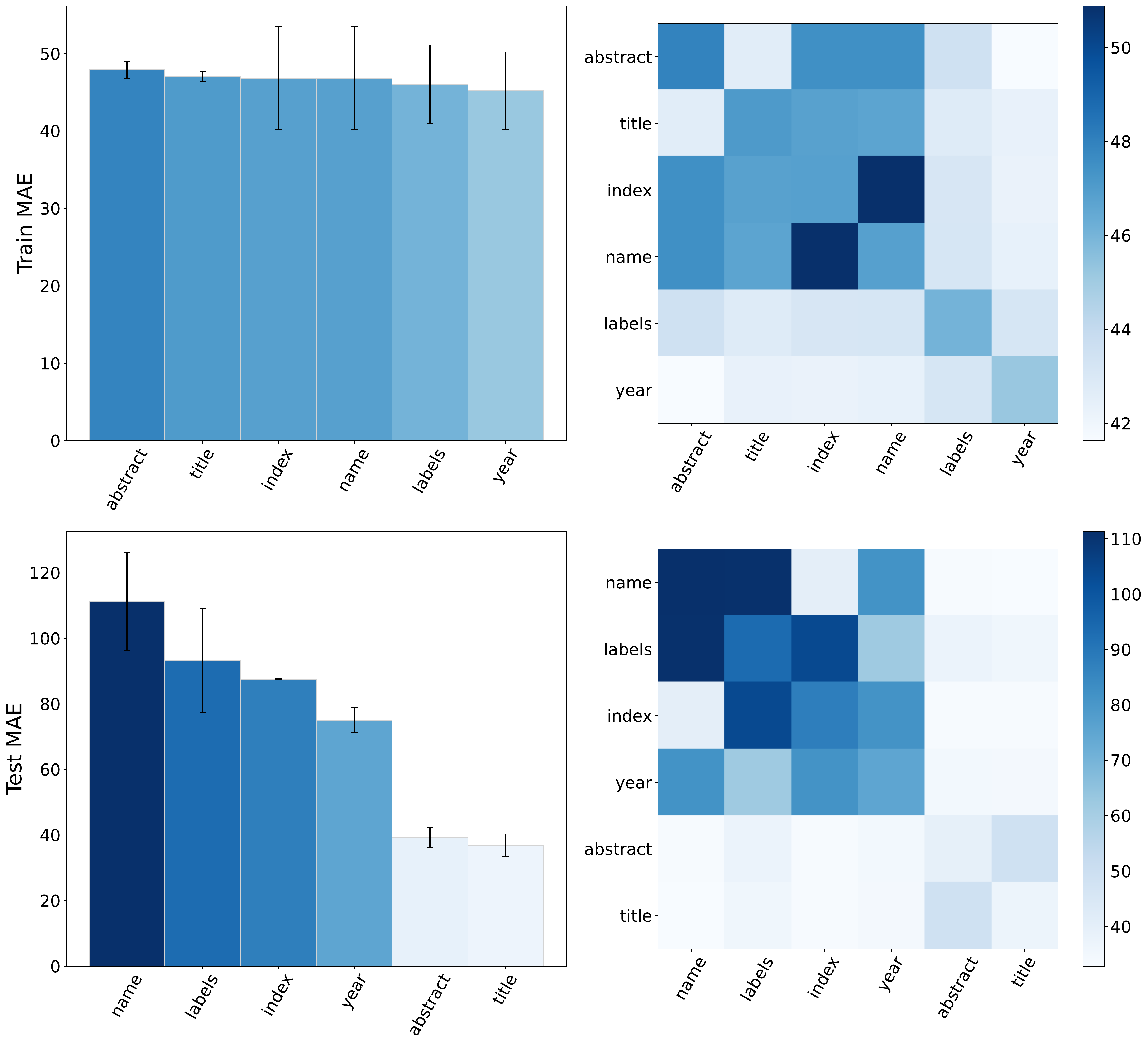}
    \vspaceSQ{-0.25em}
    \caption{\textbf{Neo4j citations (node regression, GCN-only results)}. Impact from different properties and their combinations on the MAE.
            Here, we do not use green/red colors, because the baselines with no labels/properties could not converge. Instead,
            we use only one-color (blue) shades to indicate relative improvements.}
    \label{fig:citations-hm-gcn}
\end{figure}

\begin{figure}[h]
\vspaceSQ{-1.5em}
    \centering
    \includegraphics[width=0.8\textwidth]{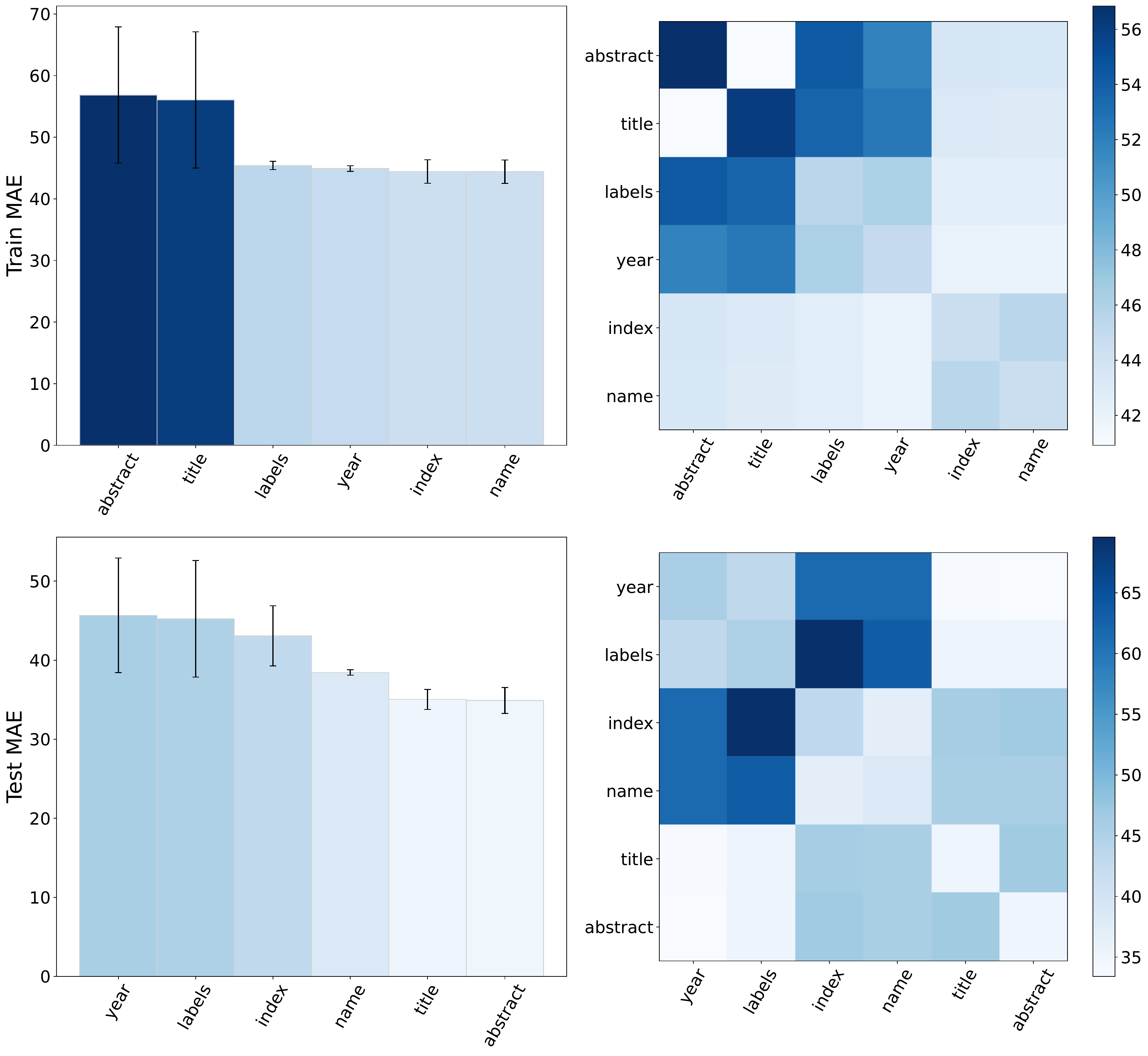}
    \vspaceSQ{-0.25em}
    \caption{\textbf{Neo4j citations (node regression, GAT-only results)}. Impact from different properties and their combinations on the MAE.
            Here, we do not use green/red colors, because the baselines with no labels/properties could not converge. Instead,
            we use only one-color (blue) shades to indicate relative improvements.}
    \label{fig:citations-hm-gat}
\end{figure}

\begin{figure}[h]
\vspaceSQ{-1.5em}
    \centering
    \includegraphics[width=0.8\textwidth]{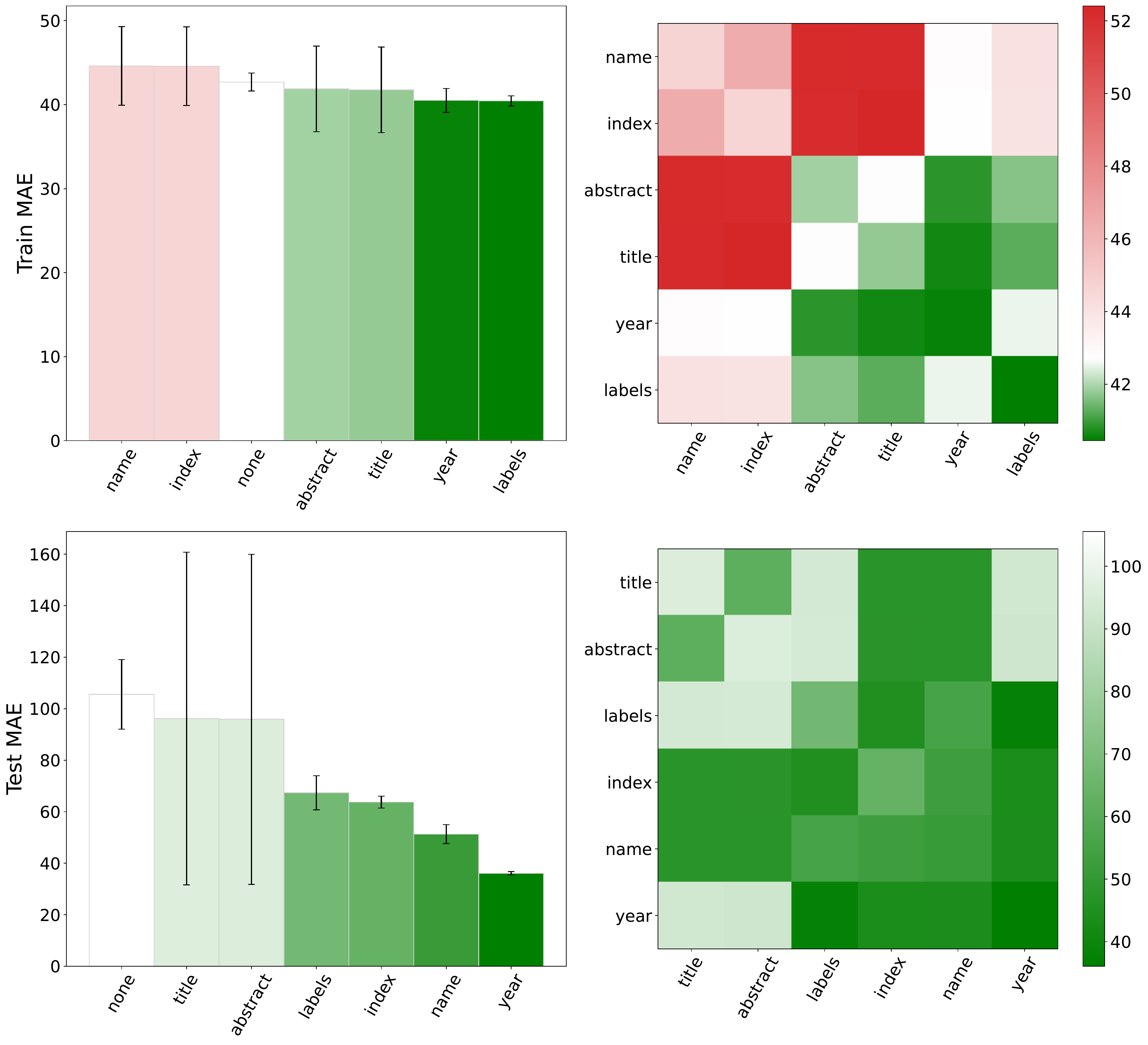}
    \vspaceSQ{-0.25em}
    \caption{\textbf{Neo4j citations (node regression, GIN-only results)}. Impact from different properties and their combinations on the MAE.
    Green: MAE is better than that of a graph with
    no labels/properties; red: the MAE is worse than that of a graph with
            no labels/properties.}
    \label{fig:citations-hm-gin}
\end{figure}

\begin{figure}[h]
\vspaceSQ{-1.5em}
    \centering
    \includegraphics[width=1.0\textwidth]{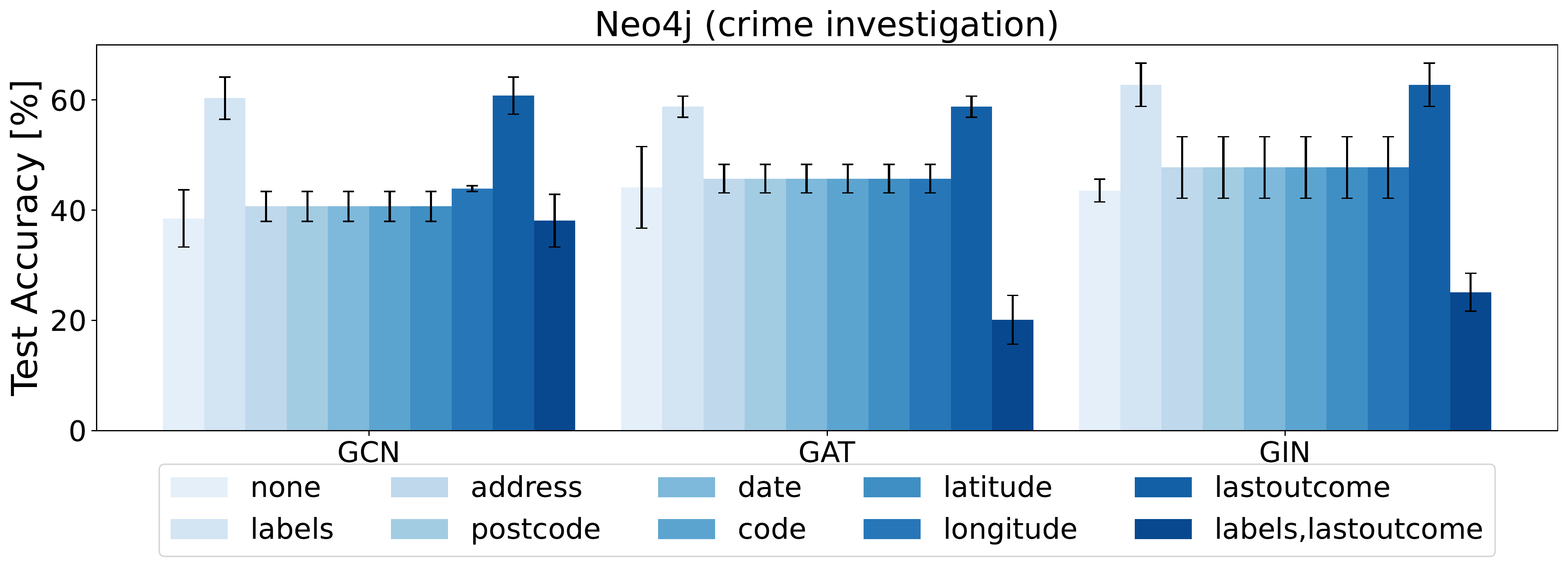}
    \vspaceSQ{-0.25em} \caption{Advantages of preserving the
    information encoded in LPG labels and properties, for node classification in the Neo4j crime investigation dataset.}
    \label{fig:pole-main}
\end{figure}

Figures~\ref{fig:makg-small-hm}--\ref{fig:citations-hm-gin}
illustrate the impact of using each of the many available properties, and pairs
of properties, on the final prediction accuracy. We show results separately for
each GNN model and also aggregated for all thee models, for the completeness of
the analysis. To facilitate comparing the data, we also replot the results for
the GIN model for MAKG small and Neo4j Twitter analyses from Section~\ref{sec:eval}.
Finally, Figure~\ref{fig:pole-main} shows results
for an additional Neo4j dataset modeling crime investigations.

\marginpar{\Large\vspace{1em}\colorbox{yellow}{\textbf{x3Ya}}}

\hl{Interestingly, the largest accuracy increase for MAKG is
consistently obtained when including the title property. This is
the case for all the considered GNN models. Similarly, when
detecting trolls, including the counts of friends or followers
was crucial in consistent accuracy improvements. This indicates
that it is more important to appropriately understand the data
and include the right information in the input feature vectors,
and once this is achieved, different GNN models would be similarly
able to extract this information for more accurate outcomes.}

\clearpage
\section{\colorbox{yellow}{Details of Embedding Construction}}

\marginpar{\Large\vspace{-1em}\colorbox{yellow}{\textbf{n77k}}}

\marginpar{\Large\vspace{1em}\colorbox{yellow}{\textbf{n77k}}}

\hl{We provide formal specifications of the computed LPG2vec encodings for any vertex $i$ ($\mathbf{x}_i$) and for any edge $(i,j)$ ($\mathbf{x}_{ij}$).
The specific fields are as follows: one-hot encoding of the $x$-th label ($l_x$) where $x \in \{1, ..., L\}$,
one-hot encoding of the $y$-th property that has $C_y$ potential values ($p_{y,1}, p_{y,2}, ..., p_{y,C_y}$) where $y \in \{1, ..., P\}$,
and a string encoding (e.g., BERT) of the $z$-th text feature that has $T_z$ potential fields ($f_{z,1}, f_{z,2}, ..., f_{z,T_z}$) where $z \in \{1, ..., F\}$.
This formal description assumes that all the properties are appropriately discretized and - if needed - normalized.
The encoding for edges is fully analogous (for simplicity, we assume that the set of labels and properties $L \cup P$ is common for vertices and edges).}

\footnotesize
\[
{\mathbf{x}_i} = \left( 
                                  \begin{array}{c}
                                         l_1\\
                                         l_2\\
                                         \vdots\\
                                         l_{L}\\
                                         p_{1,1}\\
                                         p_{1,2}\\
                                         \vdots\\
                                         p_{1,C_1}\\
                                         p_{2,1}\\
                                         p_{2,2}\\
                                         \vdots\\
                                         p_{2,C_2}\\
                                         \vdots\\
                                         p_{P,1}\\
                                         p_{P,2}\\
                                         \vdots\\
                                         p_{P,C_P}\\
                                         f_{1,1}\\
                                         f_{1,2}\\
                                         \vdots\\
                                         f_{1,T_1}\\
                                         f_{2,1}\\
                                         f_{2,2}\\
                                         \vdots\\
                                         f_{2,T_2}\\
                                         \vdots\\
                                         f_{F,1}\\
                                         f_{F,2}\\
                                         \vdots\\
                                         f_{F,T_F}\\
                                  \end{array}
                            \right)
\]
\normalsize

\footnotesize
\[
{\mathbf{e}_{i,j}} = \left( 
                                  \begin{array}{c}
                                         l_1\\
                                         l_2\\
                                         \vdots\\
                                         l_{L}\\
                                         p_{1,1}\\
                                         p_{1,2}\\
                                         \vdots\\
                                         p_{1,C_1}\\
                                         p_{2,1}\\
                                         p_{2,2}\\
                                         \vdots\\
                                         p_{2,C_2}\\
                                         \vdots\\
                                         p_{P,1}\\
                                         p_{P,2}\\
                                         \vdots\\
                                         p_{P,C_P}\\
                                         f_{1,1}\\
                                         f_{1,2}\\
                                         \vdots\\
                                         f_{1,F_1}\\
                                         f_{2,1}\\
                                         f_{2,2}\\
                                         \vdots\\
                                         f_{2,F_2}\\
                                         \vdots\\
                                         f_{F,1}\\
                                         f_{F,2}\\
                                         \vdots\\
                                         f_{F,C_F}\\
                                  \end{array}
                            \right)
\]
\normalsize

\newpage
\section{\colorbox{yellow}{Results for Additional Hyperparameters and Models}}

\marginpar{\Large\vspace{-1em}\colorbox{yellow}{\textbf{Cmds}}\\ \colorbox{yellow}{\textbf{n77k}}\\ \colorbox{yellow}{\textbf{x3Ya}}}

\begin{figure}[h]
    \centering
    \includegraphics[width=1.0\textwidth]{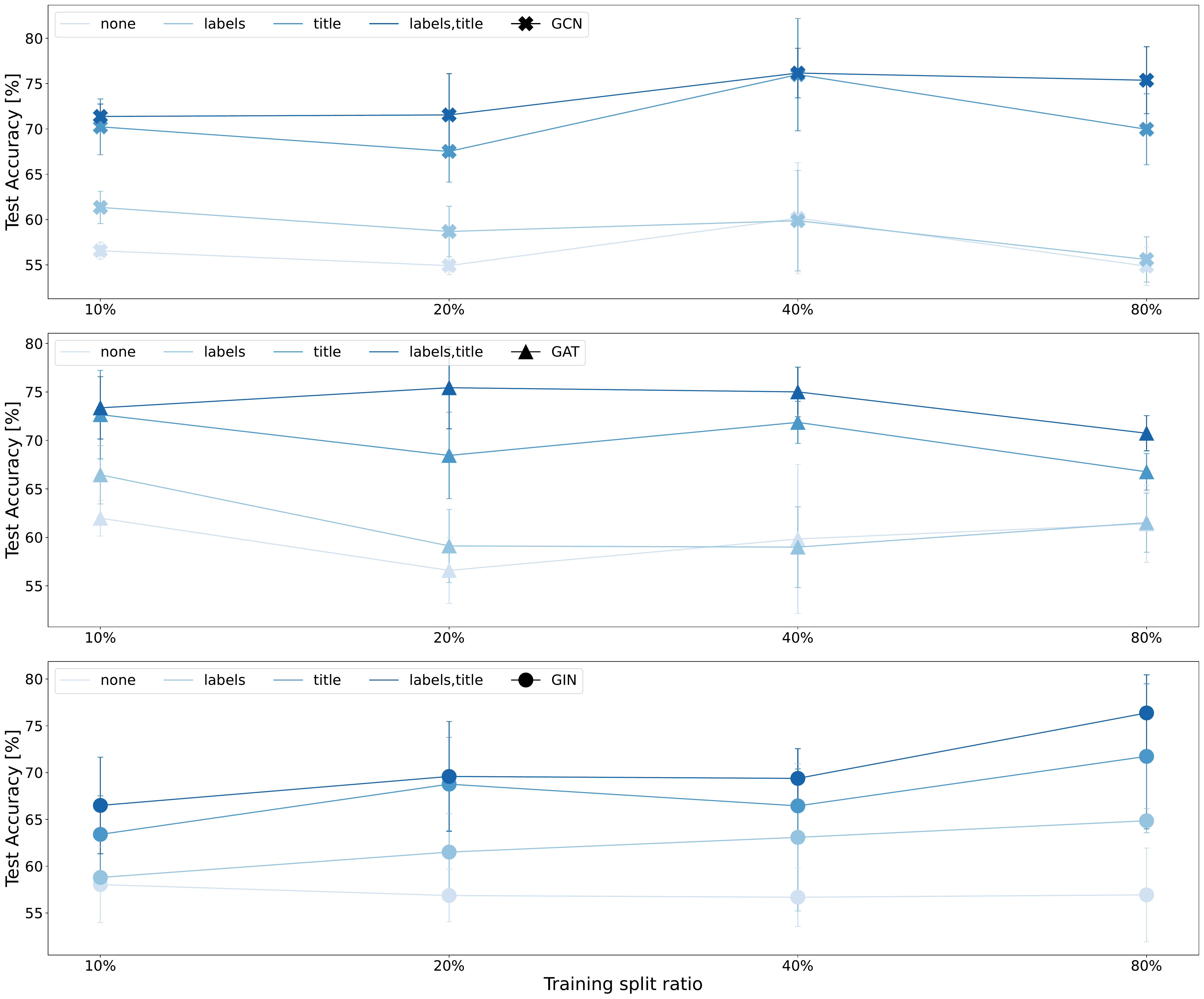}
    \vspaceSQ{-0.25em}
    \caption{\hl{\textbf{MAKG small (node classification, 4 classes)}. Impact from different split ratios (the higher the better).}}
    \label{fig:makg-split}
\end{figure}

\begin{figure}[h]
\vspaceSQ{-1.5em}
    \centering
    \includegraphics[width=1.0\textwidth]{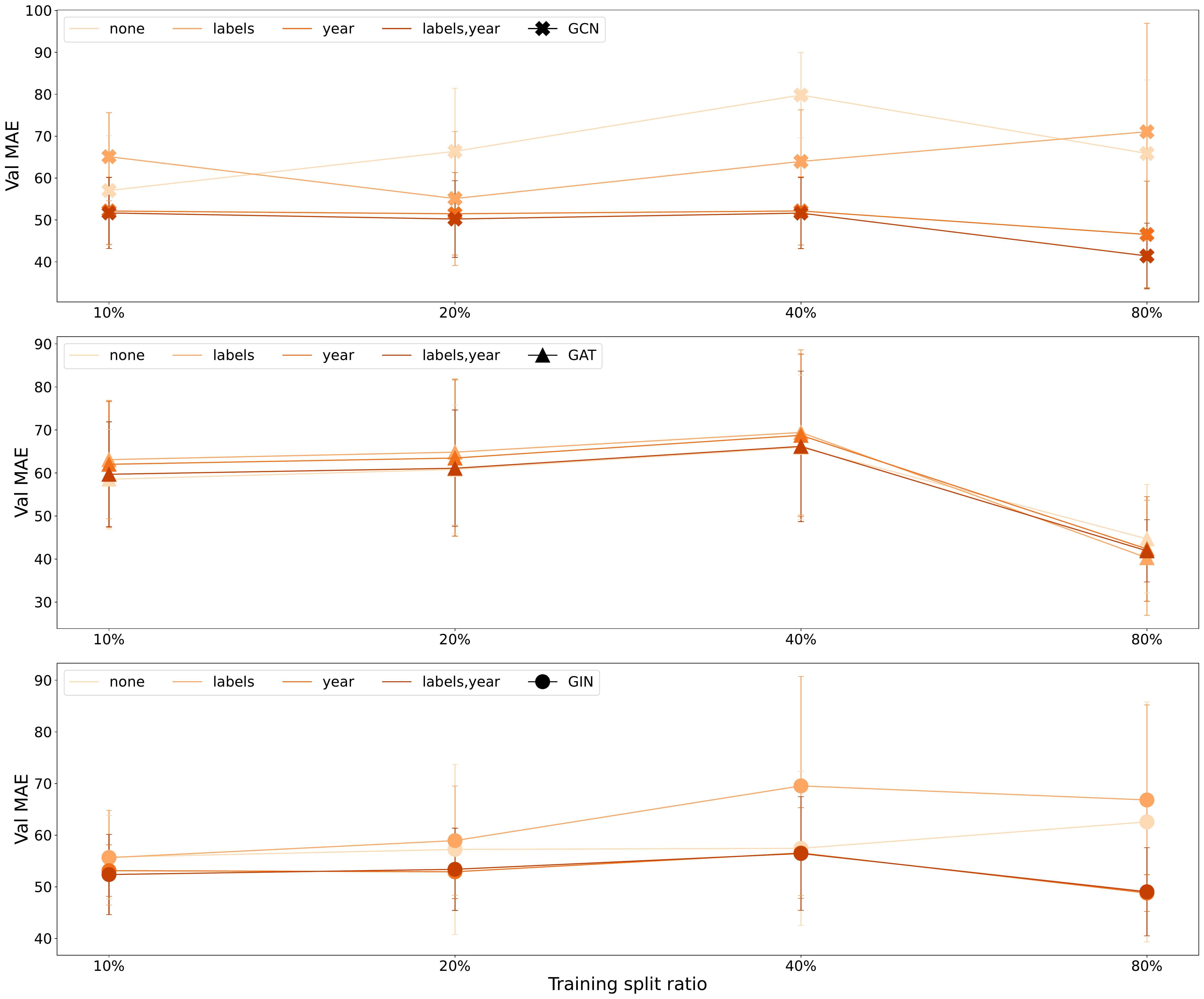}
    \vspaceSQ{-0.25em}
    \caption{\hl{\textbf{Neo4j citations (node regression)}. Impact from different split ratios (the lower the better).}}
    \label{fig:cit-split}
\end{figure}

\hl{We also investigate different training split ratios as well as the counts of convolution layers, see Figures~\mbox{\ref{fig:makg-split}} and~\mbox{\ref{fig:cit-split}}.
Adding node features generally improves the accuracy across different GNN models and splits.
Differences in the training split ratio for the MAKG dataset have little effect on the accuracy.
However, in the citations dataset, the accuracy gets worse when it uses more training data. It
indicates that, in this dataset and task, the initial 80\% split ratio for the training nodes is too high.
}

\begin{figure}[h]
    \centering
    \includegraphics[width=1.0\textwidth]{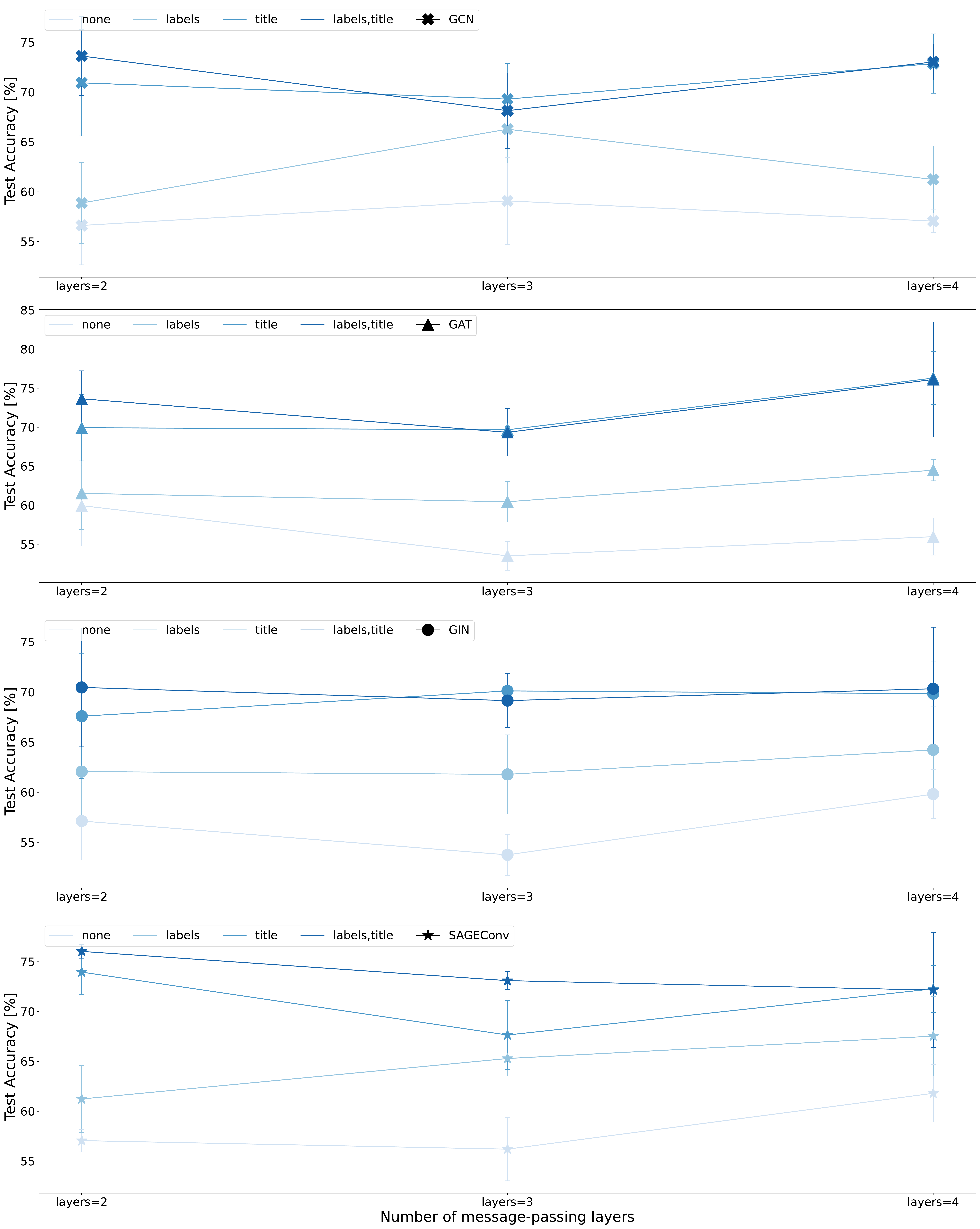}
    \vspaceSQ{-0.25em}
    \caption{\hl{\textbf{MAKG small (node classification, 4 classes)}. Impact from different counts of convolution layers (the higher the better).}}
    \label{fig:layers}
\end{figure}

\newpage
\hl{We also vary the number convolution layers, see Figure~\mbox{\ref{fig:layers}}.
Adding more layers on its own does not bring consistent improvements. This is because 
the structure of the considered graph datasets usually has a lot of locality and
is highly clustered. However, importantly, adding the information from labels and
from properties enhances the accuracy consistency across all tried layer counts.}

\newpage
\hl{We also investigated different hyperparameters for LPG2vec embeddings.
For example, we experimented with the dimensions of the constructed embeddings.
For this, we 
tried to use an additional MLP to reduce the
dimensions of the high dimensional LPG2vec feature vectors, while keeping the information within the features
intact.
We use two linear layers combined
with a dropout layer and the Leaky Relu activation.
The dimensions were reduced by different rations, between 20 and 5$\times$.
This approach on one hand resulted in much smaller input feature vectors, which could visibly reduce
the memory storage overheads for particularly large graphs. However,
we also observed consistent accuracy losses across all tried datasets and GNN models.
We left more extensive experiments into this direction for future work.}

\marginpar{\Large\vspace{-5em}\colorbox{yellow}{\textbf{n77k}}}

\begin{figure}[h]
    \centering
    \includegraphics[width=1.0\textwidth]{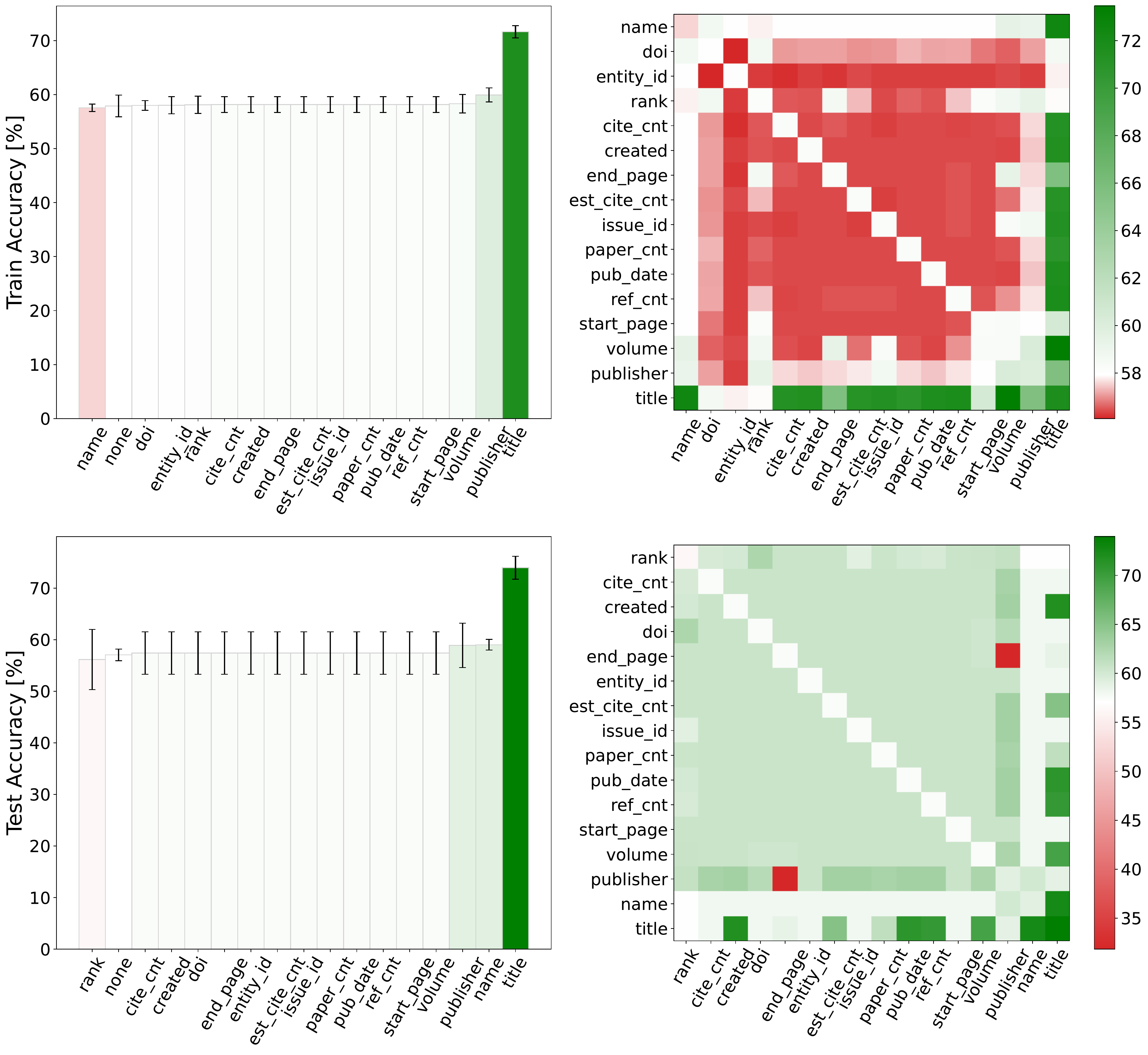}
    \vspaceSQ{-0.25em}
    \caption{\hl{\textbf{MAKG small (node classification, 4 classes, GraphSAGE-only results)}. Impact from different properties and their combinations on the accuracy.
    Green: the accuracy is better than that of a graph with
    no labels/properties; red: the accuracy is worse than that of a graph with
        no labels/properties.}}
    \label{fig:sageconv}
\end{figure}

\begin{figure}[h]
\vspaceSQ{-1.5em}
    \centering
    \includegraphics[width=1.0\textwidth]{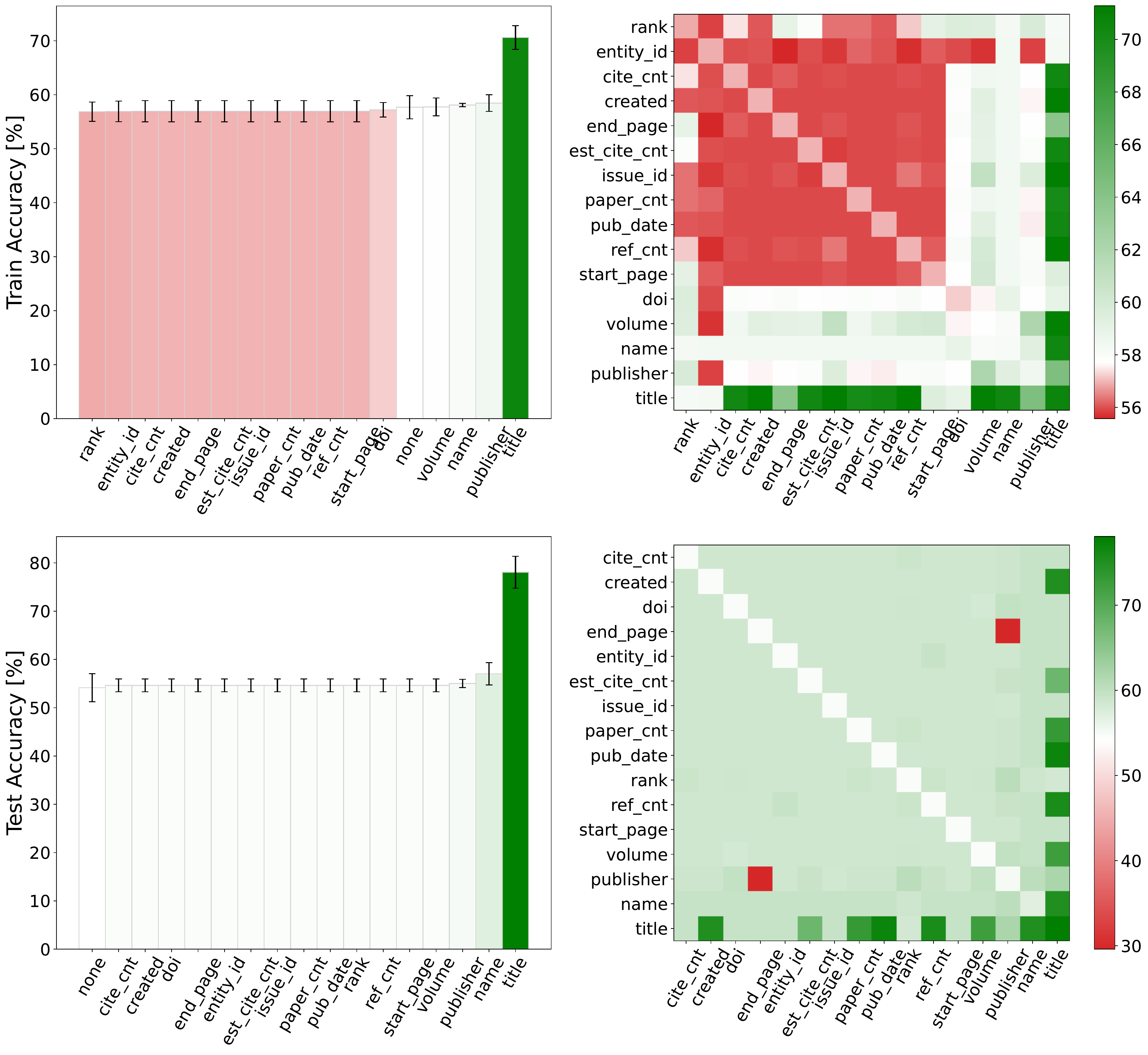}
    \vspaceSQ{-0.25em}
    \caption{\hl{\textbf{MAKG small (node classification, 4 classes, MLP-only results)}. Impact from different properties and their combinations on the accuracy.
    Green: the accuracy is better than that of a graph with
    no labels/properties; red: the accuracy is worse than that of a graph with
        no labels/properties.}}
    \label{fig:mlp}
\end{figure}

\hl{Finally, we also investigate additional models, GraphSAGE (Figure~\mbox{\ref{fig:sageconv}}) and plain MLP (Figure~\mbox{\ref{fig:mlp}}).
As with GCN, GIN, and GAT, adding more labels and more properties enhances the accuracy. MLP comes with much lower accuracy than GraphSAGE
for most tried settings (i.e., with most of labels and properties tried). However, interestingly, it becomes only slightly
less powerful than GraphSAGE when including the title property. This further shows the importance of harnessing LPG data - 
when the right data is included into the initial embeddings, it may offer very high accuracy even without considering the
graph structure.}

\marginpar{\Large\vspace{-5em}\colorbox{yellow}{\textbf{Cmds}}\\ \colorbox{yellow}{\textbf{n77k}}\\ \colorbox{yellow}{\textbf{x3Ya}}}

\end{document}